\documentclass[journal]{IEEEtran}
\usepackage{subcaption}
\usepackage{amsfonts}
\usepackage{mathtools} 
\DeclarePairedDelimiter{\norm}{\lVert}{\rVert}
\usepackage{graphicx}
\usepackage{parskip}
\usepackage{placeins}
\usepackage{amssymb} 
\usepackage{float}
\usepackage{enumitem}
\usepackage{array,multirow,tabularx}
\usepackage[utf8]{inputenc}
\usepackage{xspace}
\usepackage{wrapfig}
\usepackage[dvipsnames,table,svgnames]{xcolor}
\usepackage[many]{tcolorbox}
\usepackage{tikz,lipsum,lmodern}
\usepackage{mathtools, cuted}
\usepackage{algpseudocode}
\usepackage{algorithm}
\usepackage{multicol}
\usepackage{gensymb}
\usepackage{units}
\usepackage{parskip}
\usepackage[font={footnotesize,it},labelfont=bf,
   justification=justified,
   format=plain]{caption}
\usepackage{titlesec}
\titleformat{\section}
  {\normalfont\fontsize{11}{15}\bfseries}{\thesection.}{1em}{}

\titleformat{\subsection}
  {\normalfont\fontsize{10}{15}\bfseries}{\thesubsection}{1em}{}

\graphicspath{{./pdf/}{./jpg/}}
\DeclareGraphicsExtensions{.pdf,.jpg,.png}
\usepackage{lipsum}

\makeatletter
\newcommand{\mathleft}{\@fleqntrue\@mathmargin0pt}
\newcommand{\mathcenter}{\@fleqnfalse}
\makeatother

\DeclarePairedDelimiter{\abs}{\lvert}{\rvert}

\floatstyle{plain}
\newfloat{twocolequfloat}{b}{zzz}
\floatname{twocolequfloat}{Equation}

\renewcommand{\thesubsection}{\Alph{subsection}}
\ifCLASSINFOpdf
\else
\fi
\begin{document}
%
\title{Efficient Attitude Estimators: A 
Tutorial and Survey}
%
%
%

\author{Hussein~Al-Jlailaty, Mohammad~M.~Mansour
\thanks{M. Al-Jlailaty and M. Mansour are with the Department
of Electrical and Computer Engineering, Maroun Semaan Faculty of Engineering   and   Architecture, American University of Beirut, Lebanon.
 e-mail: (hma98@mail.aub.edu, mmansour@aub.edu.lb).}
}

%
%

\markboth{{Efficient Attitude Estimators: A Tutorial and Survey}}%
{Shell \MakeLowercase{\textit{et al.}}: Bare Demo of IEEEtran.cls for IEEE Journals}
%



\maketitle

\begin{abstract}
Inertial sensors based on micro-electro-mechanical systems (MEMS) technology, such as accelerometers and angular rate sensors, are cost-effective solutions used in inertial navigation systems in a broad spectrum of applications that estimate position, velocity and orientation of a system with respect to an inertial reference frame. Although they present several advantages in terms of cost and form factor, they are prone to various disturbances such as noise, biases, and random walk that degrade their orientation estimation. The task of an orientation filter is to compute an optimal solution for the attitude state, consisting of roll, pitch and yaw, through the fusion of angular rate, accelerometer, and magnetometer measurements, regardless of the underlying environmental constraints. The aim of this paper is threefold: first, it serves researchers and practitioners in the signal processing community seeking the most appropriate attitude estimators that fulfills their needs, shedding light on the drawbacks and the advantages of a wide variety of designs. Second, it serves as a survey and tutorial for existing estimator designs in the literature, assessing their design aspects and components, and dissecting their hidden details for the benefit of researchers. Third, a comprehensive list of algorithms is discussed for a fully functional inertial navigation system, starting from the navigation equations and ending with the filter equations, keeping in mind their suitability for power-limited embedded processors. The source code of all algorithms is published, with the aim of it being an out-of-box solution for researchers in the field. The reader will take away the following concepts from this article: understand the key concepts of an inertial navigation system; be able to implement and test a complete stand alone solution; be able to evaluate and understand different algorithms; understand the trade-offs between different filter architectures and techniques; and understand efficient embedded processing techniques, trends and opportunities. 
\end{abstract}

\begin{IEEEkeywords}
Coning and sculling compensation, embedded processors, Kalman filter, measurement model, navigation equations, quaternions, Rotation matrix, system-error dynamics, tilt errors
\end{IEEEkeywords}

%
\IEEEpeerreviewmaketitle

\section{Introduction}\label{intro}
%
%
%
%
\IEEEPARstart{A}{n} integrated navigation system exploits the complementary characteristics of different navigation sensors, such as gyroscopes, accelerometers, magnetometers, and global navigation satellite systems GNSS, to increase the precision of the navigation solution. For example, it can compute a high quality pose estimate of a vehicle's position and orientation (up to an accuracy of 0.00l degrees per roll, pitch and yaw axis~\cite{winkler2008high}). Specifically, the underlying attitude (orientation) estimation problem is common to a wide area of applications, ranging from unmanned aerial vehicles (UAVs), virtual reality applications, underwater submersible systems, robots and ground vehicles, to medical instruments and surveying equipment. In addition, recent advances in MEMS have led to a very wide range of low-cost, light-weight and accurate components that increase the reliability of the navigation solution significantly. Nowadays, the navigation technique common to almost all integrated navigation systems is the \emph{strapdown} inertial navigation technique. In a strapdown inertial navigation system (INS), an inertial measurement unit (IMU), mounted to the vehicle, senses accelerations and angular rates for all six degrees of freedom of the vehicle. From this data, a strapdown algorithm (SDA) can compute a navigation solution, assuming that initial position, velocity and attitude are known~\cite{friedland1978analysis}. INS-SDAs must be reliable and computationally-efficient in order to suit low-end applications with power-limited processing capabilities~\cite{huddle1989advances}.
\begin{figure}[t]
    \centering
    \includegraphics[width=1\linewidth]{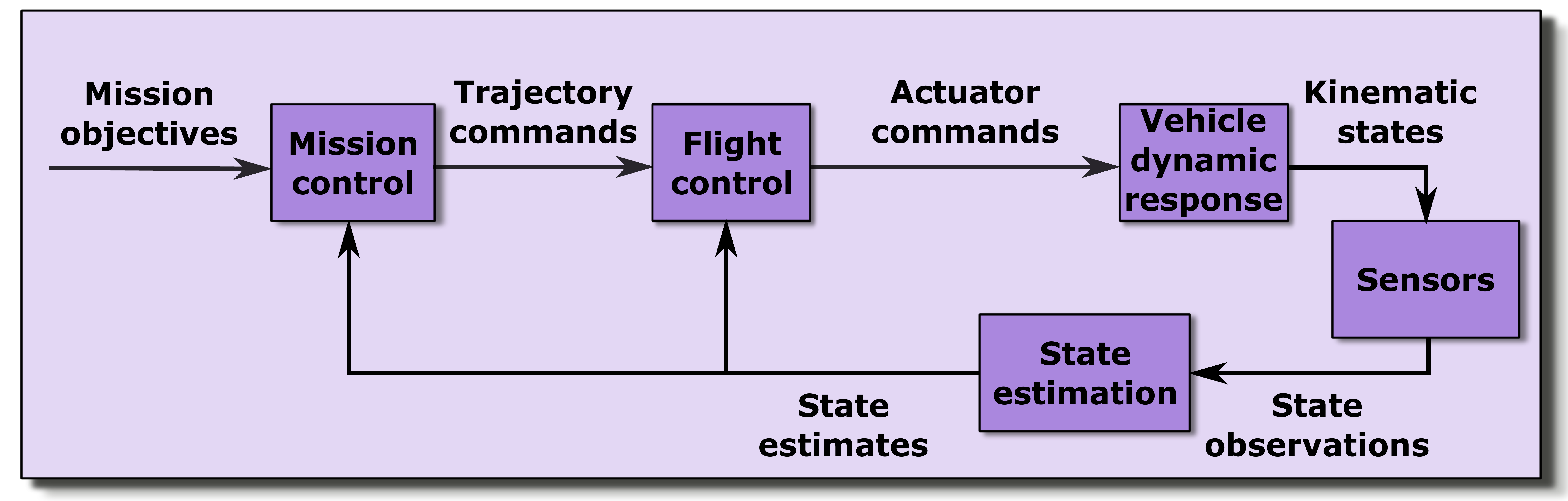}
    \caption{Block diagram illustrating the basic elements in controlling a robot.}
\label{fig_uas}
\end{figure}

The brain of a typical robot vehicle is called a navigation computer or controller~\cite{kealy2011evaluating}. The controller uses on-board sensors to estimate its current position and orientation. Figure~\ref{fig_uas} shows a simplified block diagram of a generic autonomous vehicle computer, including a mission controller, a state estimator and a command controller. State estimation is implemented by fusing the raw measurements from a set of state-observing sensors, and forming an estimate of the vehicle's state (position and attitude)~\cite{barton2012fundamentals}. The vehicle controller algorithms automatically manipulate the actuators on-board the vehicle to achieve a set of trajectory commands using the system states (position and orientation) as feedback. The trajectory commands are generated by the mission controller, which could be a human operator or a set of algorithms that convert mission objectives into trajectory commands~\cite{li2004evolution}. 

This paper aims at providing an overview of various attitude estimation techniques, building up the knowledge of the reader from basic principles, as well as providing  insight in an intuitive manner for concepts hidden between the lines of sophisticated equations governing the system as a whole. The rest of the article is organized as follows:
\begin{itemize}[label={\tiny\raisebox{1ex}{\textbullet}}, leftmargin=*]
\item Section~\ref{background} provides background on the context of why attitude estimation filters are important.
\item Section~\ref{components} gives a basic overview of attitude  estimators currently in use, their components and applications.
\item Section~\ref{coordinates} describes various coordinate systems used in inertial navigation systems and defines the transformation of coordinates from one frame to another. Of primary concern is their relative orientation.
\item Section~\ref{applied} discusses and elaborates on applied inertial navigation algorithms, and presents efficient pseudo-codes of various INS algorithms.
\item Section~\ref{principles} develops the basic INS error equations and gives insights on deriving a simplified system model of the same error equations without delving deeply into rigorous mathematical proofs. This aims at giving the reader intuition into one of the most basic components of an inertial navigation system. 
\item Section~\ref{Kalman} discusses the components and the computational aspects of Kalman filters. Also efficient algorithms targeted for embedded processors are presented. 
\item Section~\ref{insights} is dedicated to insights of the Kalman filter implementations. It points the  readers' attention to some of the practical aspects to be considered when designing an attitude estimation filter.
\item Section~\ref{other} presents different approaches to solving the attitude estimation problem, which are very efficient in terms of computational load and power consumption.
\end{itemize}

\section{Background on Filters}\label{background}
In this section, we provide a background introduction of the position of attitude estimation filters in the context of inertial navigation systems in general.   We also demonstrate the development phases, and provide a brief description of the major mile-stones in its history.

\subsection{Attitude Estimation Filters and Inertial Systems}
\begin{figure}[hbtp]
    \centering
    \includegraphics[width=0.9\linewidth]{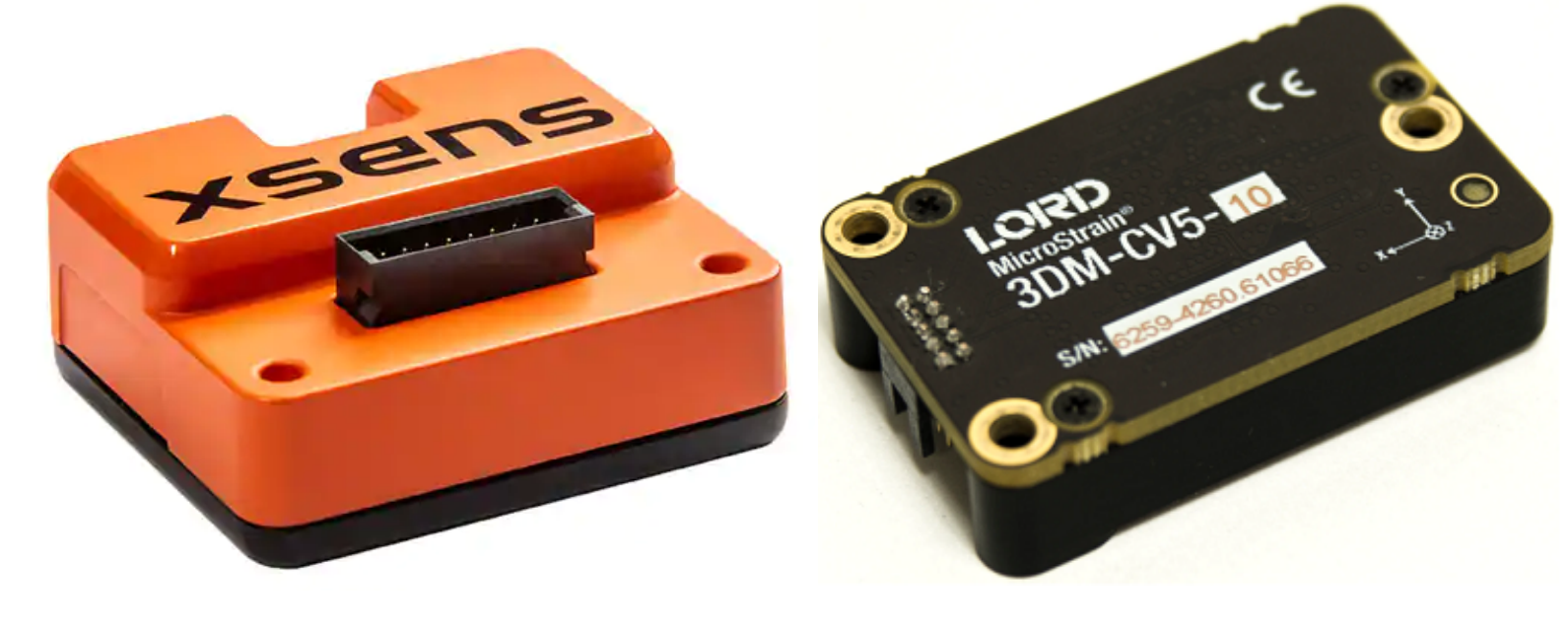}
    \caption{Commercial attitude and heading reference systems with built-in IMUs from XSens Technologies BV  and LORD MicroStrain, with three accelerometers, three gyroscopes, three magnetometers and three temperature sensors for sensor calibration. These IMUs include coning and sculling compensation (see Section~\ref{applied}), which enables them to deliver attitude data at low rates without loosing accuracy.}
    \label{fig_imu}
\end{figure}

The heart of any inertial navigation system is a fully calibrated and embedded inertial measurement unit (IMU) like the 3DM-CV5 from LORD MicroStrain or the series of IMUs from XSens Technologies BV (Fig.~\ref{fig_imu}.). IMUs should deliver accurate temperature-compensated (see Section~\ref{insights} for sensor calibration and compensation) inertial sensor data from three gyroscopes and three accelerometers to a navigation floating-point digital signal processor (DSP)~\cite{nguyen2012loosely,wang2006quadratic}. 

\begin{figure}[hbtp]
    \centering
    \includegraphics[width=0.5\linewidth]{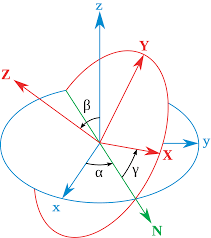}
    \caption{Euler Angles ($\alpha$, $\beta$, and $\gamma$).}
    \label{fig_eul}
\end{figure}

\begin{figure}[hbtp]
    \centering
    \includegraphics[width=1\linewidth]{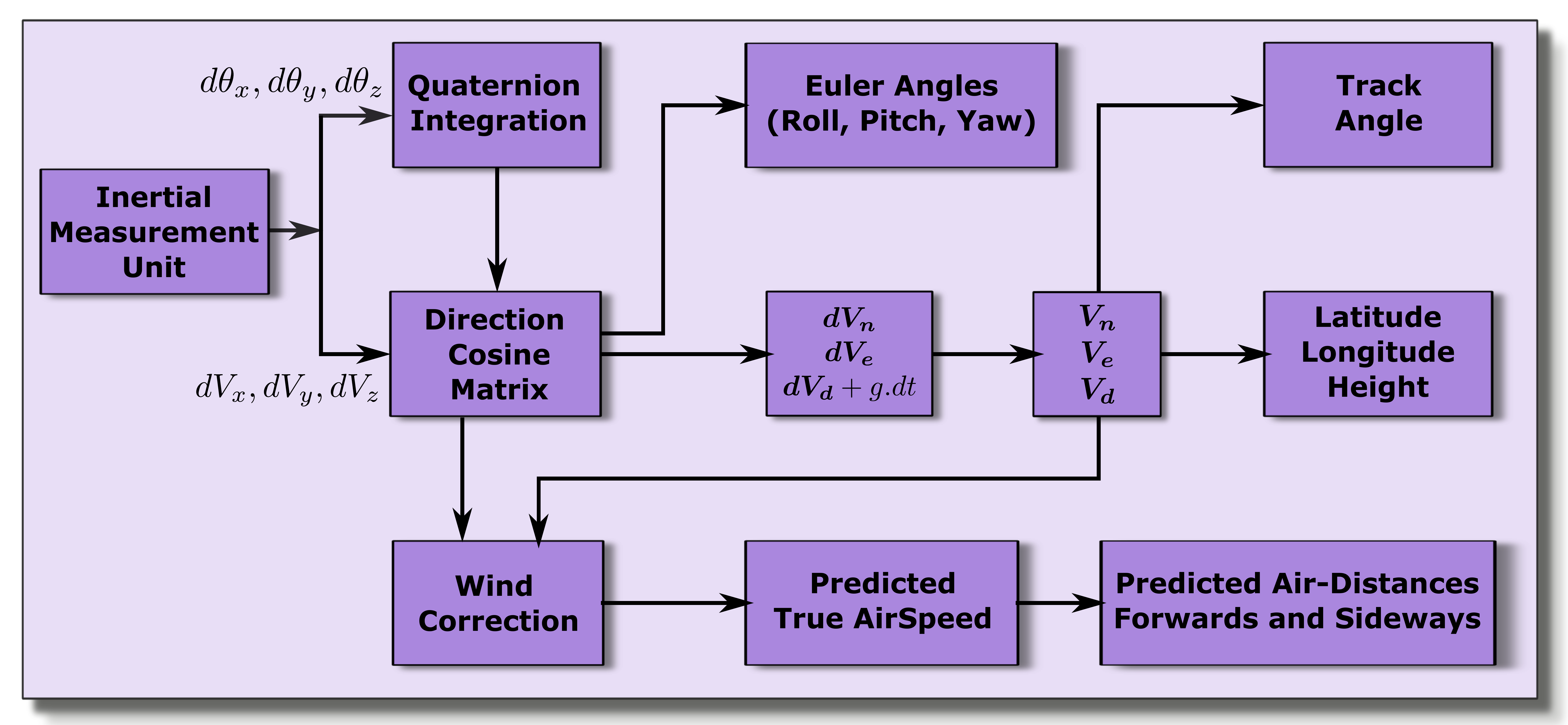}
    \caption{The inertial measurement unit is the basic element in any inertial navigation system. Raw data acquired from the IMU in  the form of delta velocities and delta angles are integrated and converted to the navigation frame. When in the navigation frame, delta velocities can be added to estimated wind speeds to predict true airspeed and thus predict air-distances. Air distances are essential in estimating the amount of energy consumption (fuel or battery) for any flying vehicle.}
    \label{fig_ins}
\end{figure}

Small angular increments ($d\phi $\footnote{Small angular increments, also called Delta theta terms $d\phi$, are small measured angular changes over one IMU cycle.\label{dtheta}}) terms obtained from   gyroscope sensors  are compensated for fine gyro-bias\footnote{Gyro bias is the mean angular change per second measured by the gyro when the actual angular rate is zero (stationary case).} corrections and then integrated using fast quaternion algorithms (Section~\ref{applied}) to derive a $3\times3$ ($9$ element) direction cosine matrix (Fig.~\ref{fig_ins}), which defines the instantaneous orientation of the vehicle relative to the local level earth-centered coordinates (\textit{North, East, Down})~\cite{wang2018gnss,jiang2016novel}. Sensed velocity increments ($dV $\footnote{Small velocity increments delivered by the accelerometer sensors during a  time step $dT$, also called delta velocity terms.}) obtained from  accelerometers are then transformed into delta velocity incremental components (with the aid of the direction cosine matrix) in the local-level earth-referenced (see Fig.~\ref{fig_eul}) coordinate frame. 

Since the IMU also senses gravity, the delta velocities contain components of integrated gravity~\cite{wendel2001direct}. To compensate for this  gravity component in the down direction, we subtract it from the delta-velocity components and then integrate to give velocity components in the local-earth referenced coordinate frame ($V_n, V_e$ and $V_d$), which are subsequently integrated further to produce updated values of latitude, longitude and altitude. Using the direction cosine matrix, heading, roll and pitch values are computed~\cite{martin2010design}.

Usually a high speed Kalman filter propagated and updated at a predefined computation rate is used to estimate, align, and correct system computed states and residual fine inertial sensor bias values. This is done using measurement aiding  from multiple  sources including, GPS-receiver's position and velocity, barometric pressure sensors and magnetic heading. Velocity measurements is required to enable the system to maintain a mathematical representation of horizontal~\cite{werries2016adaptive}. When the vehicle is moving, and especially when velocity is changing, you have no means of separating sensed gravitational acceleration from true acceleration, and a slight mathematical misalignment in your horizontal model will result in erroneous measurements of acceleration (see Section~\ref{principles}) since you will be sensing a component of gravity. If you have a velocity reference (from airspeed sensors, wheel-encoders or GPS, etc.) you can correct the misalignment, maintain a true representation of horizontal, and integrate acceleration and velocity correctly~\cite{shaghaghian2019improving}. The way one maintains alignment in modern day navigation systems is to use a Kalman filter~\cite{pham201515}.

\subsection{Development History}
\begin{figure}[t]
    \centering
      \begin{subfigure}{0.8\linewidth}
    \centering
    \includegraphics[width=\linewidth]{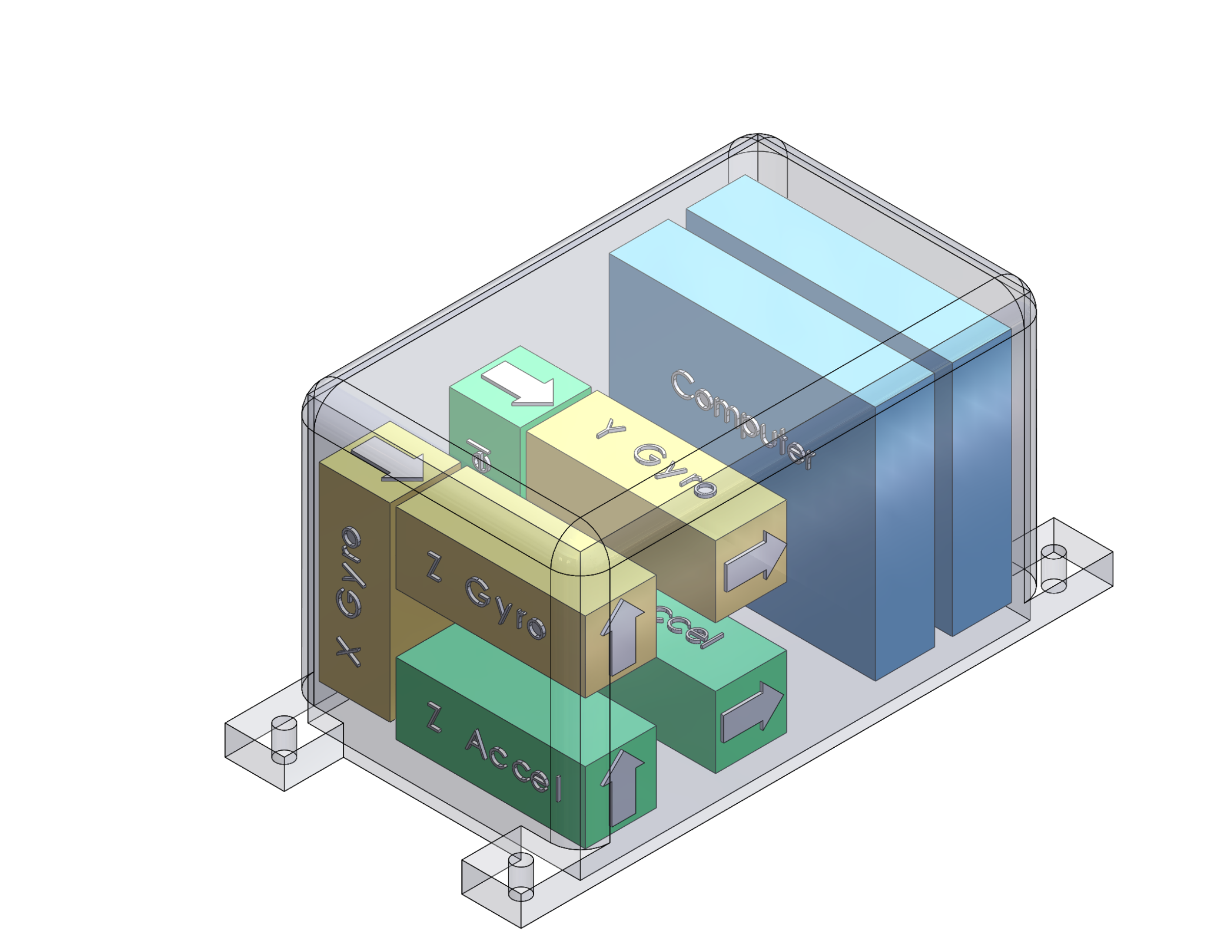}
    \caption{IMU cluster (strapdown system)}
    \label{fig_cluster}
    \end{subfigure}
    \begin{subfigure}{0.8\linewidth}
    \centering
    \includegraphics[width=\linewidth]{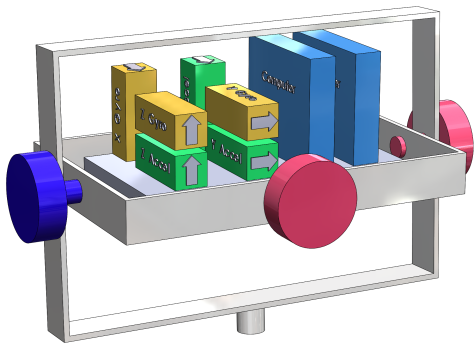}
    \caption{INS gimballed platform}
    \label{fig_gimbal}
    \end{subfigure}
 \caption{The strapdown system replaces gimbals with a computer that simulates their presence electronically. In the strapdown system, the gyroscopes and accelerometers are rigidly mounted to the vehicle structure so that they move with the vehicle. In a three axis gimballed platform, the gyros alone will try to maintain the platform aligned in inertial space. If the platform is operating in local-level coordinates, the navigation computer must keep the platform horizontal. It does this by sending command signals to the gyros that otherwise would fight the gimbals motion.}
\label{fig_mul}
\end{figure}

The fundamental principles (laws of mechanics and gravitation) on which inertial navigation is based was discovered by Isaac Newton in the seventeenth century~\cite{titterton2004strapdown}. Despite of this, it was about another two centuries before inertial navigation techniques could be demonstrated. A brief chronology of the history of inertial navigation systems is given as follows:
 \begin{itemize}[label={\small\raisebox{0.2ex}{\textbullet}}]
 \item \textbf{1852} - The gyroscopic effect was discovered by Foucault who was the first to use this word.
 \item \textbf{1923} - Schuler invents a device that enables a vertical reference to be defined~\cite{schuler1923storung}.
 \item \textbf{1920} - Directional gyroscopes and artificial horizon instruments were produced for aircrafts.
 \item \textbf{1930} Boykow introduced the idea of using accelerometers and gyroscopes to build a functional inertial navigation system.
 \item \textbf{1949} - The first publication suggesting the strapdown inertial navigation concepts. 
 \item \textbf{1950's} - The accuracy of gyroscopes increases incredibly, reducing their errors from $\unitfrac[15^{\circ}]{\!}{hour}$ to about $\unitfrac[0.01^{\circ}]{\!}{hour}$.
 \item \textbf{1960's} - The start of the ring laser gyroscope and wide spread of the so-called stable platform technology.
 \item \textbf{1961} - NASA awarded MIT laboratory (later to become the Charles Stark Draper Laboratory), a contract for preliminary design study of a guidance and navigation system for Apollo~\cite{battin1982space}\cite{draper1965guidance}.
 \item \textbf{1970's} - Advances in technology converged to make strapdown systems available~\cite{tazartes2014historical}.
  \item \textbf{1980's} - Developments of higher-order gravity models enabled trajectory accuracy improvements of approximately an order of magnitude~\cite{reams1983effects}\cite{ford1983gravitational}
  \item \textbf{1990's} -A non-gyroscopic inertial measurement unit was proposed that consisted of a triad of
accelerometers mounted on three orthogonal platforms rotating at constant angular velocities~\cite{merhav1982nongyroscopic}.
 \end{itemize}
Nearly all IMUs fall into one of the two categories; stable platform systems Fig.~\ref{fig_gimbal}
or, strapdown systems, Fig.~\ref{fig_cluster}. The difference between the two
categories is the frame of reference in which the rate-gyroscopes and accelerometers operate.

\begin{figure*}[hbtp]
    \centering
    \includegraphics[scale=0.165]{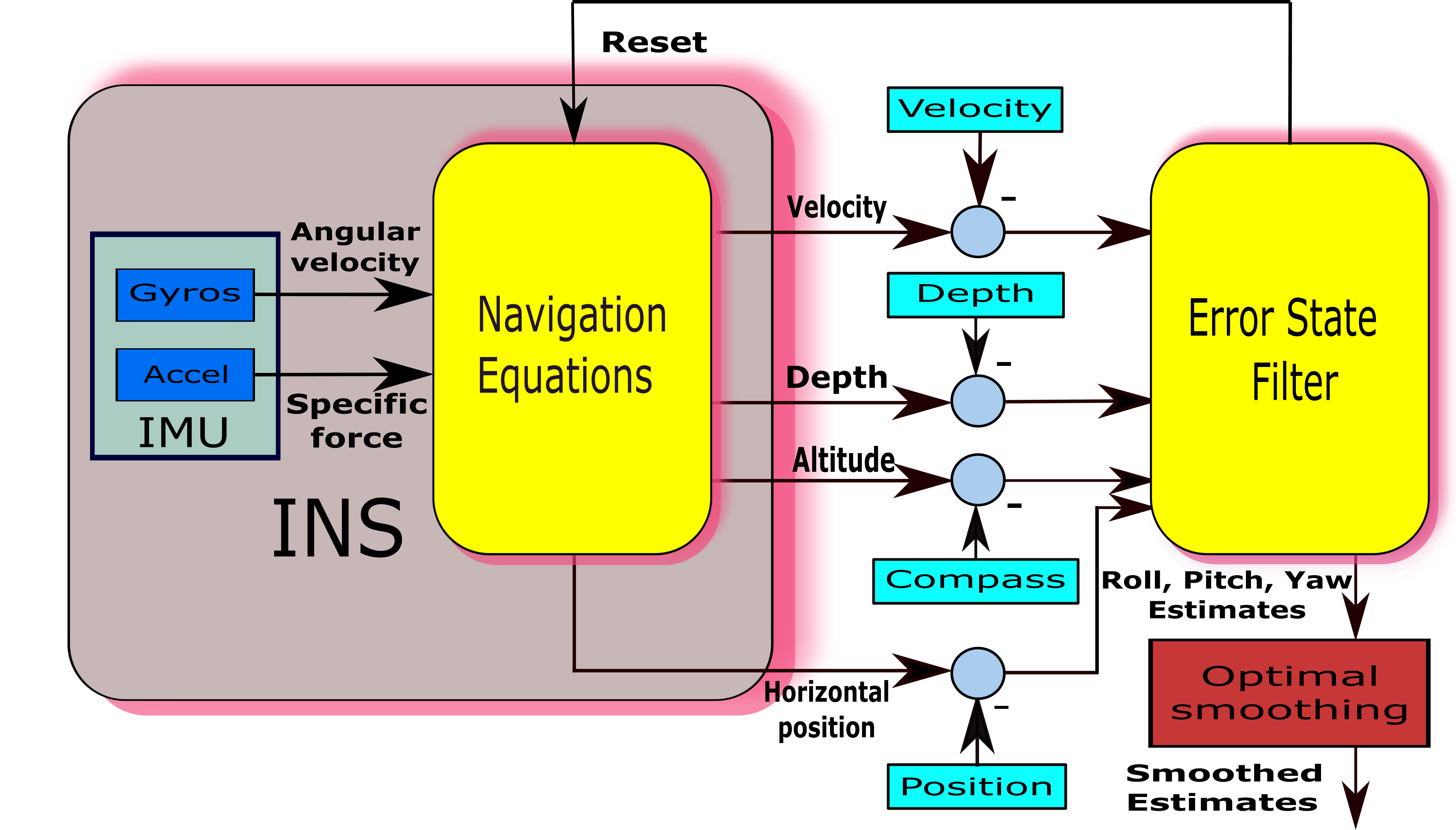}
    \caption{Any GNSS-INS system is composed of a) INS  responsible of predicting velocity, position and heading, b) State fusion filter that combines the readings from both INS and GNSS to optimally estimate attitude of the system.}
    \label{fig_nav_alg}
\end{figure*}

\subsubsection{Stable Platform Systems}
In stable platform system types, the inertial sensors are mounted on a platform which is isolated from
any external rotational motion. In other words the platform is held in alignment with the global frame.
This is achieved by mounting the platform using gimbals (frames) which allow the platform freedom in
all three axes. The platform mounted gyroscopes detect any platform rotations.
These signals are fed back to torque motors which rotate the gimbals in order to cancel out such rotations,
hence keeping the platform aligned with the global frame.
To track the orientation of the device the angles between adjacent gimbals can be read using angle
pick-offs. To calculate the position of the device the signals from the platform mounted accelerometers
are double integrated. Note that it is necessary to subtract acceleration due to gravity from the vertical
channel before performing the integration.

%
\subsubsection{Strapdown Systems}
In strapdown systems, the inertial sensors are mounted rigidly onto the device, and therefore output
quantities are measured in the body frame rather than the global frame. To keep track of orientation the
signals from the rate gyroscopes are ‘integrated’, as described in Section~\ref{applied}. To track position the three
accelerometer signals are resolved into global coordinates using the known orientation, as determined by
the integration of the gyro signals. The global acceleration signals are then integrated as in the stable
platform algorithm.

Stable platform and strapdown systems are both based on the same underlying principles. Strapdown
systems have reduced mechanical complexity and tend to be physically smaller than stable platform
systems. These benefits are achieved at the cost of increased computational complexity. As the cost of
computation has decreased strapdown systems have become the dominant type of INS.
 
More recently, there has been significant developments in inertial sensors, especially gyroscopes with large dynamic range giving the strapdown principles opportunity to be realized. This has enabled the complexity and size of inertial navigation systems to be reduced, as well as enabling reliable~\cite{vagner2011mems} inertial sensors to be produced at a relatively inexpensive price, which led to significant advancements in a diversity of applications ( see Section~\ref{components}).

%
\section{Aided Inertial Navigation Systems}\label{components} 
The inspiration of any system integration concept, is to get superior execution than would be conceivable by any of the stand-alone systems. This section starts with a look at the qualities of GNSS and INS systems that make them so appropriate to combine together ~\cite{gade2008introduction}. The  subtleties  for combining the systems together will be examined afterward within this section.

One of the imperative points of interest of inertial (gyroscopes and accelerometers) systems is that they require no interaction with the environment past the client. This is attractive  particularly to clients where outside supporting cannot be depended upon or is rare. In another sense, no external interference besides the client is needed for the INS to work properly. On the contrary, GNSS systems, which depend on signals transmitted from satellites, obviously, can't be ensured in all cases, and transitory blackouts going from seconds to minutes might be conceivable, contingent upon the application and working environment.

The only restriction concerning the output rate of the inertial system, is the computational power of the INS host computer. Some INS's are able of delivering the navigation state vector at $\unit[100]{Hz}$ or more. On the other hand, most GNSS receivers have data rates of $1$ to $\unit[20]{Hz}$, in spite of the fact that a few specialized receivers can give yield up to $\unit[100]{Hz}$. Expressed in an another way, the bandwidth of the navigation states delivered by inertial system is regularly much higher than with GNSS, which is vital in guidance and control and for high-dynamic applications.

GNSS and INS are complementary in terms of the information they provide. In particular, despite the fact that GNSS can provide an attitude solution, this is usually dodged in practice because it includes employing multiple receiver antennas and expensive equipment, while attitude is the main output of INS algorithms.

Most critically, GNSS and INS systems are also complementary in terms of their errors. While low-cost INS inertial sensors are error unbounded, GNSS provides velocity and position estimates that are limited and bounded in terms of their errors. Also, GNSS systems are dominated by high-frequency errors while INS systems are susceptible to low-frequency errors due to the integration (effectively a low-pass filter) of the mechanization equations (see Section~\ref{principles}).

With respect to what has been mentioned, whenever GNSS and INS systems are fused, the GNSS can deliver high-fidelity position and velocity measurements that can bound the INS system generated errors, which in turn delivers high frequency navigation states (attitude, position and velocity) needed for guidance and control of vehicles. The INS system can also maintain good accuracy in case of outages of GNSS during temporary blockage of receiver antennas. These are the main reasons that motivate the integration of both systems nowadays. 

In this article, we will demonstrate to the reader how both systems (INS and GNSS) can be fused together for an optimal navigation solution. It is helpful to always remember that the navigation solution is obtained by integrating the acceleration readings to obtain  velocity and by double integrating the sensed accelerometers to obtain position.

%
\subsection{Applications of Inertial Navigation Systems}
Inertial navigation systems are used extensively in every day applications, covering aircraft navigation, spacecrafts, robots, unmanned aerospace vehicles and ships. As well, many  novel applications include, active suspension of high performance racing cars, Stewart platform simulators and surveying of underground oil pipelines and wells.
They can also be applied to many advanced medical equipment, such as MRI devices, surgical robots and intelligent beds. The use of inertial navigation systems is widely spreading in the medical field, for example, in the manufacturing of wheel chairs based on inertial systems. They have been placed on head trackers of disabled people where they can choose where to go and in what direction solely by moving their heads.

\begin{figure}[H]
    \centering
    \includegraphics[scale=0.5]{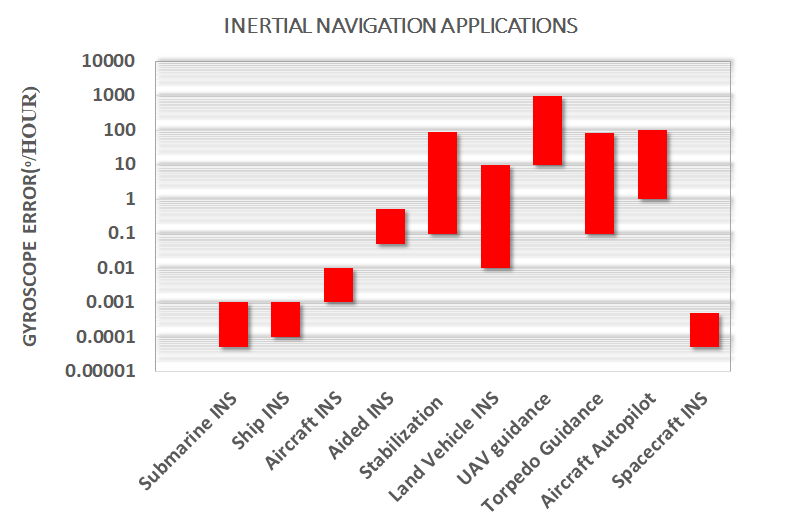}
    \caption{Diverse applications need different accuracies of gyros in strapdown inertial navigation systems.}
    \label{fig_excel}
\end{figure}

Due to such diversity of  fields where inertial systems may be applied, a broad range of inertial sensor accuracy is required (especially for gyroscopes, see Fig.~\ref{fig_excel}).

Also, since inertial systems differ in the amount of time they will be required to report accurate data, it is necessary to choose the sensors accordingly. For example, many airborne systems may need to provide accurate position and attitude data for several hundreds of kilometers or several hours. In this instance, it is necessary to rely on inertial sensors having very low residual gyroscope biases, having the order of $0.001$ degrees per hour. Other cases involving marine or space applications may be required to provide accurate data for weeks or even months. In these extreme cases, gyroscopes having bias errors on the order of $0.0001$ degrees per hour are mandatory. In some cases such as torpedo guidance operating for a few minutes, it is sufficient to rely on sensors with moderate accuracy ($0.1$ to $100$ degrees per hour, see Fig.~\ref{fig_excel}).

\begin{table*}[!bht]
    \centering
\tcbset{enhanced,width=0.9\textwidth,
        fonttitle=\bfseries\large\sffamily,
        fontupper=\normalsize\sffamily,
        colback=yellow!10!white, colframe=red!50!black,
        colbacktitle=Salmon!30!white,
        coltitle=black, center title}
\begin{tcolorbox}[tabularx={|>{\bfseries\hsize=0.1\hsize}X| 
                              >{\centering\hsize=0.2\hsize}X| 
                              >{\hsize=0.4\hsize}X| 
                              c|
                              >{\hsize=0.3\hsize}X|} ]  
        \multicolumn{2}{>{\centering\hsize=0.25\hsize}X|}{%
            \textbf{}}
        &   \multicolumn{2}{>{\centering\hsize=0.45\hsize}X|}{%
            \textbf{MEMS Gyro Error Characteristics}}
                &   \hfil\textbf{}          \\  \cline{3-5}
                
      \multicolumn{2}{>{\centering\hsize=0.25\hsize}X|}{%
            \textbf{}}
        &   \hfil \textbf{Description}
        &\hfil\textbf{Units}
                &   \hfil\textbf{Result of Integration}          \\  \hline
\multirow{9}{*}{\Large TYPE} 
    &  {\color{Red}Constant Bias}
    &  The average output from the gyroscope when it is not undergoing any rotation
                        & \color{Magenta}$\unit{\degree/sec}$   &  A steadily growing angular error.     \\  \cline{2-5}
    &  {\color{Red} White Noise \\(Angle Random Walk ARW)}& Very high frequency noise that is added to the signal that has an average amount equal to sigma ($\sigma$) and with a long term average equal to zero.
                        & \color{Magenta} $\unit{\degree /sec/\sqrt{Hz}}$   & To find error in orientation due to gyro white noise multiply ARW by the square root of the integration time (t).
      \\  \cline{2-5}
    &  {\color{Red} Bias Stability \\ (Sometimes called Bias Instability)}& A bias stability measurement describes how the bias of a device may change over a specified period
of time. (Bias Fluctuations)~\cite{board1998ieee}
                        & \color{Magenta} $\unit{\degree/sec}$    &       \\  \cline{2-5}
    &  {\color{Red} Rate Random Walk} & This is a rate error due to white noise in angular
acceleration~\cite{vaccaro2011statistical}
                        & \color{Magenta}$\unit{\degree/sec^{1.5}}$  &    Introduces the opportunity to plan for re-calibration in critical
applications that require extended life.    \\  \hline
    
      \multicolumn{2}{>{\centering\hsize=0.25\hsize}X|}{%
            \textbf{}}
        &   \multicolumn{2}{>{\centering\hsize=0.5\hsize}X|}{%
            \textbf{Accelerometer Error Characteristics}}
                &   \hfil\textbf{}          \\  \cline{3-5}

      \multicolumn{2}{>{\centering\hsize=0.25\hsize}X|}{%
            \textbf{}}
        &   \hfil \textbf{Description}
        &\hfil\textbf{Units}
                &   \hfil\textbf{Result of Integration}          \\  \hline

 \multirow{9}{*}{\Large TYPE} 
    & {\color{Red} Constant Bias}
    &  The average output from the accelerometer when it is not undergoing any movement.
                        & \color{Magenta} $\unit{m/sec^2}$   &  A steadily growing velocity error.     \\  \cline{2-5}
    &  {\color{Red} White Noise \\(Velocity Random Walk VRW)}&  Very high frequency noise that is added to the signal that has an average amount equal to sigma ($\sigma$) and with a long term average equal to zero.
                        &  \color{Magenta} $\unit{m/sec^2/\sqrt{Hz}}$   &       \\  \cline{2-5}
    &  {\color{Red} Bias Stability \\ (Sometimes called Bias Instability)}& A bias stability measurement describes how the bias of a device may change over a specified period
of time
                        &  \color{Magenta} $\unit{m/sec}$   & To find error in velocity due to accelerometer white noise multiply VRW by the square root of the integration time (t).      \\  \cline{2-5}
    &  {\color{Red} Acceleration Random Walk} &  This is an acceleration error due to white noise
in jerk (derivative of acceleration)~\cite{petkov2010stochastic}. 
                        & \color{Magenta} $\unit{m/sec^{1.5}}$  &       \\  \hline 
        \end{tcolorbox}
   \caption{Types of random error noise sources}
    \label{tab:my_label}
\end{table*}

%
\section{Coordinate Frames and Transformations}\label{coordinates} 
The attitude of a vehicle is defined as its orientation with respect to a reference frame. It is substantial to understand the different coordinate frames used in inertial navigation systems and their transformations to grasp its concepts. In this section we will discuss the basic coordinate frames that have three orthogonal unit vectors and that follow the right-hand rule~\cite{leffens1982kalman}.

\subsection{Coordinate Frames}
The measurement sensed by an Inertial Measurement Unit (IMU) are three orthogonal components of the body rotation rates and three accelerations in a coordinate frame, which is not directly related to any coordinate frame. These measurements have to be analytically integrated and transformed through several coordinate frames. It is important therefore that all coordinate frames involved in the transformation of the measurements, and results of integration are well defined before any discussion of an inertial navigation system is presented.

%
\subsubsection{Earth-Centered Inertial (ECI) Frame}
Newton defines the inertial frame as the frame of reference that does not rotate or accelerate. 
\begin{figure}[H]
    \centering
    \includegraphics[width=0.7\linewidth]{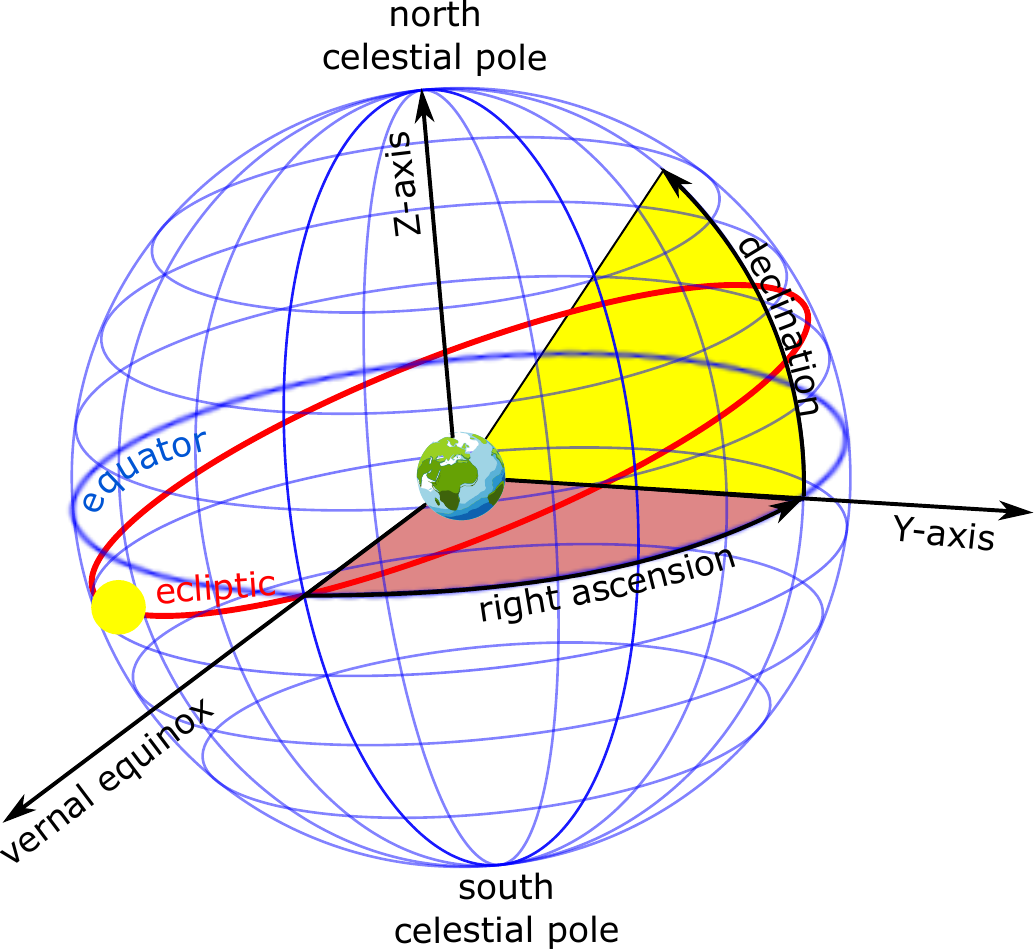}
    \caption{Earth-centered inertial (ECI) frame}
    \label{fig_sph_nav}
\end{figure}

Such a frame is not practically realized although, theoretically well defined. It is best approximated as one that is fixed with respect to the distant stars. For all practical purposes the inertial frame can be treated as the frame that has the following (see Fig.~\ref{fig_sph_nav}):
\begin{itemize}
	\item $x_i$-axis towards the mean vernal equinox.
    \item $y_i$-axis completes a right handed system.
     \item $z_i$-axis towards the north celestial pole.
\end{itemize}

%
\subsubsection{Earth-Centered, Earth-Fixed (ECEF) Frame}
It is a right-handed coordinate system that rotates with and is attached to the earth, which is why it is called earth fixed.
This frame is not inertial since, it revolves  around the sun  at an average orbital speed of $\unitfrac[29.78]{km}{sec}$ and rotates at a rate of $\unitfrac[7.292115 . 10^{-5}]{rad}{sec}$. The Earth-fixed frame can be defined as follows:  
\begin{itemize}
	\item Its origin at the mass center of the earth.
    \item $x_e$-axis pointing towards the Greenwich meridian in the equatorial plane.
     \item $y_e$-axis 90 degrees of Greenwich meridian, in the equatorial plane.
     \item $z_e$-axis is the axis of rotation of the earth and passes through the north pole.
\end{itemize}
In fact, it is important to note that the Global Positioning System (GPS) reports the position and velocity of the satellites in the ECEF coordinates system.\\

%
\subsubsection{Local-Level or Navigation Frame}
It is a non-inertial frame that is commonly used to describe the navigation of a vehicle in a local-level frame. Its axes are aligned along the geodetic directions defined by the earth's surface. Is is defined as follows:
\begin{itemize}
	\item Its origin is at the mass center of the vehicle under study.
    \item The $x_n$-axis points north parallel to the geoid surface.
     \item The $y_e$-axis points east parallel to the geoid surface, along a latitude curve.
     \item The $z_d$-axis points downward, toward the Earth surface, anti-parallel to the surface outward normal $N$.
\end{itemize}

\begin{figure}[H]
    \centering
    \includegraphics[width=0.7\linewidth]{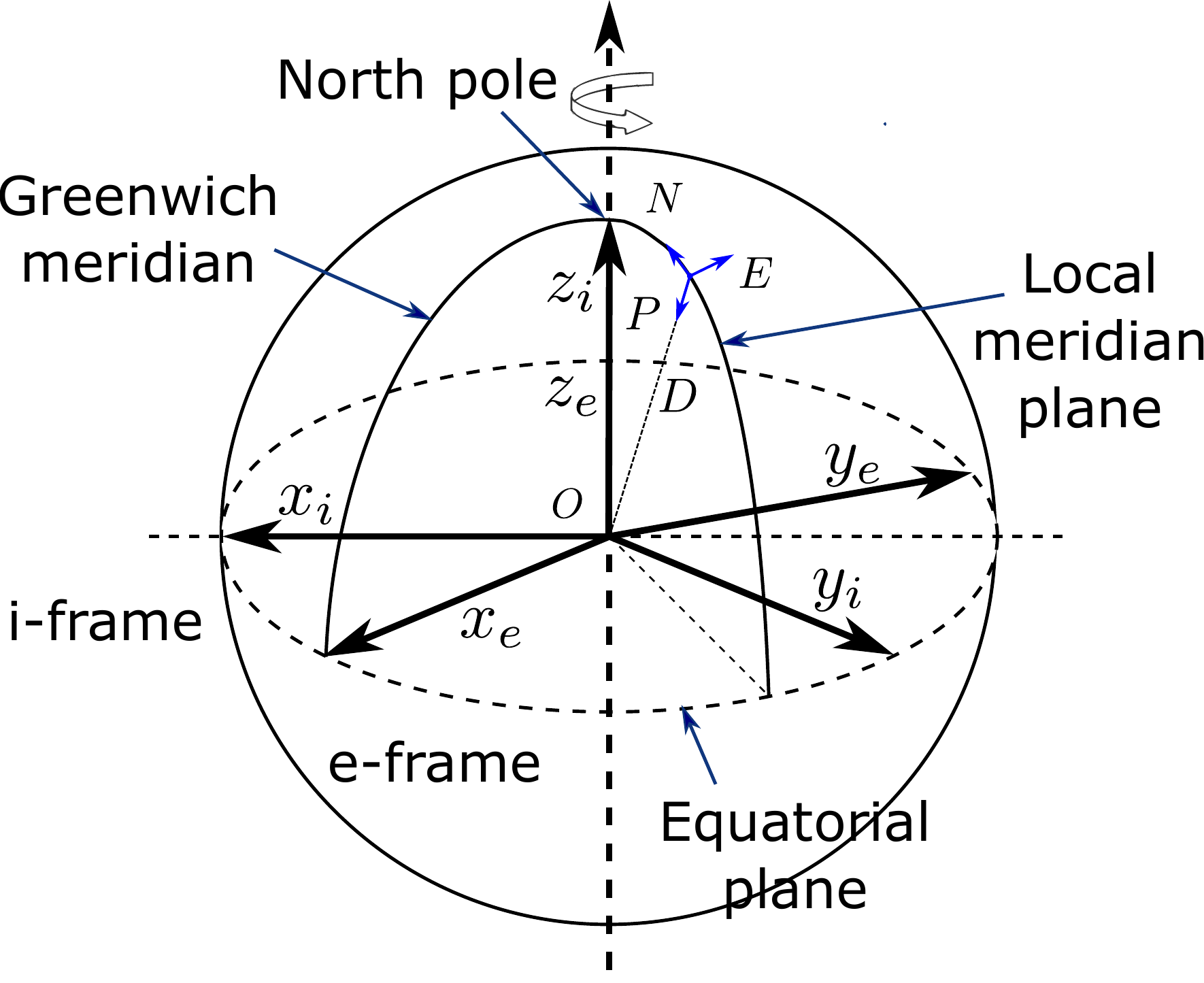}
    \caption{The relative orientation and position of the Earth, inertial frame, and navigation frame.}
    \label{fig_sph_new}
\end{figure}

%
\subsubsection{Body Frame}
\begin{figure}[H]
    \centering
    \includegraphics[width=0.7\linewidth]{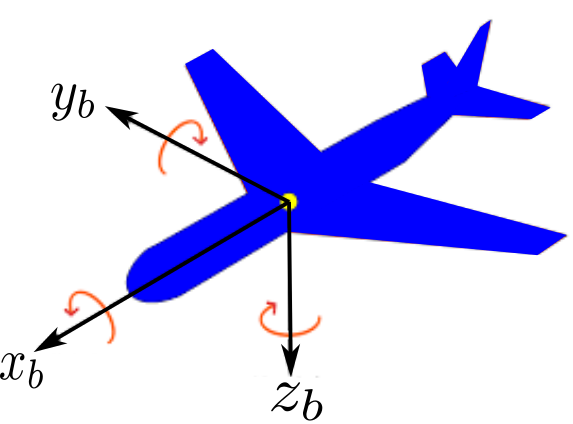}
    \caption{The $x$-axis of the body frame is aligned with the longitudinal axis of the air-frame. The $y$-axis is aligned with the right wing, while the $z$-axis completes the triad.}
    \label{fig_body_frame}
\end{figure}

The body frame is a non-inertial reference frame, in which the measurements of a strapdown inertial navigation system are reported. Its axes are aligned with the output axes of the gyroscopes and accelerometers of the Inertial Measurement Unit (IMU). Thus, the raw data composed of the rotation rates and the accelerations experience by the body are coordinatized along the body axes. It is noted that the navigation frame can be rotated to the body frame by three consecutive right-handed rotations about its three axes (see Fig.~\ref{fig_body_frame}). The definition of the body frame of an inertial navigation system can be summarized a follows:  
\begin{itemize}
	\item Its origin is at the mass center of the inertial navigation system.
    \item The $x_b$-axis points towards the front of the INS.
    \item The $y_b$-axis points towards the right of the INS.
    \item The $z_b$-axis points downwards and perpendicular to the $x$-$y$ plane.
\end{itemize}

%
\subsubsection{Platform Frame}
The platform frame is a virtual frame created mainly for the derivation of the error equations. It is an image of the navigation frame which is recognized on an on-board computer using the outputs from the sensors. Since these sensors are dominated by noise, the platform frame does not coincide with the navigation frame and has a small deviation error from the navigation frame. The definition of the platform frame is as follows:
\begin{itemize}
	\item Its origin at the mass center of the vehicle under study.
    \item  $x_p$-axis  slightly misaligned due to attitude errors with the $x_n$-axis of the navigation frame.
     \item $y_p$-axis slightly misaligned with the $x_e$-axis of the navigation frame and perpendicular to the $x_p$-axis.
     \item $z_p$-axis completes an orthogonal right-handed system.
\end{itemize}

%
\subsubsection{Sensor Frame}
Due to installation errors, the body frame  does not coincide with the sensitivity axes of the sensors ( accelerometers and gyros) used in our inertial navigation system. These errors can be compensated during manufacturing by appropriate calibration. For this reason, we  assume here  that the body frame and the sensor frame coincide (they are interchangeable).

%
\subsection{Transformations}
In this section, the basic mathematical tools that define the transformation between orthogonal coordinate systems are introduced. We will focus mainly on the concepts of \emph{Rotation Matrix}, \emph{Quaternions}, and \emph{Rotation Vectors}.

%
\subsubsection{Rotation Matrix}
A common coordinate transformation in this article is the rotation from the \textit{North-East-Down} coordinate frame to the body \textit{x-y-z} coordinate frame via the ordered Euler angles (see BOX~\ref{box:dir_cos_matrix}) yaw $(\psi)$, pitch $(\theta)$, and roll $(\phi)$. 
\begin{figure}[hbtp]
    \centering
    \includegraphics[width=0.7\linewidth]{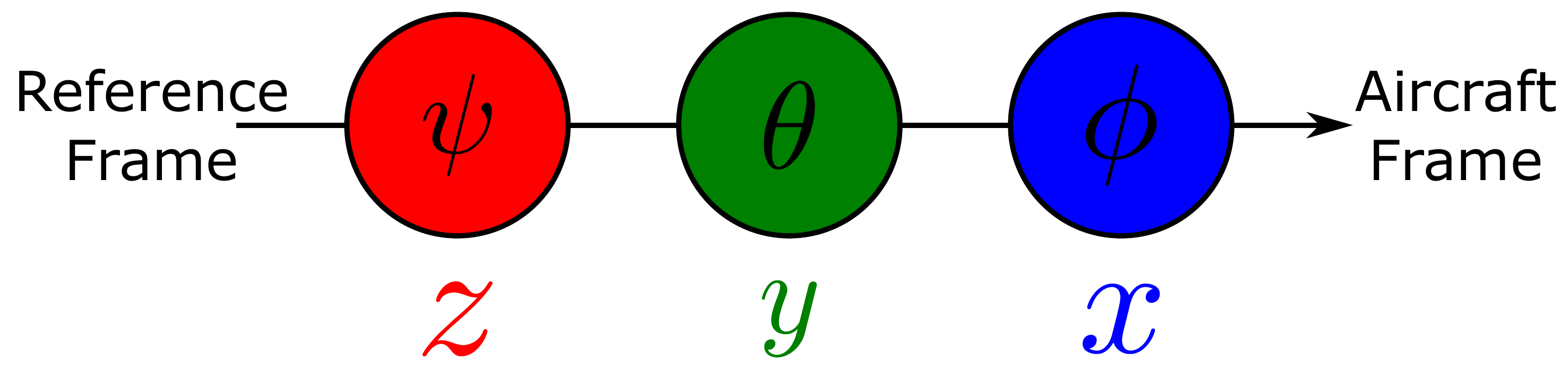}
    \caption{Euler angle sequence corresponding to the three consecutive rotations about the $z$, $y$, and $x$ axes, respectively.}
    \label{fig_euler_angles}
\end{figure}
A sequence of such distinctive rotations is often called a Euler angle sequence of rotations. The restriction stated  above  that successive axes of rotations be distinct still permits at least 12 Euler angle sequences. The sequence \textbf{xzy} means a rotation about the \textbf{x}-axis, followed by a rotation about the new \textbf{z}-axis, followed by a rotation about the newer \textbf{y}-axis.
 
Specifically the first rotation is $\psi$ about \textit{z} which is denoted here as $C_z(\psi)$. The second rotation is $C_y(\theta)$, or $\theta$ about \textit{y}.
Finally, the third rotation is $C_x(\phi)$ or $\phi$ about \textit{x} (see Fig.~\ref{fig_euler_angles}). These three single axis rotations are written as:
\begin{align}
    C_x(\phi) &= \left[ \begin{array} { c c c  }
    1 & 0 & { 0 } \\ 
    0 & \cos\phi & \sin\phi   \\ 
    0  & -\sin\phi & \cos\phi  
    \end{array} \right], 
    \label{eq_Cx_phi}
    \\[1.5em]
    C_y(\theta) &= 
        \left[ \begin{array} { c c c  }
        \cos\theta & 0 & -\sin\theta  
            \\ 0 & 1 & 0  
            \\  \sin\theta  &0 & \cos\theta
            \end{array} \right], 
    \label{eq_Cx_theta}
    \\[1.5em]
    C_z(\psi) &= 
        \left[ \begin{array} { c c c  } 
        \cos\psi & \sin\psi  & { 0 } \\
            -\sin\psi & \cos\psi & 0  \\ 
            0  & 0 & 1 
            \end{array} \right].
    \label{eq_Cx_psi}
\end{align}
Thus, the transformation from body \textit{x-y-z} coordinate frame coordinate frame to the \textit{n}-frame (North-East-Down)  is written as a cascade of the three single-axis rotations above, which can be solved using standard matrix multiplication:

\begin{twocolequfloat}%
\begin{align}
C_{n}^b=C_z(\psi)C_y(\theta)C_x(\phi)=
\begin{bmatrix}
\cos\psi\cos\theta & \cos\psi\sin\theta \sin\phi +\sin\theta \sin\phi& \cos\psi \sin\theta \sin\phi +\sin\theta \sin\phi\\
-\sin\psi \cos\theta &-\sin\psi \sin\theta \sin\phi +\cos\theta \cos\phi&\sin\psi \sin\theta \cos\phi +\cos\theta \sin\phi \\ 
\sin\theta  &-\cos\theta \sin\phi &\cos\theta \cos\phi 
\end{bmatrix}\,.\label{eq_CDCM}
\end{align} 
\end{twocolequfloat}%

%

\newtcolorbox[use counter=myboxcounter, number format=\Alph]{mybox}[2][]{%
colback=yellow!10!white, colframe=blue!75!black,colbacktitle=red!80!black, fonttitle=\bfseries, title= Box \thetcbcounter: #2,#1}

\begin{mybox}[label={box:dir_cos_matrix}]{Direction Cosine Matrix}
The Euler angles are three angles introduced by Leonhard Euler to describe the orientation of a rigid body with respect to a fixed coordinate system. Leonard Euler (1707-1783) was one of the giants inn mathematics~\cite{kuipers1999quaternions}. Euler stated and proved a theorem that states that:\par

\textit{Any two independent orthonormal coordinate frames can be related by a sequence of rotations (not more than three) about coordinate axes, where no two successive rotations may be about the same axis.}

A sequence  of  such  rotations  is  often  called  a  Euler  angle sequence of rotations. The restriction stated in the above theorem that successive axes of rotations be distinct still permits  at  least  12  Euler  angle  sequences.  The  sequence xzy means  a  rotation  about  the x-axis,  followed  by  a rotation about the new z-axis, followed by a rotation about the newer y-axis.\par
We are specifically interested in the well-known Euler sequence called the Aerospace sequence. This sequence (\textbf{zyx}) is commonly used in aircraft and aerospace applications. For example, a primary flight instrument used in air-crafts, continuously relates the orientation of the aircraft to the local-level \textit{n}-frame mentioned above.\par
\begin{figure}[H]
    \centering
    \includegraphics[width=0.71\linewidth]{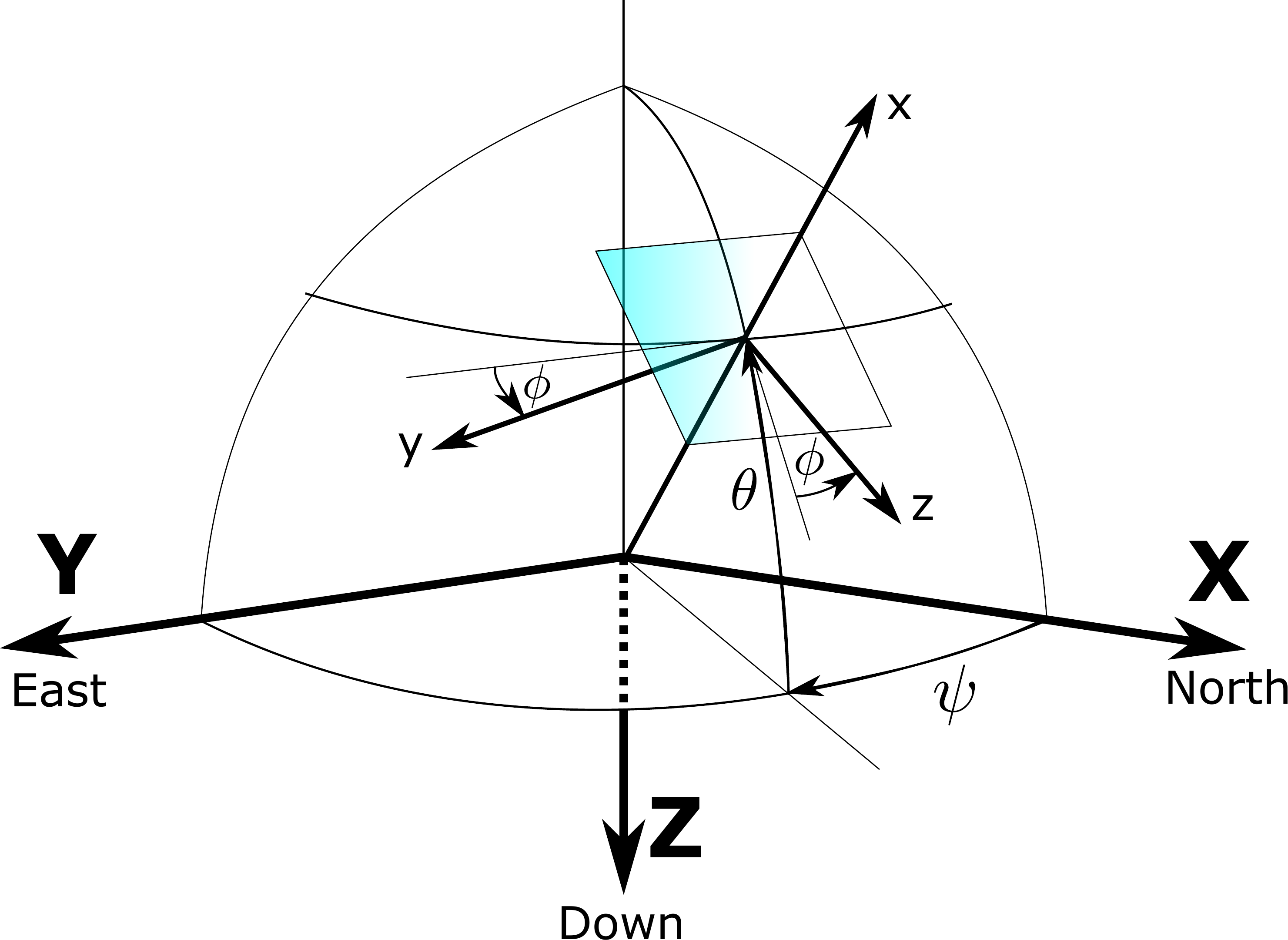}
    \caption{The Euler angles of a vehicle when aligned with the \textbf{x}-axis direction.}
    \label{fig_euler}
\end{figure}
From the \textit{n}-frame, first a rotation through the angle $\psi$ about the \textbf{z}-axis defines the aircraft heading. This is followed by a rotation about the new \textbf{y}-axis through an angle $\theta$ which defines the aircraft pitch. Finally, the aircraft roll angle $\phi$, is a rotation about the newest \textbf{x}-axis. These three Euler angle rotations relate the body coordinate frame of the aircraft to the local-level \textit{n}-frame. 
\end{mybox}

%
\newpage
\subsubsection{Quaternions}
The rotation matrix describes the rotation of 3 degrees of freedom with 9 quantities, with redundancy. Euler angles and rotation vectors are compact but with singularity as mentioned before.

Normal complex numbers can describe rotations in a plane. Recall that in order to rotate a two degrees-of-freedom vector represented be  complex number in the plane  by an angle $\theta$, we  multiply by $\mathrm{e}^{i\theta}$. It can be written in the usual form
\begin{equation}
    \mathrm{e}^{i \theta}=\cos \theta  + i \sin \theta.
\end{equation}

A quaternion $\mathbf{q}$ has a real part and three imaginary parts. Usually the real part is written first and the three imaginary parts next, as
\begin{equation}
    \mathbf{q}= q_0 + q_1 i + q_2 j + q_3 k,
\end{equation}
where $i, j, k$ are three imaginary parts of the quaternion. These imaginary parts satisfy the following equations:
\begin{equation}
    \left\{ \begin{array}{l}
         {i^2} = {j^2} = {k^2} = -1   \\
         ij = k, ji = -k \\
         jk = i, kj = -i \\
         ki = j, ik = -j \\
    \end{array}    \right.
\end{equation}
Alternatively, quaternions are often represented using a scalar and a vector as:
\[
\mathbf{q} =  \left[s, \mathbf{v} \right]^\mathrm{T}, \quad s =q_0 \in \mathbb{R}, \quad \mathbf{v} = [q_1, q_2, q_3]^\mathrm{T} \in \mathbb{R}^3.
\]
Here $s$ is the real part of the quaternion and $\mathbf{v}$ is its imaginary part. If the imaginary part of the quaternion is $\mathbf{0}$ it is called a \textbf{real quaternion} and if the real part is $0$ it is called \textbf{imaginary quaternion}.\par
$ $\hfill

%
\subsubsection{Rotation Vector}
In fact, a rotation can be described by a \textbf{ rotation vector and a rotation angle}. Thus we can use a vector whose direction is parallel to the axis of rotation and whose magnitude is equal to the angle of rotation.

Let us introduce a rotation vector $\mathbf{\Phi}$, which is directed along the axis of rotation and has a magnitude equal to the rotation angle in radians. The equation of the rotation vector can be defined as
\begin{equation}
    \mathbf{\Phi} = \norm{\mathbf{\Phi}}\mathbf{n}
    = \left[\begin{array}{c}
    \mathbf{\phi}_x\\
    \mathbf{\phi}_y\\
    \mathbf{\phi}_z\\
    \end{array} \right] 
    =\norm {\mathbf{\Phi}}
    \left[\begin{array}{c}
    \mathbf{\cos\alpha}\\
    \mathbf{\cos\beta}\\
    \mathbf{\cos\gamma}\\
    \end{array}
        \right],
\end{equation}
where $\mathbf{n}$ is the unit vector in the direction of the rotation vector.
$\alpha, \beta, \gamma$ are the angles between the rotation vector and the coordinate frame axis. The quaternion elements can be represented through the parameters of the rotation vector $\mathbf{\Phi}$ as
\begin{equation}
        \begin{array}{c}
        \begin{aligned}
         q_0 &= \cos\frac{\mathbf{\phi}}{2} \,,  \\
         
         q_1 &= \sin\frac{\mathbf{\|\Phi\|}}{2}\frac{{\phi}_x}{2}\,, \\
         
         q_2 &= \sin\frac{\mathbf{\|\Phi\|}}{2}\frac{{\phi}_y}{2}\,, \\
         
         q_3 &= \sin\frac{\mathbf{\|\Phi\|}}{2}\frac{{\phi}_z}{2}\,. \\
         \end{aligned}
        \end{array}
        \label{eq_quaternion}
\end{equation}

\begin{mybox}[label={box:nav_frame_rel}]{Navigation Frame Relations}
  
We can find the transformation between the \textit{n}-frame and the \textit{e}-frame by using Euler angles. First we rotate about the east axis by $(\pi/2 +\phi)$, then rotate about the new z-axis by the angle $-\lambda$ (see Fig.~\ref{fig_cne}:
\begin{equation}
    C_n^e=C_z(-\lambda)C_y(\pi/2 +\phi)
\end{equation}
 The result is 
 \begin{equation}
 C_n^e=\begin{bmatrix}-\sin\phi\cos\lambda &\sin\lambda &-\cos\phi\cos\lambda\\
 -\sin\phi\sin\lambda&\cos\lambda&-\cos\phi\sin\lambda\\
 \cos\phi & 0& -\sin\phi
 \end{bmatrix}
\end{equation}
\begin{figure}[H]
    \centering
    \includegraphics[width=0.8\linewidth]{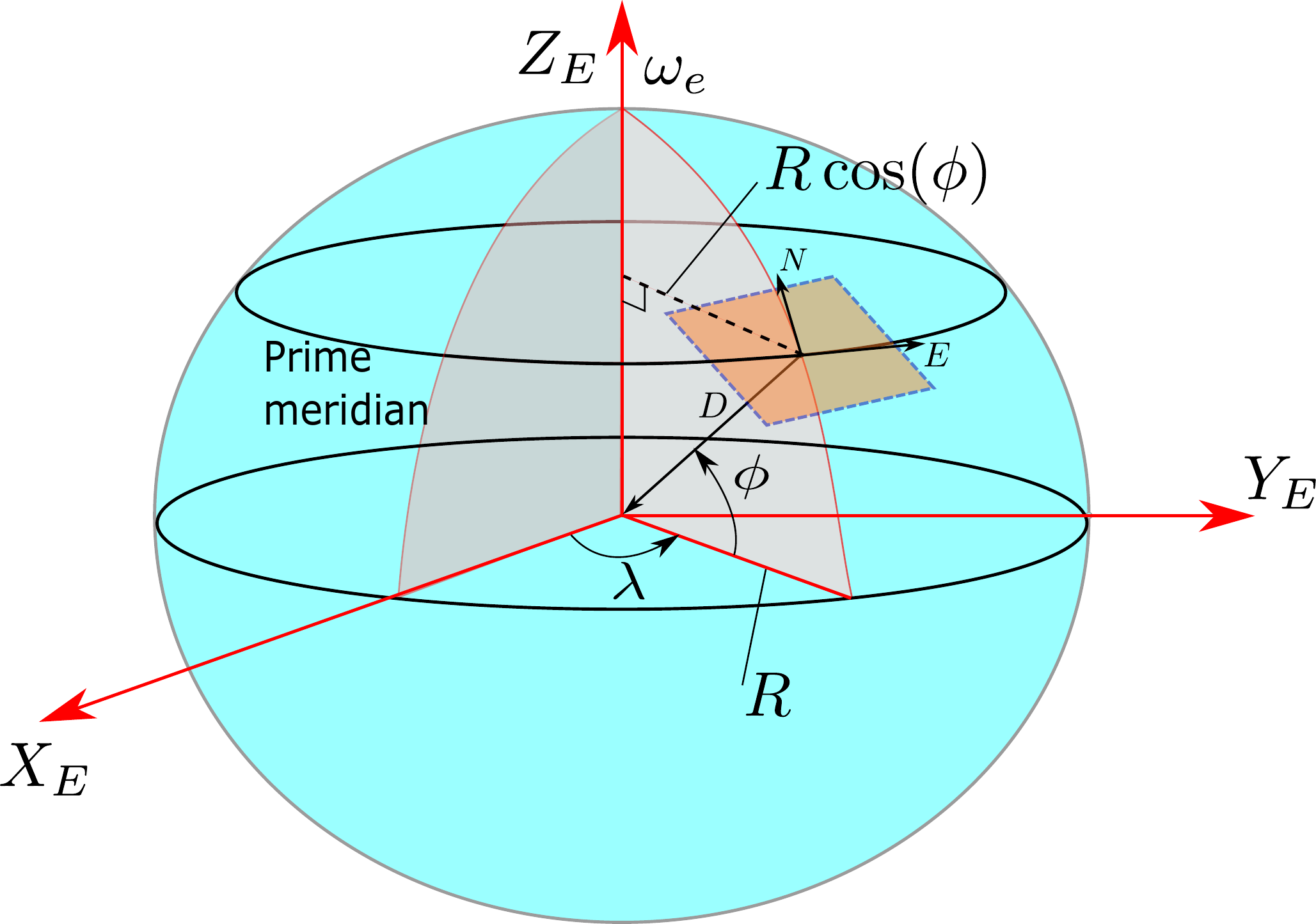}
    \caption{The figure shows the geometrical relationship between the navigation frame and the Earth frame.}
    \label{fig_cne}
\end{figure}
Looking closely at Fig.~\ref{fig_cne} we can find the equation for the angular velocity of the \textit{n}-frame with respect to the \textit{e}-frame coordinatized in the \textit{n}-frame, $\omega_{en}^n$. Clearly, moving along the north direction is accompanied with a mandatory rotation rate, $\dot\phi$,  of the \textit{n}-frame around the east axis, to keep it level. Also, any motion in the east direction is accompanied by a rotation rate, $\dot\lambda$, of the \textit{n}-frame around an axis parallel to the $Z_E$ direction, Since the $Z_E$ direction makes an angle $\phi$ with the north axis, its components along the north and down axes are respectively, $\cos\phi$ and $-\sin\phi$. Therefore:
\begin{equation}\label{eq_wen}
    \omega_{en}^n=(\dot{\lambda}\cos\phi, -\dot\phi, -\dot{\lambda}\sin\phi)
    \end{equation}
    
    From similar geometric considerations, the angular rate of the n-frame with respect to the i-frame is:
    \begin{equation}\label{eq_win}
    \omega_{in}^n=((\dot{\lambda}+\omega_e)\cos\phi, -\dot\phi, -(\dot{\lambda}+\omega_e)\sin\phi)
    \end{equation}
    where $\omega_e$ is the Earth's rotation rate.
\end{mybox}

\begin{figure}[h]
\centering
\includegraphics[width=0.8\linewidth]{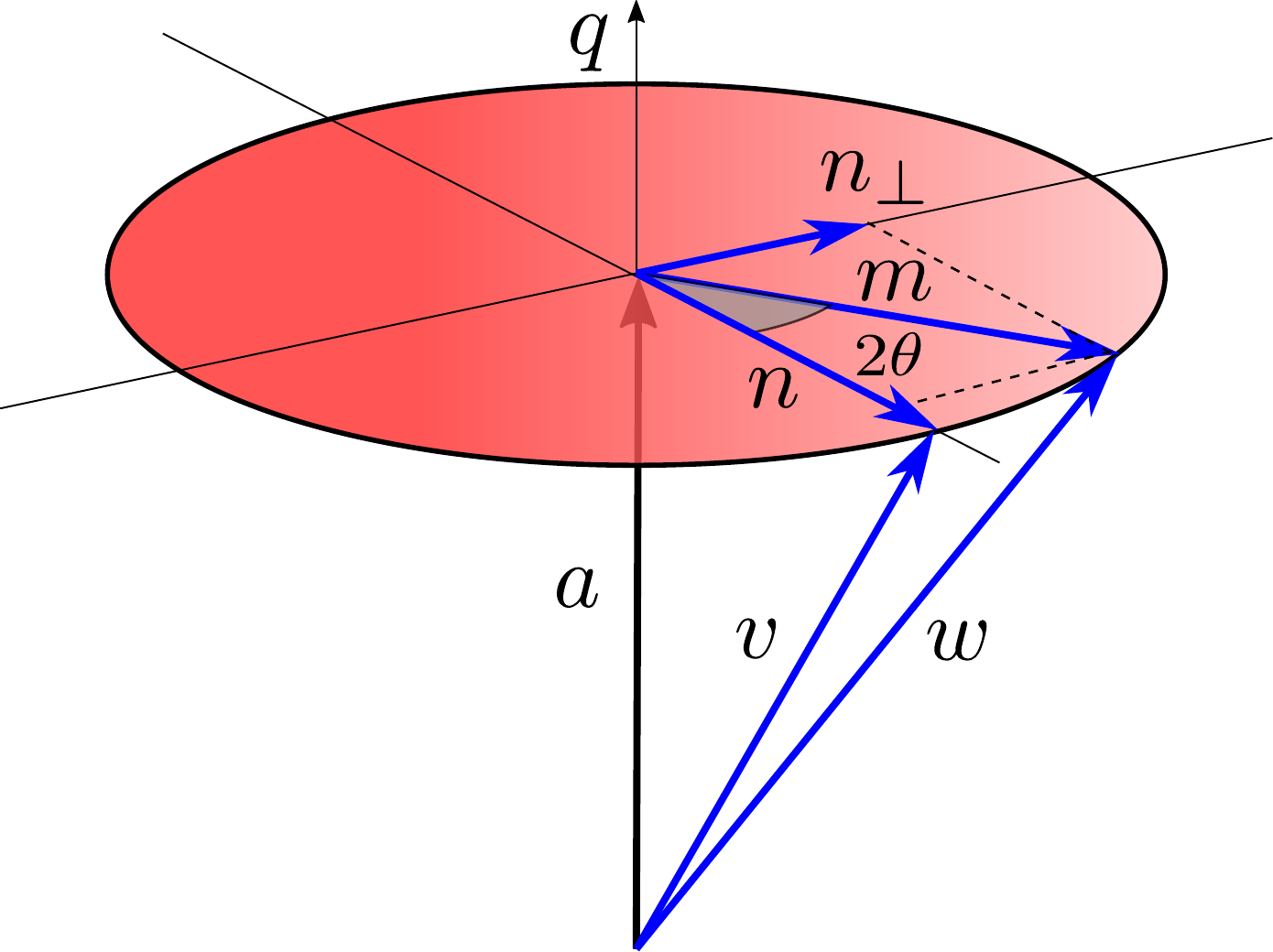}
\caption{Vector $\boldsymbol{v}$ and its image $\boldsymbol{w}$ are related by the rotation about a vector aligned with the quaternion vector.}
\label{fig_quat}
\end{figure}

We can show using Fig.~\ref{fig_quat} that the image of the vector $\boldsymbol{v}$ under rotation around  the vector part of the quaternion $\boldsymbol{q}$, and through an angle ``$2\theta$'' where $q_0=\cos\theta$ ($\phi=2\theta$ in~\eqref{eq_quaternion}) is the scalar part of the quaternion, $\boldsymbol{q}$, to be the vector $\boldsymbol{w}$.\par
The vector $\boldsymbol{v}$ can be resolved into a vector $\boldsymbol{a}$ along the quaternion and a vector $\boldsymbol{n}$ perpendicular to $\boldsymbol{a}$, such that, $\boldsymbol{v}=\boldsymbol{a}+\boldsymbol{n}$.
Since $\boldsymbol{a}$ is aligned with $\boldsymbol{q}$ it is invariant under rotation. On the other hand, it can be easily proved geometrically that $\boldsymbol{m}= \cos2\theta\boldsymbol{n} + \sin2\theta\boldsymbol{n}_{\perp}$.

%
\section{Applied Inertial Navigation}\label{applied}
The main steps in obtaining a solution for a navigation system problem are as follows:
\begin{itemize}[label={\tiny\raisebox{1ex}{\textbullet}}]
\item  Gyro bias corrections
\item Quaternion integration
\item Direction cosine computation
\item Heading, Roll and Pitch computation
\item Delta velocity transformations to earth coordinates
\item Sensed gravity component removal
\item Velocity integration
\item Position and altitude integration
\end{itemize}  

\begin{figure}[h]
    \centering
    \includegraphics[width=1\linewidth]{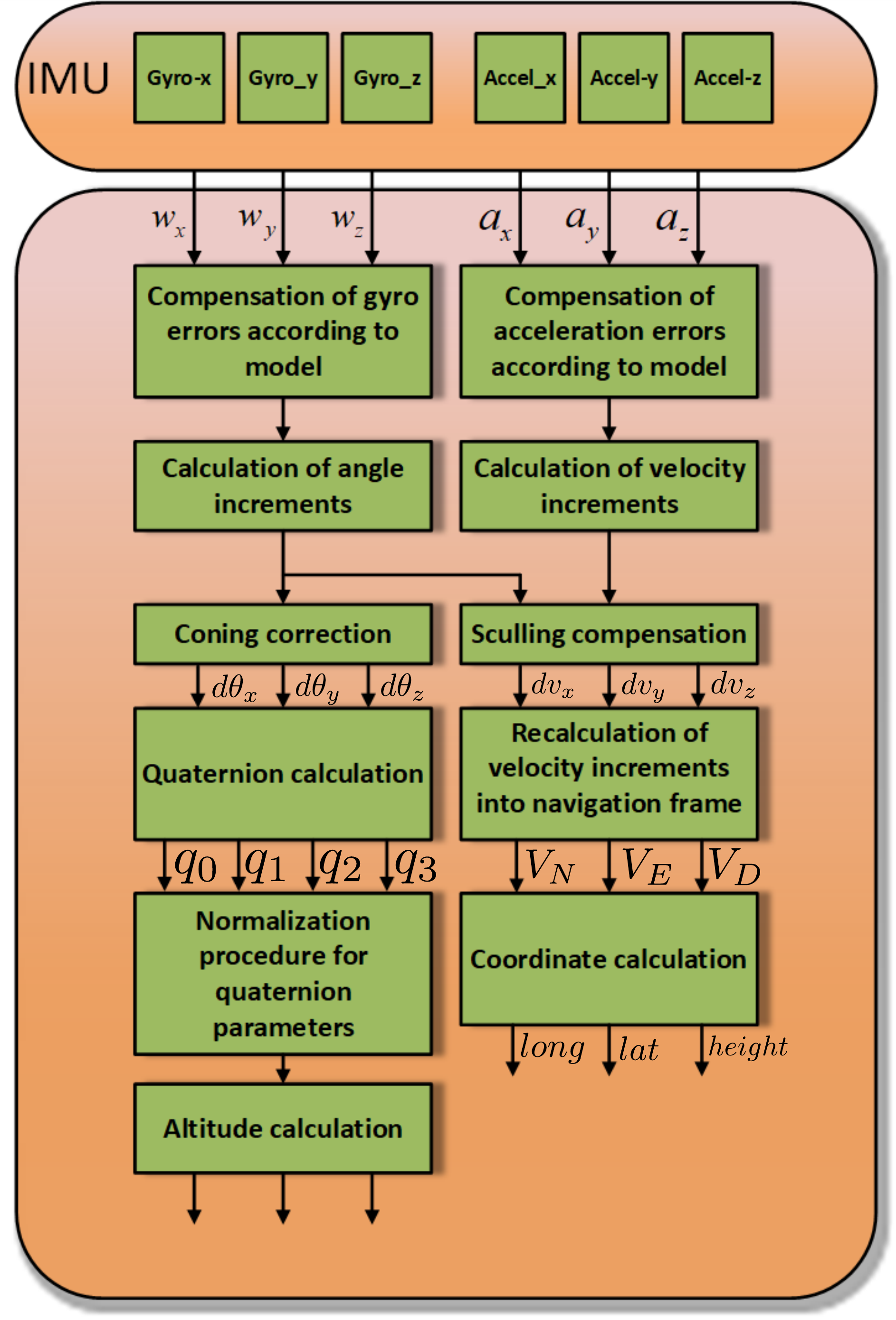}
    \caption{Navigation algorithm flow diagram}
    \label{fig_applied_nav}
\end{figure}

%
\subsection{Gyro Bias Corrections}
When looking at the output of inertial sensors like gyroscopes and accelerometers, you observe that there is a small offset in the average signal output even if the sensors are not moving. This phenomena is known as sensor bias. This bias is the result of physical properties of the sensors that change over time which often lead to increase in sensor bias. The physical properties of sensors are different and so each sensor needs to be calibrated individually. Usually, gyroscopes are factory calibrated for coarse biases but even though, residual biases still remain in the gyroscope outputs. Remember that bias changes so there is no constant value that can be used to compensate for bias. Advanced algorithms are run in real-time to estimate and adjust for these biases.\par
If not corrected for bias, the output orientation of the system will drift with time. Consider a bias of $\unit[0.1]{deg/s}$ in the gyroscope output. This means we would have a drift of, $0.1\times 60 =6$ degrees in the orientation after one minute. Thus, it is crucial to estimate theses biases and compensate for them in the host computer of your inertial navigation system. On-board filters use sensor fusion to predict the biases and correct for them.
The $d\phi$ terms extracted from the IMU are compensated for gyroscope biases by a simple subtraction operation as shown in Algorithm (\ref{alg1}).

\begin{algorithm}[h]
\caption{Gyro bias corrections}
 \hspace*{\algorithmicindent} \textbf{Input:} Estimated biases $\delta\boldsymbol{w}_G=(\delta\omega_{Gx}^s, \delta\omega_{Gy}^s, \delta\omega_{Gz}^s)$  and angular rates $(w_x,w_y,w_z)$.\\
 \hspace*{\algorithmicindent} \textbf{Output:} De-biased angular increments  $d\phi_x,d\phi_y,d\phi_z$. 
\begin{algorithmic}[1]
\While {$IMUread$}
\State ${d\phi}_x=w_x *dT$
\State ${d\phi}_y=w_y *dT$
\State ${d\phi}_z=w_z *dT$
\State ${d\phi}_x={d\phi}_x -\delta\omega_{Gx}^s *dT $
\State ${d\phi}_y={d\phi}_y -\delta\omega_{Gy}^s *dT$
\State ${d\phi}_z={d\phi}_z -\delta\omega_{Gz}^s *dT $
\EndWhile
\State \textit{Gyro Biases estimated maybe either  angular rates in $\deg/\sec$, or angular increments in degrees. In our cases thay are angular rates so they should be multiplied by $dT$ to convert to angular increments\footnotemark.}
\end{algorithmic}
\label{alg1}
\end{algorithm}
\footnotetext{The adapted terminology here for estimated gyroscope biases is $\delta\boldsymbol{w}_G=( \delta\omega_{Gx}^s, \delta\omega_{Gy}^s, \delta\omega_{Gz}^s )   $}

%
\subsection{Coning Correction}
In old navigation systems, the inertial measurement unit (IMU) was mounted on a gimballed platform that was maintained in a horizontal position whenever the vehicle rotated, so that the gyroscopes and the accelerometers did not rotate with the vehicle. Nowadays, in
strap-down systems, the gyroscopes and accelerometers are attached and rotate with the vehicle. To obtain velocities and angles you have to time integrate the reported acceleration and angular rates which is a highly non-linear operation. In particular, if you have high speed motion occurring you have to do the integration really fast to prevent errors from creeping in. This implies the IMU should be sampled at a  really high rate (e.g., \unit[1000]{Hz}) to provide raw angular rates and accelerometer data for the integration process. Since this integration is part of the Kalman filter (see Section~\ref{Kalman}) process, it places a heavy burden on your processor. This forms a major confliction. On the one hand you need to get data a high rate to preserve integration accuracy. On the other hand you can't afford dealing with this high throughput of data. So what the coning algorithm does is to reduce the heavy burden on the navigation processor by performing accurate high speed integration on-board the IMU processor. The output is in the form of delta theta which is the integration of the raw  angular rate data. The benefit is that the delta theta quantity has already captured the integration non-linearities using a high speed coning algorithm. The resulting output still retains accuracy even at a slow rate (e.g., \unit[100]{Hz}). These delta theta quantities are used by the quaternion integration block of the navigation processor to find the attitude of the system.  \par

In order to implement the quaternion integration, the delta theta quantities which form components of the  rotation vector $\boldsymbol{\phi} = [{d\phi}_x, {d\phi}_y,{d\phi}_z]$ for one time step should be calculated. The general equation for the dynamics of this vector  $\dot{\boldsymbol{\phi}}$ can be expressed by the following equation: 
\begin{equation}\label{eq_equation phi dot}
     {\dot{\boldsymbol{\phi}}}= \boldsymbol{\omega}+\frac{1}{2}\boldsymbol{\phi}\times\boldsymbol{\omega}+\frac{1}{\phi^2}\left(1-\frac{\phi\sin\phi}{2(1-\cos\phi)}\right)\boldsymbol{\phi}\times(\boldsymbol{\phi}\times\boldsymbol{\omega}),
\end{equation}
where $\boldsymbol{\phi}$ is the rotation vector that defines the attitude of the  body frame $\boldsymbol{B}$ at general time $t$  relative to frame  $\boldsymbol{B}$ at time $t_{m-1}$, and $\boldsymbol{\omega}$ is the angular rotation rate of frame $\boldsymbol{B}$ relative to inertial space coordinatized in frame $\boldsymbol{B}$.

A more convenient form  for practical implementation would require writing the sine and cosine terms as series expansions and ignoring any terms higher than third order. For example, through a series expansion, the scalar multiplier of the $\boldsymbol{\phi}\times(\boldsymbol{\phi}\times\boldsymbol{\omega})$ term in~\eqref{eq_equation phi dot} can be written as:
\begin{equation}
\frac{1}{\phi^2}\left(1\!-\!\frac{\phi\sin\phi}{2(1\!-\!\cos\phi)}\right)\!=\!\frac{1}{12}\left(1\!+\!\frac{1}{60}\phi^2 \!+\! \cdots \right)\!\approx\!\frac{1}{12}.
\end{equation}
Hence, the rate of change of the rotation vector is given by
\begin{equation}
     {\dot{\boldsymbol{\phi}}}\approx \boldsymbol{\omega}+\frac{1}{2}\boldsymbol{\phi}\times\boldsymbol{\omega}+\frac{1}{12}\boldsymbol{\phi}\times(\boldsymbol{\phi}\times\boldsymbol{\omega}).
\end{equation}
It can be shown through analysis that, to second order accuracy of $\phi$,
\begin{equation}
   \frac{1}{2}\boldsymbol{\phi}\times\boldsymbol{\omega}+\frac{1}{12}\boldsymbol{\phi}\times(\boldsymbol{\phi}\times\boldsymbol{\omega}) \approx  \frac{1}{2}\boldsymbol{\alpha}\times\boldsymbol{\omega},
\end{equation}
with 
\begin{equation}
\boldsymbol{\alpha} = \int_{t_{m-1}}^{t}\boldsymbol{\omega}d\tau,
\end{equation}
where $\boldsymbol{\alpha}$ is the integral of $\boldsymbol{\omega}$ from time ${t_{m-1}}$ to time $t$. 
Thus,~\eqref{eq_equation phi dot} becomes to second order accuracy:
\begin{equation}\label{eq_equation phi approx}
     {\dot{\boldsymbol{\phi}}}\approx \boldsymbol{\omega}+\frac{1}{2}\boldsymbol{\alpha}\times\boldsymbol{\omega}.
\end{equation}
Using~\eqref{eq_equation phi approx}, it is possible to determine the attitude rotation vector that relates the body B frame attitude at time $t_m$ relative to time $t_{m-1}$
\begin{equation}\label{eq_phimequation}
\boldsymbol{\phi}_m = \int_{t_{m-1}}^{t
_m}\boldsymbol{\left[\boldsymbol{\omega}+\frac{1}{2}\boldsymbol{\alpha}\times\boldsymbol{\omega}\right]}d\tau = \boldsymbol{\alpha}_m + \boldsymbol{\beta}_m,
\end{equation}
with 
\begin{align}
    \boldsymbol{\alpha}_m &= \int_{t_{m-1}}^{t_m}\boldsymbol{\omega}d\tau,
    \label{eq_alphaequation}
    \\
    \boldsymbol{\beta}_m &=\frac{1}{2} \int_{t_{m-1}}^{t_m}\left(\boldsymbol{\alpha}\times\boldsymbol{\omega}\right)d\tau,
    \label{eq_betam}
\end{align}
where $\boldsymbol{\beta}_m$ is by definition, the coning attitude motion from time $t_{m-1}$ to time $t_m$. The variable $\boldsymbol{\beta}_m$ has been named coning term since it measures the effects of coning motion present in $\boldsymbol{\omega}$. Coning motion is the condition where the angular velocity vector is itself rotating. As can be easily seen from~\eqref{eq_alphaequation}, $\boldsymbol{\alpha}$ and $\boldsymbol{\omega}$ remain parallel when the angular velocity vector does not rotate. Hence, the $\boldsymbol{\beta}_m$ terms zeroes out since the cross product in its integrand is zero. In this case,~\eqref{eq_phimequation} reduces to 
\begin{equation}
\boldsymbol{\phi}_m = \int_{t_{m-1}}^{t
_m}\boldsymbol{\boldsymbol{\omega}}d\tau.
\end{equation}
This condition can  also be seen directly from~\eqref{eq_equation phi dot} since the second and third terms on the right-hand-side zero out.

%
\subsubsection{Coning algorithm}
In this section, we will develop an efficient digital algorithm for calculating the coning term. The integration time in~\eqref{eq_betam} can be divided into a time up to and after $t_{l-1}$, where $t_{l-1}$ is between $t_{m-1}$ and $t_{m}$. From ~\eqref{eq_betam},
\begin{equation}\label{eq_betadigital}
\begin{array}{c}
\boldsymbol{\beta}_l = \boldsymbol{\beta}_{l-1}+\Delta\boldsymbol{\beta}_l, \quad   \boldsymbol{\beta}_m = \boldsymbol{\beta}_l\bigg\rvert_{t_l=t_m}\,,\\
\\
\boldsymbol{\beta}_l\bigg\rvert_{{t_l=t_{m-1}}} = 0\,,\\
\\
\Delta\boldsymbol{\beta}_l =\frac{1}{2} \int_{t_{l-1}}^{t_l}\left(\boldsymbol{\alpha}\times\boldsymbol{\omega}\right)d\tau\,.
\end{array}
\end{equation}.

A similar process can be utilized to digitize~\eqref{eq_alphaequation} giving the following
\begin{equation}\label{eq_alphadigital}
\begin{array}{c}
\boldsymbol{\alpha}_l = \boldsymbol{\alpha}_{l-1}+\Delta\boldsymbol{\alpha}_l, \quad  \boldsymbol{\alpha}_m = \boldsymbol{\alpha}_l\bigg\rvert_{t_l=t_m}\,,\\
\\
{\alpha}_l\bigg\rvert_{t_l=t_{m-1}} = 0\,,\\
\\
\Delta\boldsymbol{\alpha}_l = \int_{t_{l-1}}^{t_l}\boldsymbol{\omega}d\tau\,.
   \end{array}
\end{equation}

Substituting $\boldsymbol{\alpha} = \boldsymbol{\alpha}_{l-1}+\Delta\boldsymbol{\alpha}(t)$ in $\Delta\boldsymbol{\beta}_l$ of~\eqref{eq_betadigital} we obtain 
\begin{equation}\label{eq_betadigitalfin}
\begin{array}{c}
\Delta\boldsymbol{\beta}_l =\frac{1}{2}\left(\boldsymbol{\alpha}_{l-1}\times\Delta\boldsymbol{\alpha}_l\right)+\frac{1}{2} \int_{t_{l-1}}^{t_l}\left(\Delta\boldsymbol{\alpha}(t)\times\boldsymbol{\omega}\right)d\tau\,,\\
\\
\boldsymbol{\beta}_l = \boldsymbol{\beta}_{l-1}+\Delta\boldsymbol{\beta}_l, \quad   \boldsymbol{\beta}_m = \boldsymbol{\beta}_l\bigg\rvert_{t_l=t_m}\,,\\
\\
\boldsymbol{\beta}_l\bigg\rvert_{{t_l=t_{m-1}}} = 0\,.\\
\end{array}
\end{equation}

Equations~\eqref{eq_alphadigital} and~\eqref{eq_betadigitalfin} form the basis for a recursive digital algorithm at the high $l$ rate of the on-board IMU processor  to calculate the $\boldsymbol{\alpha}_m$ and the coning term $\boldsymbol{\beta}_m$ of the low $m$ rate  of~\eqref{eq_phimequation}. What remains is to determine a digital integration algorithm for the integral term in~\eqref{eq_betadigitalfin}.

In order to digitize the integral term in~\eqref{eq_betadigitalfin}, it is wise to consider an linear analytical form for the angular rate vector $\boldsymbol\omega$ between any two time steps $t_{l-1}$ and $t_l$. Approximating $\boldsymbol\omega$ profile as a constant $\boldsymbol{a}$ added to a linear build-up in time having rate $\boldsymbol{b}$, we obtain
\begin{equation}\label{eq_omega}
    \boldsymbol\omega \approx \boldsymbol{a} + \boldsymbol{b}(t-t_{l-1}), 
\end{equation}
where both $\boldsymbol{a}$ and  $\boldsymbol{b}$ are constant vectors. Therefore, both constants can  be determined from current and previous values of $\Delta\boldsymbol{\alpha}_l$
\begin{equation}\label{eq_AB}
    \boldsymbol{a} =\frac{1}{2T_l} \left(\Delta\boldsymbol{\alpha}_l+\Delta\boldsymbol{\alpha}_{l-1} \right)\,,\quad \boldsymbol{b} =\frac{1}{T_{l}^{2}} \left(\Delta\boldsymbol{\alpha}_l-\Delta\boldsymbol{\alpha}_{l-1}   \right)\,.
\end{equation}

Substituting~\eqref{eq_AB} in~\eqref{eq_omega} and the integral part of~\eqref{eq_betadigitalfin} gives
\begin{equation}
\frac{1}{2} \int_{t_{l-1}}^{t_l}\left(\Delta\boldsymbol{\alpha}(t)\times\boldsymbol{\omega}\right)d\tau = \frac{1}{12} \left(\Delta\boldsymbol{\alpha}_{l-1}\times\Delta\boldsymbol{\alpha}_{l}\right).
\end{equation}
When substituted in~\eqref{eq_betadigitalfin}, the final result is
\begin{equation}
\Delta\boldsymbol{\beta}_l =\frac{1}{2}\left(\boldsymbol{\alpha}_{l-1}+\frac{1}{6}\Delta\boldsymbol{\alpha}_{l-1}\right)\times \Delta\boldsymbol{\alpha}_l.
\end{equation}

The overall digital algorithm for $\boldsymbol{\alpha}_m$ and the coning term $\boldsymbol{\beta}_m$ is determined from the above results and abbreviated in Algorithm \ref{coning_alg}. 

\begin{algorithm}[h]
\caption{Coning algorithm}\label{coning_alg}
 \hspace*{\algorithmicindent} \textbf{Input:} $({d\phi}_x, {d\phi}_y,{d\phi}_z)$ at high-speed cycle index $l$. \\
 \hspace*{\algorithmicindent} \textbf{Output:} $\boldsymbol{\phi}_m $ vector of angular increments at low-speed index $m$.
\begin{algorithmic}[1]
\State ${\alpha}_l \gets 0\quad at\quad (t=t_{m-1})$ \Comment \textit{$m$ the low-speed computer cycle index} 
\State ${\beta}_l \gets 0\quad at\quad (t=t_{m-1})$
\While {$t_l < t_m$}\Comment \textit{$l$ is the high-speed computer cycle index  }
\State $\Delta\boldsymbol{\alpha}_l \gets ({d\phi}_x, {d\phi}_y,{d\phi}_z)$
\State $\boldsymbol{\alpha}_l = \boldsymbol{\alpha}_{l-1}+\Delta\boldsymbol{\alpha}_l$
\State $\Delta\boldsymbol{\beta}_l =\frac{1}{2}\left(\boldsymbol{\alpha}_{l-1}+\frac{1}{6}\Delta\boldsymbol{\alpha}_{l-1}\right)\times \Delta\boldsymbol{\alpha}_l$
\State $\boldsymbol{\beta}_l = \boldsymbol{\beta}_{l-1}+\Delta\boldsymbol{\beta}_l$
\EndWhile
\State $\boldsymbol{\alpha}_m = \boldsymbol{\alpha}_l\quad at\quad (t_l=t_m)$
\State $\boldsymbol{\beta}_m = \boldsymbol{\beta}_l\quad at\quad (t_l=t_m) $
\State $\boldsymbol{\phi}_m =  \boldsymbol{\alpha}_m + \boldsymbol{\beta}_m$
\State \Comment \textit{$\boldsymbol{\phi}_m$ vector contains the integration of delta theta terms between two $m$ computer cycle indices with very high accuracy.}
\end{algorithmic}
\end{algorithm}

%
\subsection{Sculling Compensation}
Sculling on the other hand is basically analogous to coning but it has to do with the accelerometers instead of the gyroscopes. Coning relates specifically to an error in your angle measurement and so fundamentally it is coming from gyro data. On the other hand, sculling happens when you have a cyclic linear acceleration in combination with cyclic rotation. We call this sculling because it results in an apparent but erroneous velocity, and the characteristic motion
that gives you this erroneous velocity looks like the sculling type of oar, where the oar sweeps back and forth. Without compensation, this would come out in the delta velocity quantity and the output would have that corruption built into it. For example, if you have very fast motion, especially a vibration-like oscillating motion at the same time that you have a slow sampling
rate, you will be in trouble without the sculling compensation provided.\par

Therefore, in order to prevent delta velocity errors from creeping in, it is convenient to account for the body frame rotation $C_{B_{
(t)}}^{B_{m-1}}$ during the $m$th computer cycle index period. To find delta velocities, we integrate the reported accelerometer measurements according to the following
\begin{equation}\label{eq_deltaV}
\Delta\boldsymbol{v}_m = \int_{t_{m-1}}^{t_m}C_{B_{(t)}}^{B_{m-1}} \boldsymbol{a}_{SF}dt,
\end{equation}
where $\boldsymbol{a}_{SF}$ is the accelerometer reported values and $C_{B_{(t)}}^{B_{m-1}}$ the general direction cosine matrix defining the attitude of Frame ${B}$ relative to Frame ${B_{m-1}}$ for time $t$ greater than $t_{m-1}$.

The $C_{B_{(t)}}^{B_{m-1}}$ term in~\eqref{eq_deltaV} can be expressed as:
\begin{equation}\label{eq_C}
     C_{B_{(t)}}^{B_{m-1}}=I + \frac{\sin\phi(t)}{\phi(t)}\left[\boldsymbol{\phi}(t)\times\right]+\frac{1-\cos\phi(t)}{{\phi(t)}^2}{\left[\boldsymbol{\phi}(t)\times\right]}^2,
\end{equation}
where $\boldsymbol{\phi}(t)= $ Rotation vector that defines the attitude of the  body frame ${B}$
at general time $t$  relative to frame  ${B_{m-1}}$ at time $t_{m-1}$, and $\phi(t)=$ Magnitude  of $\boldsymbol{\phi}(t)$.\par
  A first order approximation for~\eqref{eq_C} neglects ${\left[\boldsymbol{\phi}(t)\times\right]}^2$ and approximates ${\sin\phi(t)}/{\phi(t)}$ by unity. Assuming that the $m$ cycle rate is selected fast enough to maintain $\boldsymbol{\phi}(t)$ small, e.g., less that 0.05 radians, we can write $\boldsymbol{\phi}(t) \approx  \boldsymbol{\alpha}(t) $. In this case~\eqref{eq_C} becomes
\begin{equation}\label{eq_Capprox}
     C_{B(t)}^{B_{m-1}}\approx I +\left[\boldsymbol{\alpha}(t)\times\right]\,.
\end{equation}

Substituting~\eqref{eq_Capprox}  in~\eqref{eq_deltaV} then yields to first order 
\begin{equation}\label{eq_deltaVapprox}
\begin{array}{c}
\begin{split}
\Delta\boldsymbol{v}_m &= 
\int_{t_{m-1}}^{t_m}\left(I +\left[\boldsymbol{\alpha}(t)\times\right] \boldsymbol{a}_{SF}\right)dt\\ 
&= \int_{t_{m-1}}^{t_m}{\boldsymbol{a}_{SF}dt} + \int_{t_{m-1}}^{t_m} \left(\boldsymbol{\alpha}(t)\times\right) \boldsymbol{a}_{SF}dt\,,
 \end{split}\\
\\
\Delta\boldsymbol{v}_m =\boldsymbol{v}_m +\int_{t_{m-1}}^{t_m}\left(\boldsymbol{\alpha}(t)\times \boldsymbol{a}_{SF}\right)dt\,, \\
\\
\boldsymbol{\alpha}(t) = \int_{t_{m-1}}^{t}\boldsymbol{\omega}d\tau, \quad \boldsymbol{\alpha}_m =  \boldsymbol{\alpha}(t_m)\,,\\
\\
\boldsymbol{v}(t)=\int_{t_{m-1}}^{t}{\boldsymbol{a}_{SF}d\tau} ,\quad \boldsymbol{v}_m =  \boldsymbol{v}(t_m)\,.
\end{array}
\end{equation}

Equation~\eqref{eq_deltaVapprox} can be further synthesized if we work on the integral term by first noting that:
\begin{equation}\label{eq_y}
\begin{split}
     \frac{d}{dt}\left(\boldsymbol{\alpha}(t)\times {\boldsymbol{v}}(t)\right)&= \boldsymbol{\alpha}(t)\times \dot{\boldsymbol{v}}(t) +\dot{\boldsymbol{\alpha}}(t)\times {\boldsymbol{v}}(t)\\
     &= \boldsymbol{\alpha}(t)\times \dot{\boldsymbol{v}}(t) - {\boldsymbol{v}}(t)\times \dot{\boldsymbol{\alpha}}(t)\,.
     \end{split}
\end{equation}
Upon re-arranging this equation, we obtain
\begin{equation}\label{eq_tt}
   \boldsymbol{\alpha}(t)\times \dot{\boldsymbol{v}}(t) = \frac{d}{dt}\left(\boldsymbol{\alpha}(t)\times \boldsymbol{v}(t)\right) + \boldsymbol{v}(t)\times \dot{\boldsymbol{\alpha}}(t)\,.
\end{equation}

Trivially,
\begin{equation}\label{eq_ty}
   \boldsymbol{\alpha}(t)\times \dot{\boldsymbol{v}}(t) =  \frac{1}{2}\boldsymbol{\alpha}(t)\times \dot{\boldsymbol{v}}(t) + \frac{1}{2} \boldsymbol{\alpha}(t)\times \dot{\boldsymbol{v}}(t)\,.
\end{equation}
We now substitute for one of the terms on the right to obtain
\begin{equation}\label{eq_tx}
\begin{split}
   \boldsymbol{\alpha}(t)\times \dot{\boldsymbol{v}}(t) &=  \frac{1}{2}\frac{d}{dt}\left(\boldsymbol{\alpha}(t)\times \boldsymbol{v}(t)\right) + \\
   &\frac{1}{2} \left(\boldsymbol{\alpha}(t)\times \dot{\boldsymbol{v}}(t)+
   \boldsymbol{v}(t)\times \dot{\boldsymbol{\alpha}}(t)\right)\,.
   \end{split}
\end{equation}
Knowing that $\dot{\boldsymbol{\alpha}}(t) = \boldsymbol\omega$ and $\dot{\boldsymbol{v}}(t)= \boldsymbol{a}_{SF}$,~\eqref{eq_tx} becomes
\begin{equation}\label{eq_ty}
\begin{split}
   \boldsymbol{\alpha}(t)\times\boldsymbol{a}_{SF} &=  \frac{1}{2}\frac{d}{dt}\left(\boldsymbol{\alpha}(t)\times \boldsymbol{v}(t)\right) + \\
   &\frac{1}{2} \left(\boldsymbol{\alpha}(t)\times \boldsymbol{a}_{SF} +\boldsymbol{v}(t)\times \boldsymbol{\omega}\right)\,.
   \end{split}
\end{equation}

Substituting~\eqref{eq_ty} for the integrand in~\eqref{eq_deltaVapprox} yields the following
\begin{equation}\label{eq_deltaVapproxnew}
\begin{split}
\Delta\boldsymbol{v}_m &= \boldsymbol{v}_m+\frac{1}{2}\left(\boldsymbol{\alpha}_m \times \boldsymbol{v}_m \right)+\\
&\int_{t_{m-1}}^{t_m}\frac{1}{2}  \left(\boldsymbol{\alpha}(t)\times \boldsymbol{a}_{SF} +\boldsymbol{v}(t)\times \boldsymbol{\omega}\right)dt\,.\\
\end{split}
\end{equation}
It is easily verified that the integrand in~\eqref{eq_deltaVapproxnew} vanishes for the cases where the angular velocity term $\boldsymbol{\omega}$ and the specific force $\boldsymbol{a}_{SF}$ are  non-rotating and having constant magnitudes. We conclude that the integral term in~\eqref{eq_deltaVapproxnew} represents a contribution from rotating high frequency components in $\Delta\boldsymbol{v}_m $.

The integral term in~\eqref{eq_deltaVapproxnew}, denoted as ``sculling", measures the ``constant" contribution to $\Delta\boldsymbol{v}$ under classical sculling motion (mariners propel boats using a single oar with an undulating motion) where the $\boldsymbol{\alpha}(t)$ angular excursion term about one body frame  axis is at the same frequency and in phase with the specific force $\boldsymbol{a}_{SF}$ along another axis.

The other terms in~\eqref{eq_deltaVapproxnew}, $\boldsymbol{v}_m+\frac{1}{2}\left(\boldsymbol{\alpha}_m \times \boldsymbol{v}_m \right)$, represent a combination of both low-frequency and  high frequency effects. In particular, $\frac{1}{2}\left(\boldsymbol{\alpha}_m \times \boldsymbol{v}_m \right)$ is denoted as  velocity rotation compensation term. With this terminology,~\eqref{eq_deltaVapproxnew} can be re-written as 
\begin{equation}\label{eq_deltaVapproxnewnew}
 \begin{split}
 \begin{array}{c}
\Delta\boldsymbol{v}_m = \boldsymbol{v}_m+ \Delta\boldsymbol{v}_{Rot_{m}} + \Delta\boldsymbol{v}_{Scul_{m}}\,,\\
\\
\Delta\boldsymbol{v}_{Scul_{m}}=\int_{t_{m-1}}^{t_m}\frac{1}{2}  \left(\boldsymbol{\alpha}(t)\times \boldsymbol{a}_{SF} +\boldsymbol{v}(t)\times \boldsymbol{\omega}\right)dt\,,\\
\\
\boldsymbol{\alpha}(t) = \int_{t_{m-1}}^{t}\boldsymbol{\omega}d\tau, \quad\boldsymbol{\alpha}_m =  \boldsymbol{\alpha}(t_m)\,,\\
\\
\boldsymbol{v}(t)=\int_{t_{m-1}}^{t}{\boldsymbol{a}_{SF}d\tau}, \quad \boldsymbol{v}_m =  \boldsymbol{v}(t_m)\,,
\end{array}
\end{split}
\end{equation}
and
\begin{equation}\label{deltaVapproxnewnnn}
\Delta\boldsymbol{v}_{Rot_{m}} =\frac{1}{2}\left(\boldsymbol{\alpha}_m \times \boldsymbol{v}_m \right)\,.\\
\end{equation}
where $\Delta\boldsymbol{v}_{Rot_{m}}=$ ``Velocity Rotation Compensation" term, and 
$\Delta\boldsymbol{v}_{Scul_{m}}= $ ``Sculling" term.
 In order to develop a digital algorithm for calculating the terms in~\eqref{eq_deltaVapproxnewnew}, we follow an identical procedure to that used for the coning algorithm. We consider the integration in~\eqref{eq_deltaVapproxnewnew} as divided into portions up to and after a general time $t_{l-1}$ within the $t_{m-1}$ to $t_{m}$interval so that it becomes
\begin{equation}\label{eq_digitalscul}
\begin{array}{c}
\Delta\boldsymbol{v}_{Scul}(t) =\Delta\boldsymbol{v}_{Scul_{l-1}} +\delta\boldsymbol{v}_{Scul}(t)\,,\\
\\
\delta\boldsymbol{v}_{Scul}(t) = \int_{t_{l-1}}^{t}\frac{1}{2}  \left(\boldsymbol{\alpha}(\tau)\times \boldsymbol{a}_{SF} +\boldsymbol{v}(\tau)\times \boldsymbol{\omega}\right)d\tau\,,
\end{array}
\end{equation}
Now let us define the next $l$ cycle time within the $t_{m-1}$ to $t_{m}$ interval so that at $t_l$ we can write
\begin{equation}\label{eq_digitalscul}
\begin{array}{c}
\boldsymbol{\alpha}_l = \boldsymbol{\alpha}_{l-1}+\Delta\boldsymbol{\alpha}_l, \quad  \boldsymbol{\alpha}_m = \boldsymbol{\alpha}_l\bigg\rvert_{t_l=t_m}\,,\\
\\
\Delta\boldsymbol{\alpha}(\tau) = \int_{t_{l-1}}^{\tau}\boldsymbol{\omega}dt, \quad \Delta\boldsymbol{\alpha}_l = \int_{t_{l-1}}^{t_l}\boldsymbol{\omega}dt\,,\\
\\
{\alpha}_l\bigg\rvert_{t_l=t_{m-1}} = 0,\\
\\
\boldsymbol{v}_l = \boldsymbol{v}_{l-1}+\Delta\boldsymbol{v}_l, \quad   \boldsymbol{v}_m = \boldsymbol{v}_l\bigg\rvert_{t_l=t_m}\,,\\
\\
\Delta\boldsymbol{v}(\tau) = \int_{t_{l-1}}^{\tau}\boldsymbol{a}_{SF}dt \quad \Delta\boldsymbol{v}_l = \int_{t_{l-1}}^{t_l}\boldsymbol{a}_{SF}dt\,,\\
\\
{v}_l\bigg\rvert_{t_l=t_{m-1}} = 0\,,\\
\\
\Delta\boldsymbol{v}_{Scul_l} =\Delta\boldsymbol{v}_{Scul_{l-1}} +\delta\boldsymbol{v}_{Scul_l}\,,\\
\\
\delta\boldsymbol{v}_{Scul}(t) =\int_{t_{l-1}}^{t}\frac{1}{2}  \left(\boldsymbol{\alpha}(\tau)\times \boldsymbol{a}_{SF} +\boldsymbol{v}(\tau)\times \boldsymbol{\omega}\right)d\tau\,,\\
\\
\Delta\boldsymbol{v}_{Scul_m} =\Delta\boldsymbol{v}_{Scul_{l}}\bigg\rvert_{t_l=t_m}, \Delta\boldsymbol{v}_{Scul_{l}}\bigg\rvert_{t_l=t_{m-1}}=0\,. \\
\\
\end{array}
\end{equation}

Substituting for the terms $\boldsymbol{\alpha}$ and $\boldsymbol{v}$ using~\eqref{eq_deltaVapprox} and incorporating the definition for $\Delta\boldsymbol{\alpha}_l$ and $\Delta\boldsymbol{v}_l$,~\eqref{eq_digitalscul} becomes
\begin{equation}\label{eq_digitalsculnewnnn}
\begin{split}
\delta\boldsymbol{v}_{Scul_l} &=\frac{1}{2}\left( \boldsymbol{\alpha}_{l-1}\times  \Delta\boldsymbol{v}_l + \boldsymbol{v}_{l-1} \times \Delta\boldsymbol{\alpha}_l  \right)+\\
&\phantom{=}\int_{t_{l-1}}^{t_l}\frac{1}{2}  \left(\Delta\boldsymbol{\alpha}(t)\times \boldsymbol{a}_{SF} +\Delta\boldsymbol{v}(t)\times \boldsymbol{\omega}\right)dt\,.\\
\end{split}
\end{equation}

As in the coning algorithm design process, we base our development on an assumed form for the angular rate and specific-force vectors during the  $t_{l-1}$ to $t_l$ time interval. In this case, we propose a linearly changing angular rate and specific-force vector over the $t_{l-1}$ to $t_l$ time interval, where its coefficients are computed from current and past $l$ cycle sensor samples.Thus we have:
\begin{equation}\label{eq_lineqscul}
    \boldsymbol{\omega}\approx\boldsymbol{a}+\boldsymbol{b}(t-t_{l-1}), \quad  \boldsymbol{a}_{SF}\approx\boldsymbol{c}+ \boldsymbol{d}(t-t_{l-1})\,,
\end{equation}
where $\boldsymbol{a}.\boldsymbol{b},\boldsymbol{c},\boldsymbol{d} = $ Constant vectors.
With~\eqref{eq_lineqscul} and the $\Delta\boldsymbol{\alpha}$ and $\Delta\boldsymbol{v}$ definitions in~\eqref{eq_digitalscul}
\begin{equation}\label{eq_lineqscul2}
\begin{array}{c}
  \Delta\boldsymbol{\alpha}(t) = \boldsymbol{a}(t-t_{l-1}) +\frac{1}{2}\boldsymbol{b}(t-t_{l-1})^2\,, \\
  \\
\Delta\boldsymbol{v}(t) = \boldsymbol{c}(t-t_{l-1}) +\frac{1}{2}\boldsymbol{d}(t-t_{l-1})^2\,. \\
\end{array}
\end{equation}

Substituting~\eqref{eq_lineqscul} and~\eqref{eq_lineqscul2} for the integrand in~\eqref{eq_digitalsculnewnnn} yields:
\begin{equation}\label{eq_gg}
\begin{split}
\int_{t_{l-1}}^{t_l}\frac{1}{2}  \left(\Delta\boldsymbol{\alpha}(t)\times \boldsymbol{a}_{SF} +\Delta\boldsymbol{v}(t)\times \boldsymbol{\omega}\right)dt=\\ \frac{1}{12}\left(\boldsymbol{a} \times \boldsymbol{d} + \boldsymbol{c} \times \boldsymbol{b}  \right)T_{l}^{3}\,.
\end{split}
\end{equation}
where $T_l = $ time interval $t_l - t_{l-1}$, i.e., the $l$ cycle computation period. The constants $\boldsymbol{a}, \boldsymbol{b}, \boldsymbol{c}$, and  $\boldsymbol{d}$ can be calculated for each $t_{l-1}$ to $t_l$ time interval using successive measurements of integrated angular rate and specific force acceleration increments from the inertial sensors. To determine the constants $\boldsymbol{a}, \boldsymbol{b}, \boldsymbol{c}$,  and $\boldsymbol{d}$ uniquely, it is required to take sample measurements from two successive intervals. For sensor samples taken at the $l$ cycle rate the results are as follows:
\begin{equation}\label{eq_kk}
\begin{array}{cl}
\boldsymbol{a} = \frac{1}{2T_l}\left(\Delta\boldsymbol{\alpha}_l+\Delta\boldsymbol{\alpha}_{l-1} \right)\,, & \boldsymbol{b} = \frac{1}{T_l^2}\left(\Delta\boldsymbol{\alpha}_l-\Delta\boldsymbol{\alpha}_{l-1} \right)\\
\\
\boldsymbol{c} = \frac{1}{2T_l}\left(\Delta\boldsymbol{v}_l +\Delta\boldsymbol{v}_{l-1} \right)\,, & \boldsymbol{d} = \frac{1}{T_l^2}\left(\Delta\boldsymbol{v}_l-\Delta\boldsymbol{v}_{l-1}  \right)\,.\\
\end{array}
\end{equation}
Substituting the terms in~\eqref{eq_kk} in~\eqref{eq_gg} we obtain the desired equation for $\delta\boldsymbol{v}_{Scul_l}$:
\begin{equation}\label{eq_jk}
 \begin{split}
  \delta\boldsymbol{v}_{Scul_l} &=\frac{1}{2}\left[\left( \boldsymbol{\alpha}_{l-1}+ \frac{1}{6}\Delta\boldsymbol{\alpha}_{l-1}\right) \times \Delta\boldsymbol{v}_l  +\right. \\
 &\left.\phantom{==[}\left(\boldsymbol{v}_{l-1} +\frac{1}{6}\Delta\boldsymbol{v}_{l-1} \right)\times\Delta\boldsymbol{\alpha}_l \right]\,.
 \end{split}
\end{equation}

A digital algorithm from the above results and from the coning equations yields the  sculling Algorithm \ref{sculling}.

\begin{algorithm}[h]
\caption{Sculling algorithm}\label{sculling}
 \hspace*{\algorithmicindent} \textbf{Input:} $(dv_x, dv_y,dv_z)$ at high-speed computer cycle-index $l$.\\
 \hspace*{\algorithmicindent} \textbf{Output:} $\Delta\boldsymbol{v}_m=(dv_x, dv_y,dv_z)$ vector of integrated delta velocity terms at low-speed cycle index $m$.
\begin{algorithmic}[1]
\State ${v}_l \gets 0\quad at\quad (t=t_{m-1})$ \Comment \textit{$m$ the low-speed computer cycle index.}
\State $\Delta\boldsymbol{v}_{Scul_l} \gets 0\quad at\quad (t=t_{m-1})$
\While {$t_l < t_m$}\Comment \textit{$l$ is the high-speed computer cycle index}
\State $\Delta\boldsymbol{v}_l \gets (dv_x, dv_y,dv_z)$
\State $\boldsymbol{v}_l = \boldsymbol{v}_{l-1}+\Delta\boldsymbol{v}_l$
\State \begin{equation}\begin{split}
\delta\boldsymbol{v}_{Scul_l}=&\frac{1}{2}\left[\left(\boldsymbol{\alpha}_{l-1}+\frac{1}{6}\Delta\boldsymbol{\alpha}_{l-1}\right)\times \Delta\boldsymbol{v}_l +\right. \\
&\left. \phantom{=[}\left(\boldsymbol{v}_{l-1}+\frac{1}{6}\Delta\boldsymbol{v}_{l-1}\right)\times \Delta\boldsymbol{\alpha}_l \right]\end{split}\end{equation} 
\State $\Delta\boldsymbol{v}_{Scul_l}=\Delta\boldsymbol{v}_{Scul_{l-1}} + 
\delta\boldsymbol{v}_{Scul_l} $
\EndWhile
\State $\boldsymbol{v}_m = \boldsymbol{v}_l\quad at\quad (t_l=t_m)$
\State $\Delta\boldsymbol{v}_{Scul_m}= \Delta\boldsymbol{v}_{Scul_{l}}\quad at\quad (t_l=t_m) $
\State $\Delta\boldsymbol{v}_m  =  \boldsymbol{v}_m + \Delta\boldsymbol{v}_{Scul_m} $ \Comment  \textit{vector that contains the integration of delta velocity terms between two $m$ computer cycle indices with no loss of accuracy.} 
\end{algorithmic}
\end{algorithm}

%
\subsection{Velocity Increments Transformation}
The velocity increments $(dv_x, dv_y,dv_z)$ output from the sculling compensation algorithm are described in the body frame. In order to perform the velocity integration in the navigation coordinate frame, it is essential that we transform their values to the NED frame. This can be easily done with help of the direction-cosine-matrix $C_b^n$ as shown in Algorithm (\ref{velinteg}).

\begin{algorithm}[h]
\caption{Velocity increment transformation}\label{velinteg}
 \hspace*{\algorithmicindent} \textbf{Input:} $(dv_x, dv_y,dv_z)$ vector of integrated delta velocity terms at low-speed cycle index $m$.\\
 \hspace*{\algorithmicindent} \textbf{Output:}$(dv_n, dv_e,dv_d)$ vector of integrated delta velocity terms in navigation frame. 
\begin{algorithmic}[1]
\State $dv_n = c_{11}dv_x +c_{12}dv_y +c_{13}dv_z$  
\State $dv_e = c_{21}dv_x +c_{22}dv_y +c_{23}dv_z$  
\State $dv_d = c_{31}dv_x +c_{32}dv_y +c_{33}dv_z$ 
\State $dv_d = dv_d + g.dT$ \Comment \textit{Sensed gravity component removal.}
\end{algorithmic}
\end{algorithm}

%
\subsection{Quaternion Integration}
Quaternion integration deals with the determination of the quaternion between the body and the navigation frame.
A primary advantage of using the quaternion technique lies in the fact that only four unknowns are necessary for calculation of the  transformation matrix,  while the direction cosine method requires nine.
The quaternion can also be expressed as a 4x4 matrix. Thus
\begin{equation}
    Q = \left[\begin{array}{rrrr}
         {q_0}&{q_1}&{q_2}&{q_3}\\
         -{q_1}&{q_0}&{-q_3}&{q_2}\\ 
         -{q_2}&{q_3}&{q_0}&{-q_1}\\
         -{q_3}&{-q_2}&{q_1}&{q_0}\\
    \end{array} \right]\,,
    \label{eq_quat_dyn}
\end{equation}
where, as before, $q_0$, $q_1$, $q_2$, $q_3$ are quaternion components.

It can be shown that the quaternion analog of Puasson equation (see Section~\ref{principles} equation~\eqref{eq_CNI_dot}) has the form
\begin{equation}
    \dot{Q}=\frac{1}{2}Q[w\times],
\end{equation}
where $[w\times]$ is the skew-symmetric form of the angular velocity vector $\boldsymbol{w}$.
The recurrent solution of the above equation can be determined  (to first order) as
\begin{equation}
    Q_{k+1}=Q_{k}+ \frac{1}{2}Q_{k}[w\times] dT,
\end{equation}
or
\begin{equation}
    Q_{k+1}=Q_{k}(I+ \frac{1}{2}[w\times] dT)=Q_{k}d{\Lambda},
    \label{eq_recur}
\end{equation}
where $dT$ is the sampling period and $d{Q}=(I+ \frac{1}{2}[w\times] dT)$ is usually called the update quaternion. It is the quaternion of a small rotation that can be represented using~\eqref{eq_quaternion} as follows
\begin{equation}
         \begin{array}{c}
        \begin{aligned}
        d\Lambda &= d\lambda_0+d\lambda_{1}i+d\lambda_{2}j+d\lambda_{3}k\,,\\
         d\lambda_0 &= \cos\frac{{\|d\mathbf\Phi\|}}{2}\,,   \\
         
         d\lambda_1 &=\frac{{d\phi}_x}{\|d\Phi\|} \sin\frac{{\|d\mathbf\Phi\|}}{2}\,, \\
         
         d\lambda_2 &= \frac{{d\phi}_y}{\|d\Phi\|}\sin\frac{{\|d\mathbf\Phi\|}}{2}\,,\\
         
         d\lambda_3 &= \frac{{d\phi}_z}{\|d\Phi\|}\sin\frac{{\|d\mathbf\Phi\|}}{2}\,. \\
         \end{aligned}
        \end{array}
        \label{eq_delta_phi}
\end{equation}

Substituting in~\eqref{eq_recur} the expression obtained is:
\mathleft
\begin{equation}
\begin{split}
 {Q}_{k+1}&= \left[\begin{array}{rrrr}
         {q_0}&{q_1}&{q_2}&{q_3}\\
         -{q_1}&{q_0}&{-q_3}&{q_2}\\ 
         -{q_2}&{q_3}&{q_0}&{-q_1}\\
         -{q_3}&{-q_2}&{q_1}&{q_0}\\
    \end{array} \right]*\\
  & \qquad\qquad\quad \left[\begin{array}{rrrr}
         {d\lambda_0}&{d\lambda_1}&{d\lambda_2}&{d\lambda_3}\\
         -{d\lambda_1}&{d\lambda_0}&{-d\lambda_3}&{d\lambda_2}\\ 
         -{d\lambda_2}&{d\lambda_3}&{d\lambda_0}&{-d\lambda_1}\\
         -{d\lambda_3}&{-d\lambda_2}&{d\lambda_1}&{d\lambda_0}\\
    \end{array} \right]\,.
    \label{eq_quat_dyn}
    \end{split}
\end{equation}
\mathcenter

But $\sin\frac{{\|d\mathbf\Phi\|}}{2}$ and $\cos\frac{{\|d\mathbf\Phi\|}}{2}$ can be approximated using a third order expansion of the Taylor series:
\begin{equation}
    \sin{x} \approx x - \frac{x^3}{3!}+\frac{x^5}{5!},\quad 
     \cos{x} \approx 1 - \frac{x^2}{2!}+\frac{x^4}{4!}\,.
\end{equation}

The series expansion of~\eqref{eq_delta_phi} gives the following formula for the quaternion components:
\begin{equation}
         \begin{array}{c}
        \begin{aligned}
      
         d\lambda_0 &=1- \frac{{\|d\mathbf\Phi\|}^2}{8} + \frac{{\|d\mathbf\Phi\|}^4}{384}\,, \\
         
         d\lambda_1 &=r{{d\phi}_x}\,, \\
         
         d\lambda_2 &= r{{d\phi}_y}\,,\\
         
         d\lambda_3 &= r{{d\phi}_z}\,, \\
         \end{aligned}
        \end{array}
        \label{eq_delta_ser}
\end{equation}
where $r=\frac{1}{2}-\frac{{\|d\mathbf\Phi\|}^2}{48}+\frac{{\|d\mathbf\Phi\|}^4}{3840}$. Substituting~\eqref{eq_delta_ser} into~\eqref{eq_quat_dyn} we obtain Algorithm (\ref{qint}).

\begin{algorithm}[h]
\caption{Efficient quaternion integration}
 \hspace*{\algorithmicindent} \textbf{Input:} De-biased angular increments $({d\phi}_x,{d\phi}_y,{d\phi}_z)$. \\
 \hspace*{\algorithmicindent} \textbf{Output:} Quaternion vector $\boldsymbol{q}=(q_0,q_1,q_2,q_3)$. 
\begin{algorithmic}[1]
\State $D2 = {d\phi}_x *{d\phi}_x + {d\phi}_x *{d\phi}_x+{d\phi}_x *{d\phi}_x$ \Comment \textit{norm of rotation vector squared}
\State $D4=D2*D2$
\State $s = 0.5 - \frac{D2}{48} + \frac{D4}{3840} $
\State $c= - \frac{D2}{8} + \frac{D4}{384} $
\State $s_x = s *{d\phi}_x, \,s_y = s *{d\phi}_y,\, s_z = s *{d\phi}_z $
\State ${dq_0}=c *q_0 -s_x *q_1-s_y *q_2-s_z *q_3  $
\State ${dq_1}=c *q_0 +s_x *q_1+s_z *q_2-s_y *q_3  $
\State ${dq_2}=c *q_0 +s_y *q_1=s_x *q_2-s_z *q_3  $
\State ${dq_3}=c *q_0 -s_z *q_1-s_y *q_2-s_x *q_3  $
\State $q_0 =  q_0 +dq_0$
\State $q_1 =  q_1 +dq_1$
\State $q_2 =  q_2 +dq_2$
\State $q_3 =  q_3 +dq_3$
\end{algorithmic}
\label{qint}
\end{algorithm}

%
\subsection{Normalizing Quaternion Parameters}
According to the quaternion properties, its norm should be always equal to one, which means:
\begin{equation}
    q_0^2+q_1^2+q_2^2+q_3^2= 1\,.
\end{equation}
But unfortunately, the above condition can be violated due to calculation errors or rounding approximations. In order to remove this effect it is necessary to apply a normalization procedure.Since $ q_0^2+q_1^2+q_2^2+q_3^2\approx 1$ then we have:
\begin{equation}
    \Delta =1 - q_0^2+q_1^2+q_2^2+q_3^2=1-\norm{\boldsymbol{q}}^2\,,
\end{equation}
is a very small number. Then normalizing each quaternion parameter by dividing by $\sqrt{1-\Delta}$, and expanding using a Taylor's series formula we obtain:
\begin{equation}
    \hat{q}_{norm}= \frac{q}{\sqrt{1-\Delta}}\approx q(1+\frac{\Delta}{2})=q*0.5(3-\norm{\boldsymbol{q}}^2)\,.
\end{equation}

\begin{algorithm}[h]
\caption{Efficient quaternion normalization}
 \hspace*{\algorithmicindent} \textbf{Input:} Quaternion vector $\boldsymbol{q}=(q_0,q_1,q_2,q_3)$.  \\
 \hspace*{\algorithmicindent} \textbf{Output:} Normalized vector $\boldsymbol{q}=(q_0,q_1,q_2,q_3)$. 
\begin{algorithmic}[1]
\State $q_{00} = q_0 * q_0;$
\State $q_{11} = q_1 * q_1;$
\State $q_{22} = q_2 * q_2;$
\State $q_{33} = q_3 * q_3;$
\State  $qq = q_{00} + q_{11} +q_{22} + q_{33}; $ 
\State $q_{cor}=0.5*(3.0 -qq) $ 
\State ${q_0}=q_0 * q_{cor} $
\State ${q_1}=q_1 * q_{cor} $
\State ${q_2}=q_2  * q_{cor} $
\State ${q_3}=q_3  *q_{cor} $

\end{algorithmic}
\end{algorithm}

%
\subsection{Direction Cosine Matrix Computation}
The quaternions compose a four-element unit vector $(q_0, q_1, q_2, q_3)$ obtained from the quaternion integration step. They can be efficiently used to find the elements of the 3-by-3 Direction Cosine Matrix (DCM). The outputted DCM performs the coordinate transformation of a vector in body axes to a vector in local-level navigation frame axes. 
The following algorithm will be used in the embedded processor to find the 9-elements of the DCM:
\begin{equation}
    C_b^n= \begin{bmatrix}
    c_{11} & c_{12} & c_{13} \\
    c_{21} & c_{22} & c_{23} \\
    c_{31} & c_{32}& c_{33}
    \end{bmatrix}\,.
\end{equation}

\begin{algorithm}[h]

\caption{Efficient direction cosine matrix computation}
\label{alg:alg2}
 \hspace*{\algorithmicindent} \textbf{Input:} Normalized vector $\boldsymbol{q}=(q_0,q_1,q_2,q_3)$.  \\
 \hspace*{\algorithmicindent} \textbf{Output:} Direction-cosine-Matrix $C_b^n$
\begin{algorithmic}[1]
\State $q_{00} = q_0 * q_0;$
\State $q_{11} = q_1 * q_1;$
\State $q_{22} = q_2 * q_2;$
\State $q_{33} = q_3 * q_3;$
\State $q_{01} = q_0 * q_1;$
\State $q_{02} = q_0 * q_2;$
\State $q_{03} = q_0 * q_3;$
\State $q_{12} = q_1 * q_2;$
\State $q_{13} = q_1 * q_3;$
\State $q_{23} = q_2 * q_3;$
\State $c_{11} = q_{00} + q_{11} - q_{22} - q_{33}; $
\State $c_{12} = (q_{12} - q_{03}) * 2; $
\State $c_{13} = (q_{13} + q_{02}) * 2;$
\State $c_{21} = (q_{12} + q_{03}) * 2;$
\State $c_{22} = q_{00} - q_{11} + q_{22} - q_{33};$
\State $c_{23} = (q_{23} - q_{01}) * 2;$
\State $c_{31} = (q_{13} - q_{02}) * 2;$
\State $c_{32} = (q_{23} + q_{01}) * 2;$
\State $c_{33} = q_{00} - q_{11} - q_{22} + q_{33};$
\end{algorithmic}
\end{algorithm}

%
\subsubsection{Roll, Pitch, and Heading Calculation (Euler Angles)}
The Euler angles are three angles introduced by Leonhard Euler to describe the orientation of a rigid body with respect to a fixed coordinate system. Leonard Euler (1707-1783) was one of the giants inn mathematics~\cite{kuipers1999quaternions}. Euler stated and proved a theorem that states that:\par
\hfill\\
\textit{Any two independent orthonormal coordinate frames can be related by a sequence of rotations (not more than three) about coordinate axes, where no two successive rotations may be about the same axis.}\par
\hfill\\
When we say that two independent frames are related, we mean that a sequence of rotations about successive coordinate axes will rotate the first frame into the second. The angle of rotation about a coordinate axis is called an Euler angle. A sequence of such rotations is often called a Euler angle sequence of rotations. The restriction stated in the above theorem that successive axes of rotations be distinct still permits at least 12 Euler angle sequences. The sequence \textbf{xzy} means a rotation about the \textbf{x}-axis, followed by a rotation about the new \textbf{z}-axis, followed by a rotation about the newer \textbf{y}-axis.\par
We are specifically interested in the well-known Euler sequence called the Aerospace sequence. This sequence (\textbf{zyx}) is commonly used in aircraft and aerospace applications. For example, a primary flight instrument used in aircrafts, continuously relates the orientation of the aircraft to the local-level \textit{n}-frame mentioned above.\par
From the \textit{n}-frame,first a rotation through the angle $\psi$ about the \textbf{z}-axis defines the aircraft heading. This is followed by a rotation about the new \textbf{y}-axis through an angle $\theta$ which defines the aircraft pitch. Finally, the aircraft roll angle $\phi$, is a rotation about the newest \textbf{x}-axis. These three Euler angle rotations relate the body coordinate frame of the aircraft to the local-level \textit{n}-frame.

\tcbset{width=\textwidth}

\begin{tcolorbox}[float*=!tbph]
\begin{algorithm}[H]
\caption{Efficient roll, pitch, and heading computation}
\begin{multicols}{2}
 \hspace*{\algorithmicindent} \textbf{Input:} Direction-cosine-Matrix $C_b^n$ \\
 \hspace*{\algorithmicindent} \textbf{Output:} Roll $\phi$, Pitch $\theta$, Heading $\psi$.
\begin{algorithmic}[1]
\\\textbf{\color{Red}Compute Roll}
\State $C1 = c_{32};$
\State $C2 = c_{33};$
\State $ ANGLE=atan(C1,C2);$\footnotemark
\If {C2 $>$ 0}
\State $ROLL = ANGLE;$
\EndIf
\If {$C2 < 0$}
        \If {$C1 > 0$}
        \State $ROLL = ANGLE + \pi;$
        \Else
        \State $ROLL = ANGLE - \pi;$
        \EndIf
\EndIf

\If {$C2 == 0$}
        \If {$C1 \geq 0$}
        \State $ROLL = \frac{\pi}{2};$
        \Else
        \State $ROLL = \frac{-\pi}{2};$
        \EndIf
\EndIf
\If {$ROLL \geq \pi$}
  \State $ROLL =ROLL - \frac{\pi}{2};$
\EndIf
\If {$ROLL \leq -\pi$}
  \State $ROLL =ROLL + \frac{\pi}{2};$
\EndIf
\State \textbf{\color{Red}Compute Pitch}
\State $C1 = c_{31};$
\State $C11 = C1 * C1;$
\If {$C11 \geq 1$}
       \State $C11 = 1;$ \Comment \textit{ should never be more than 1.}
\EndIf
\State $C2 = \sqrt{1-C11};$
\State $ ANGLE=atan(C1,C2);$
\State $PITCH = ANGLE;$
\State \textbf{\color{Red}Compute Heading}
\State $C1 = c_{21};$
\State $C2 = c_{11};$
\If {$C2 == 0$}
        \If {$C1 \geq 0$}
        \State $HEADING = \frac{\pi}{2};$
        \Else
        \State $HEADING = \frac{3\pi}{2};$
        \EndIf
\Else
       \State $ ANGLE=atan(C1,C2);$
        \If {$C2 > 0$}
        \State $HEADING = ANGLE;$
        \Else
                 \If {$C1 \geq 0$}
                 \State $HEADING = ANGLE + \pi;$
                 \Else
                 \State $HEADING = ANGLE - \pi;$
                   \EndIf
        \EndIf
\EndIf
\If {$HEADING < 0$}
       \State $HEADING  =HEADING +\frac{\pi}{2} ;$ 
\EndIf
\end{algorithmic}
\end{multicols}
\end{algorithm}
\footnotetext{For a fast algorithm for calculating the arc-tangent function $atan(\cdot,\cdot)$, see Appendix.}
\end{tcolorbox}

%
\section{Principles of Inertial Navigation}\label{principles}

%
\subsection{Navigation Equations}
It is desirable to formulate the navigation equations in the earth-centered, earth-fixed frame (\textit{e}-frame), since usually the measurements of the GNSS receiver are given in the \textit{e}-frame. But usually we are more comfortable in dealing with \textit{n}-frame coordinates since it is more trivial to deal with the \textit{North-East-Down} directions.  Recall that the coordinate directions of the \textit{n}-frame are defined by the local horizon and by the vertical, and centered on the vehicle center of gravity (cog). Strictly speaking, no horizontal motion takes place in this frame since it is attached and fixed to the vehicle. Therefore,  the navigation equations are not coordinatized in the \textit{n}-frame because no horizontal motion takes place in this frame. Nevertheless, we will  still refer to the \textit{n}-frame coordinatization of the  navigation equations as an Earth-referenced formulation in which the velocity components are transformed along the \textit{n}-frame  coordinate directions. This concept is rarely discussed in the literature and usually is a source of confusion.\par

A vector in the \textit{e}-frame (navigation frame) has coordinates in the \textit{i}-frame (inertial frame) given by
\begin{equation}
    \boldsymbol{x}^i=C_{e}^{i}\boldsymbol{x}^e\,,
    \label{eq_coord_transf}
\end{equation}
where $C_{e}^{i}$ is the transformation matrix from the \textit{e}-frame to the \textit{i}-frame. The time derivative of this matrix is given by~\cite{jekeli2012inertial}
\begin{equation}
    {\dot{C}}_{e}^{i}=C_{e}^{i}{\Omega}_{ie}^{e}\,,
    \label{eq_CNI_dot}
\end{equation}
where ${\Omega}_{ie}^{e}$ denotes a skew-symmetric matrix with elements from $\boldsymbol{\omega}_{ie}^{e}=(\omega_1, \omega_2, \omega_3)$\footnote{$\quad \boldsymbol{\omega}_{ie}^{e}=$ the  angular velocity of the \textit{e}-frame with respect to the \textit{i}-frame, with coordinates in the \textit{e}-frame. Since the three axis of the \textit{e}-frame are aligned with the Earth's spin axis, then $\boldsymbol{\omega}_{ie}^{e}= (0, 0,  \omega_e)$, where $\omega_e$ is the angular rate of the Earth's rotation. } is then given by
\begin{equation}
    \Omega_{ie}^{e} = \left[ \boldsymbol{\omega}_{ie}^{e}\times\right]=\begin{bmatrix}
     0          & -\omega_3  & \omega_2\\
     \omega_3   & 0          & -\omega_1\\
     -\omega_2  & \omega_1   & 0
    \end{bmatrix}\,.
\end{equation}

We also need the second time derivative which from~\eqref{eq_CNI_dot} and the chain rule for differentiation is given by
\begin{equation}
     {\ddot{C}}_{e}^{i}=C_{e}^{i}{\dot{\Omega}}_{ie}^{e}+{C}_{e}^{i}{\Omega}_{ie}^{e}{\Omega}_{ie}^{e}\,.
\end{equation}

Now differentiating~\eqref{eq_coord_transf} twice with respect to time yields
\begin{equation}
    \begin{aligned}
        {\ddot{\boldsymbol{x}}}^{i}&= {\ddot{C}}_{e}^{i}{\boldsymbol{x}}^{e} + 2{\dot{C}}_{e}^{i}{\dot{\boldsymbol{x}}}^{e}+{{C}}_{e}^{i}{\ddot{\boldsymbol{x}}}^{e}\\
             &={{C}}_{e}^{i}{\ddot{\boldsymbol{x}}}^{e}+2{C}_{e}^{i}{\Omega}_{ie}^{e}{\dot{\boldsymbol{x}}}^{e} + {C}_{e}^{i}({\dot{\Omega}}_{ie}^{e}+{\Omega}_{ie}^{e}{\Omega}_{ie}^{e}){\boldsymbol{x}}^{e}\,.
    \end{aligned}
\end{equation}

Solving for ${\ddot{\boldsymbol{x}}}^{n}$ and combining with ${\ddot{\boldsymbol{x}}}^{i}= {\boldsymbol{g}}^{i}+{\boldsymbol{a}}^{i}$ gives the system dynamics for position in the \textit{e}-frame:
\begin{equation}
        {\ddot{\boldsymbol{x}}}^{e}= -2{\Omega}_{ie}^{e}{\dot{\boldsymbol{x}}}^{e} - ({\dot{\Omega}}_{ie}^{e}+{\Omega}_{ie}^{e}{\Omega}_{ie}^{e}){\boldsymbol{x}}^{e}+{\boldsymbol{g}}^{e}+{\boldsymbol{\boldsymbol{a}}}^{e}\,.
  \end{equation}
Since the earth has a constant angular velocity with respect to the inertial frame, then ${\dot{\Omega}}_{ie}^{e}=0$ and we obtain
\begin{equation}\label{eq_xe}
        {\ddot{\boldsymbol{x}}}^{e}= -2{\Omega}_{ie}^{e}{\dot{\boldsymbol{x}}}^{e} - {\Omega}_{ie}^{e}{\Omega}_{ie}^{e}{\boldsymbol{x}}^{e}+{\boldsymbol{g}}^{e}+{\boldsymbol{\boldsymbol{a}}}^{e}\,.
\end{equation}

We can transform the navigation equation above into the \textit{n}-frame  merely by substituting  ${\dot{\boldsymbol{x}}}^{e} = {C}_{n}^{e}{\boldsymbol{v}}^n$ on the right-hand side of~\eqref{eq_xe}
\begin{equation}
    \frac{d}{dt}C_{n}^{e}\boldsymbol{v}^n = C_{n}^{e}(\frac{d}{dt}\boldsymbol{v}^n+{\Omega}_{en}^{n}\boldsymbol{v}^{n})\,,
\end{equation}
and on the right hand side of~\eqref{eq_xe} we use the formula ${\Omega}_{ie}^{n}=C_{e}^{n}{\Omega}_{ien}^{e}C_{n}^{e}$, and the result is:
\begin{equation}\label{eq_rtyyy}
    \frac{d}{dt}\boldsymbol{v}^n=\boldsymbol{{a}}^{n}- (2{{\Omega}}_{ie}^{n}+{\Omega}_{en}^{n}){\boldsymbol{v}}^{n}+\boldsymbol{{g}}^{n} -C_{e}^{n}{\Omega}_{ie}^{e}{\Omega}_{ie}^{e}\boldsymbol{x}^{e}\,.
\end{equation}
The last two terms are, respectively, the gravitational vector and the centrifugal acceleration due to the Earth's rotation, coordinatized in the \textit{n}-frame. Together they define the gravity vector:
\begin{equation}
   {\Bar{\boldsymbol{g}}}^{n} =\boldsymbol{{g}}^{n} -C_{e}^{n}{\Omega}_{ie}^{e}{\Omega}_{ie}^{e}\boldsymbol{x}^{e}\,.
    \label{eq_nframe_dyn}
\end{equation}

The distinction between the terms gravitation and gravity, refers to the difference between the acceleration due to mass attraction, alone, and the total acceleration, gravitational and centrifugal, that is measured at a fixed point on the rotating earth. Gravity has a direction  that coincides with the direction of a plumb line  at any given point in space. The direction of a plumb line coincides with the direction of a string to which a freely suspended weight, or plumb bob, is attached.\par
Finally, by writing ${\Omega}_{en}^{n}={\Omega}_{in}^{n}+{\Omega}_{ei}^{n}={\Omega}_{in}^{n}-{\Omega}_{ie}^{n}$\footnote{Relative angular velocities can be added component-wise and they satisfy commutativity.} and by manipulating the subscripts, we also obtain

\begin{equation}\label{eq_rty}
2{{\Omega}}_{ie}^{n}+{\Omega}_{en}^{n}={{\Omega}}_{in}^{n}+{\Omega}_{ie}^{n}\,.
\end{equation}

Substituting~\eqref{eq_rty} and~\eqref{eq_nframe_dyn} into~\eqref{eq_rtyyy}, the desired form of the  \textit{n}-frame navigation equations becomes:
\begin{equation}
    \frac{d}{dt}{\boldsymbol{v}}^n={\boldsymbol{a}}^{n}- ({\Omega}_{in}^{n}+{\Omega}_{ie}^{n}){\boldsymbol{v}}^{n}+{\Bar{\boldsymbol{g}}}^{n} \,.
    \label{eq_nframe_dyn2}
\end{equation}
The components of the Earth-referenced velocity, ${\boldsymbol{v}}^n$, the sensed acceleration, ${\boldsymbol{a}}^{n}$, and the gravity vector, ${\Bar{\boldsymbol{g}}}^{n}$, can be described by their north, east, and down components in the \textit{n}-frame as follows:
\begin{equation}
    {\boldsymbol{v}}^n=\begin{bmatrix}
    {v}_n\\
    {v}_e\\
    {v}_d\\
    \end{bmatrix}\,,\quad
    {\boldsymbol{a}}^{n}=\begin{bmatrix}
    {a}_n\\
    {a}_e\\
    {a}_d\\
     \end{bmatrix}\,,\quad
    {\Bar{\boldsymbol{g}}}^{n}=\begin{bmatrix}
    {\Bar{g}}_n\\
    {\Bar{g}}_e\\
    {\Bar{g}}_d\\ 
    \end{bmatrix}\,.
    \label{eq_compo}
\end{equation}

%
\subsection{Error Dynamic Equations in the \textit{n}-Frame}
The equations of motion  depict the time development of the user's position, speed, and attitude under perfect conditions.
The way in which errors proliferate in an INS can be computed by applying a first-order Taylor series development (linearization), or perturbation investigation, to the equations of motion derived in the previous sub-section. The coordinatization of the error dynamics in the \textit{n}-frame represents the traditional and most intuitive scheme of analyzing INS errors.\\

\begin{figure}[ht]
    \centering
      \begin{subfigure}{0.45\linewidth}
    \centering
    \includegraphics[width=\linewidth]{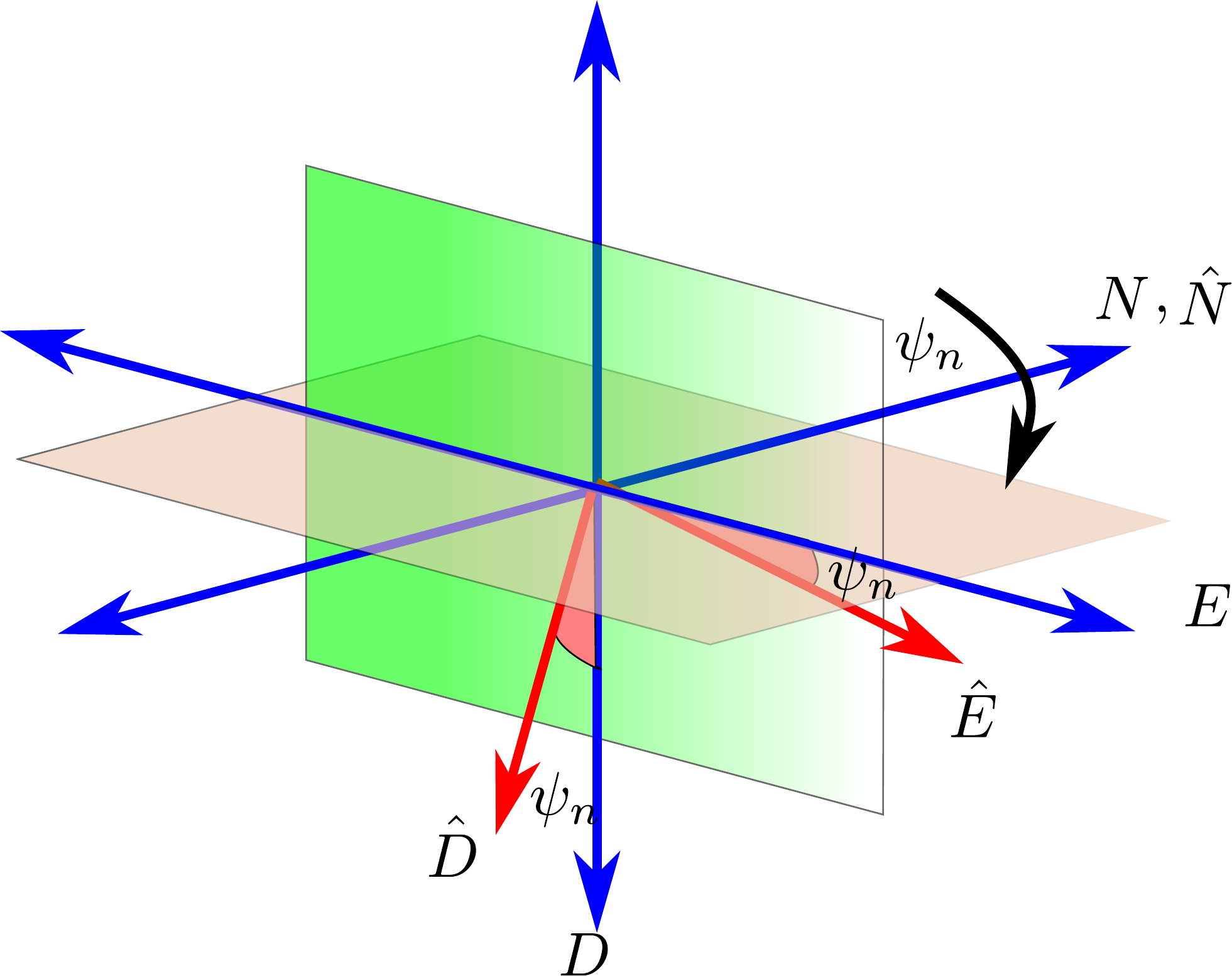}
    \caption{Tilt error in the N-direction}
    \label{fig_first}
    \end{subfigure}
    \begin{subfigure}{0.45\linewidth}
    \centering
    \includegraphics[width=\linewidth]{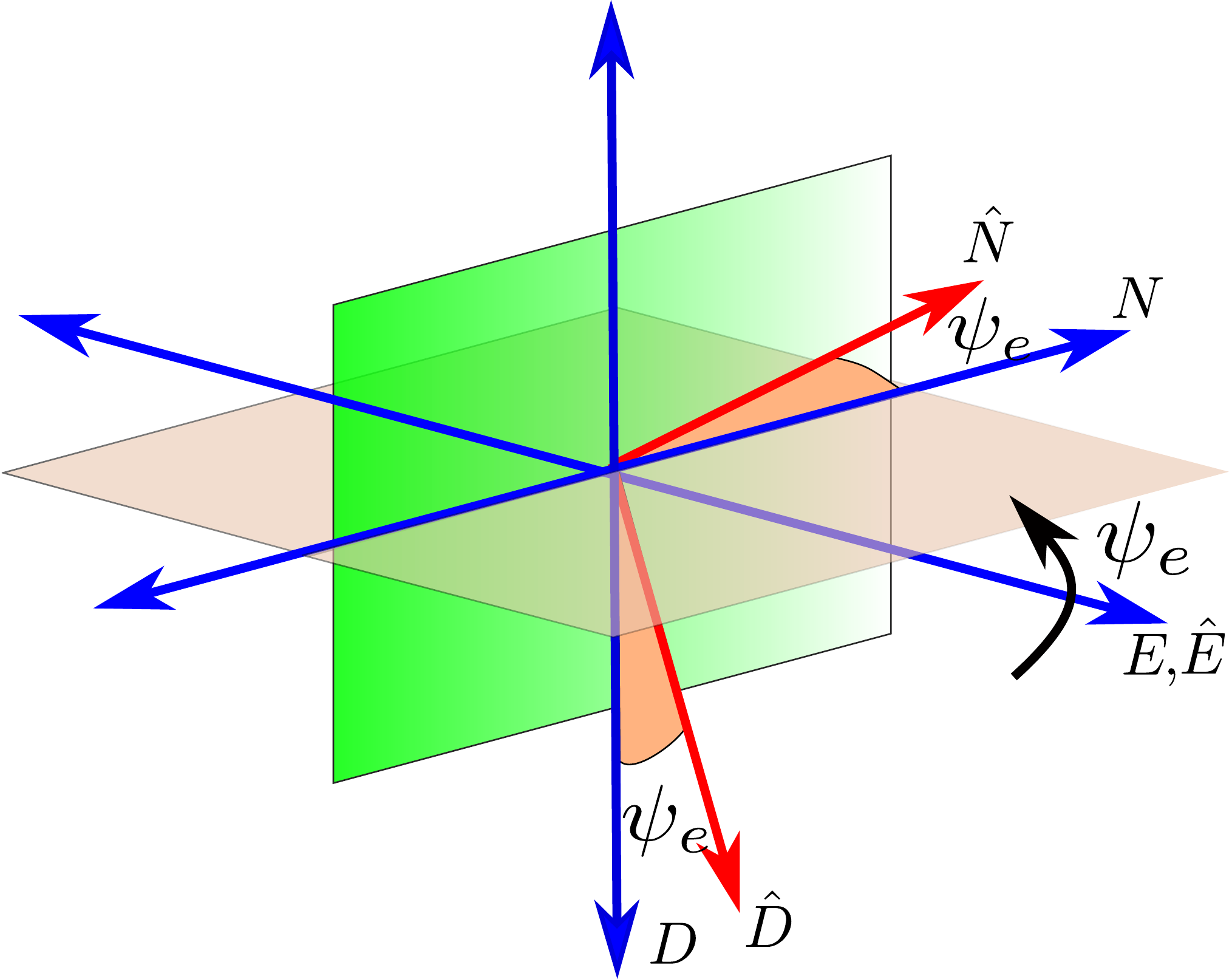}
    \caption{Tilt error in the E-direction}
    \label{fig_second}
    \end{subfigure}
     \begin{subfigure}{0.45\linewidth}
    \centering
    \includegraphics[width=\linewidth]{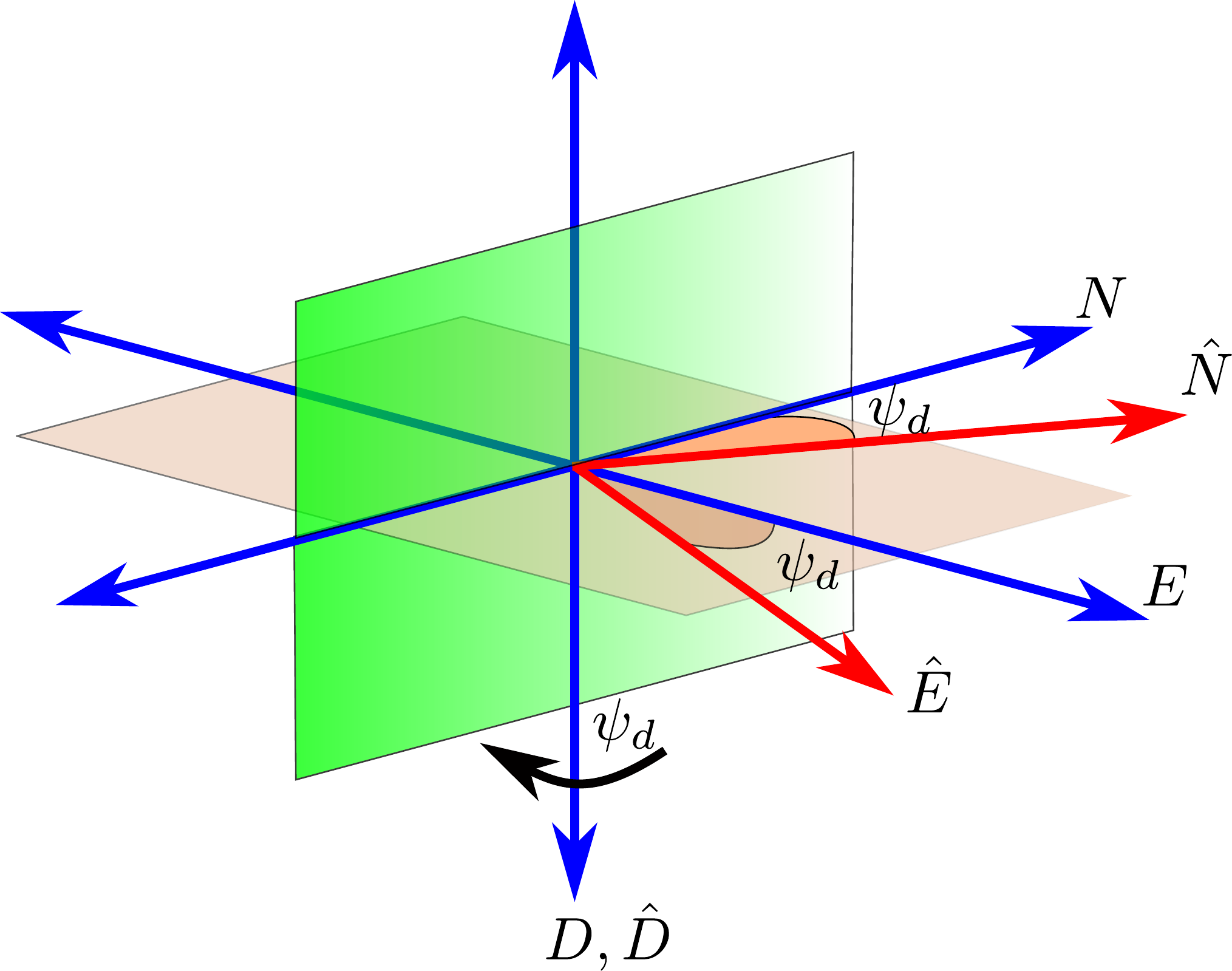}
    \caption{Tilt error in the D-direction}
    \label{fig_third}
    \end{subfigure}
    \caption{The platform frame \textit{$\hat{N}\hat{E}\hat{D}$} is a virtual frame that is slightly misaligned from the true navigation frame. It is mainly created for the derivation of the error equations. It is only recognized by the INS on-board computer and it results from the erroneous gyroscope sensors that are integrated to give this false navigation frame.}
    \label{fig_mul_images}
\end{figure}

%
\subsubsection{Orientation Error Dynamics in the \textit{n}-frame}
The perturbation $\delta \boldsymbol{a}^n$ is interpreted as the error in the expression in the \textit{n}-frame of the sensed acceleration. It represents not only accelerometer errors but also orientation errors that are committed when transforming sensed accelerations from the sensor frame (\textit{s}-frame) to the \textit{n}-frame. Taking differentials of the relation $\boldsymbol{a}^n=C_{s}^{n}\boldsymbol{a}^s$, we obtain:
\begin{equation}\label{eq_delta_an}
    \delta \boldsymbol{a}^n=\delta C_{s}^{n}{\boldsymbol{a}}^{s} + C_{s}^{n} \delta \boldsymbol{a}^s\,.
\end{equation}

The differential $\delta C_{s}^{n}$ is caused by errors in the orientation of the \textit{s}-frame with respect to the {n}-frame. It is convenient to represent $\delta C_{s}^{n}$ in terms of small error angles, one for each of the \textit{n}-frame axes: $\boldsymbol{\psi^n}=(\psi_n, \psi_e, \psi_d)^T$. This can be represented in the equivalent form of a skew-symmetric matrix:
\begin{equation}
    \boldsymbol{\Psi}^n=\left[\begin{array}{rrr}
         {0}&-{\psi_d}&{\psi_e}\\
         {\psi_d}&{0}&-{\psi_n}\\ 
         -{\psi_e}&{\psi_n}&{0}\\
          \end{array} \right]\,.
    \label{eq_tiltskew}
\end{equation}

Since $\psi_n, \psi_e$ and $\psi_d$ represent small  angle rotation errors, the transformation matrix from the true \textit{n}-frame to the erroneously computed \textit{n}-frame (inside the INS computer),  can be written as $I-\boldsymbol{\Psi}^n$. This can be easily achieved by substituting the small angle errors (see Fig.~\ref{fig_mul_images})
inside the DCM  equation~\eqref{eq_CDCM} on page \pageref{eq_CDCM}, and approximating it to first order. Therefore,
the computed transformation may be represented as a sequence of two transformations comprising first the true transformation  from  the  body frame (\textit{n}-frame or \textit{s}-frame )  to  the erroneously computed \textit{n}-frame followed by another  transformation  from the true \textit{n}-frame to the erroneously computed \textit{n}-frame:
\begin{equation}
     \hat{C}_{s}^{n}=(I-\boldsymbol{\Psi}^n) C_{s}^{n} \,.
\end{equation}

It is now clear that:
\begin{equation}\label{eq_deltaCsn}
  \delta{C}_{s}^{n}  =\hat{C}_{s}^{n}-C_{s}^{n} =-\boldsymbol{\Psi}^n C_{s}^{n} \,.
\end{equation}

Substituting~\eqref{eq_deltaCsn} in~\eqref{eq_delta_an}, we obtain
\begin{equation}\label{eq_delta_anmod}
\begin{split}
    \delta \boldsymbol{a}^n&= C_{s}^{n} \delta \boldsymbol{a}^s -\boldsymbol{\Psi}^n C_{s}^{n}{\boldsymbol{a}}^{s}\\
      &=C_{s}^{n} \delta \boldsymbol{a}^s +\boldsymbol{a}^{n}\times\boldsymbol{\psi}^n\,.
    \end{split}
\end{equation}

We now establish the dynamic behavior of the error angles ${\boldsymbol{a}}^{s}$ in the form of a differential equation. Taking the differential of $\dot{C}_{s}^{n}={C}_{s}^{n} \Omega_{ns}^s $
\begin{equation}\label{eq_delta_anmo}
\begin{split}
    \delta \dot{C}_{s}^{n}&=\delta({C}_{s}^{n} \Omega_{ns}^s)\\
    &=\delta{C}_{s}^{n} \Omega_{ns}^s+ {C}_{s}^{n} \delta\Omega_{ns}^s\,,
    \end{split}
\end{equation}
where the perturbation in angular rate, $\delta\Omega_{ns}^s$, is interpreted as the error in the corresponding computed value, denoted by $\hat{\Omega}_{ns}^s$:
\begin{equation}\label{eq_ggggg}
     \delta\Omega_{ns}^s=\hat{\Omega}_{ns}^s - {\Omega}_{ns}^s\,.
\end{equation}

Differentiating the second line of~\eqref{eq_delta_anmod} with respect to time and setting the result equal to the right side of~\eqref{eq_delta_anmo}, we get
\begin{equation}\label{eq_fffggg}
   - \dot{\boldsymbol{\Psi}}^n C_{s}^{n} -\boldsymbol{\Psi}^n C_{s}^{n}\Omega_{ns}^s=\delta{C}_{s}^{n} \Omega_{ns}^s+ {C}_{s}^{n} \delta\Omega_{ns}^s\,.
\end{equation}
Substituting for $\boldsymbol{\Psi}^n C_{s}^{n}$ and solving for $\dot{\boldsymbol{\Psi}}^n$ yields
\begin{equation}\label{eq_fffgggkkk}
   \dot{\boldsymbol{\Psi}}^n = -{C}_{s}^{n} \delta\Omega_{ns}^s {C}_{n}^{s}\,,
\end{equation}
in terms of vectors, it is easily verified that this is equivalent to 
\begin{equation}\label{eq_fffgggkkkjj}
   \dot{\boldsymbol{\psi}}^n = -{C}_{s}^{n} \delta\boldsymbol\omega_{ns}^s\,, 
\end{equation}
where $\delta\boldsymbol\omega_{ns}^s$ is the error in the rotation rate of the \textit{s}-frame with respect to the \textit{n}-frame. For small vehicle velocities, the  angular velocity of the navigation frame  is negligible\footnote{From pure geometric  observations of Fig.~\ref{fig_cne} in Box~\ref{box:nav_frame_rel} on page \pageref{box:nav_frame_rel}, it is evident that the vehicle velocity in the north and east direction are related to latitude and longitude rate respectively  through, $\dot\phi={\boldsymbol{v}_n}/{R}$ and $\dot\lambda= \boldsymbol{v}_e / R\cos\phi$ where $R\approx\unit[6370]{km}$  is the radius of the Earth.\label{foot1}}, consequently,  $\boldsymbol\omega_{ns}^s\approx \boldsymbol\omega_{is}^s $, where $\boldsymbol\omega_{is}^s$ is the angular velocity vector delivered by the gyroscope\footnote{Gyroscopes deliver vehicle  angular velocities  with respect to the inertial frame, and since in a strapdown mechanization these inertial sensors are fixed to the vehicle body, their readings are referenced to  this frame. (Usually the sensor and body frame are considered aligned to each other with a probable offset between their origins).}.

Equation~\eqref{eq_fffgggkkkjj} can be discovered directly with no need for any rigorous derivation, simply by recognizing that, the errors in the orientation angular rates, $\dot\psi_n, \dot\psi_e$, and $\dot\psi_d$ are nothing but the transformations of the  gyroscope  angular rate errors, $\delta\omega_{Gx}^s, \delta\omega_{Gy}^s$, and  $\delta\omega_{Gz}^s$\footnote{The Gyro bias vector adapted in this manuscript is written as $\delta\boldsymbol{\omega}_G = (\delta\omega_{Gx}^s,\delta\omega_{Gy}^s ,\delta\omega_{Gz}^s )$ (notation adopted  from ~\cite{petkov2010stochastic}). }, (gyroscope biases in practice) from the body frame to the navigation frame, which when expanded can be written in the form (for convenience the body frame and the sensor frame are considered coincident):
\begin{equation}\label{eq_disc1}
\begin{bmatrix}
 \dot{{\psi}}_n \\
  \dot{{\psi}}_e\\
   \dot{{\psi}}_d
\end{bmatrix}
=\begin{bmatrix}
    c_{11} & c_{12} & c_{13} \\
    c_{21} & c_{22} & c_{23} \\
    c_{31} & c_{32}& c_{33}
    \end{bmatrix}\begin{bmatrix}
    \delta\omega_{Gx}^s \\
    \delta\omega_{Gy}^s \\
    \delta\omega_{Gz}^s
    \end{bmatrix}\,.
\end{equation}
In~\eqref{eq_disc1} the $3\times3$ matrix can be determined by using Algorithm 
\ref{alg:alg2}, and the $\delta\omega$ terms represent the gyroscope biases along the three axes \textbf{x}, \textbf{y}, and \textbf{z} of the body frame.\par
Since we are interested in implementing our algorithms on an embedded processor we need to discretize~\eqref{eq_disc1}. With the help of BOX~\ref{box:nav_frame_rel} on page \pageref{box:nav_frame_rel}, we obtain the following discretized version:
\begin{equation}\label{eq_disc2}
\begin{aligned}
&\begin{bmatrix}
 {{\psi}}_n\langle k+1\rangle \\
  {{\psi}}_e\langle k+1\rangle\\
   {{\psi}}_d\langle k+1\rangle
\end{bmatrix}=\begin{bmatrix}
 {{\psi}}_n\langle k\rangle \\
  {{\psi}}_e\langle k\rangle\\
   {{\psi}}_d\langle k\rangle
\end{bmatrix}+\\
&\qquad\begin{bmatrix}
    c_{11}\langle k\rangle dT & c_{12}\langle k)dT & c_{13}\langle k\rangle dT \\
    c_{21}\langle k\rangle dT & c_{22}\langle k)dT & c_{23}\langle k\rangle dT \\
    c_{31}\langle k\rangle dT & c_{32}\langle k)dT& c_{33}\langle k\rangle dT
    \end{bmatrix}\begin{bmatrix}
    \delta\omega_{Gx}^s\langle k\rangle \\
    \delta\omega_{Gy}^s\langle k\rangle \\
    \delta\omega_{Gz}^s\langle k\rangle
    \end{bmatrix}\,.
    \end{aligned}
\end{equation}

\tcbset{width=\columnwidth}
\begin{mybox}[label={state_transition}]{State-Space Discretization}
  
A state-space representation of a general continuous time system is written as $\dot{\boldsymbol{x}}(t)=A(t){\boldsymbol{x}}(t)+B(t)\omega(t)$. At a certain time epoch $t=t_k$ it gives~\cite{salychev2012mems}: 
\begin{equation}\label{eq_bla}
\dot{\boldsymbol{x}}(t_k)=A(t_k){\boldsymbol{x}}(t_k)+B(t_k)\omega(t_k)    
\end{equation}
For a short sampling period, $dT=t_{k+1}-t_{k}$, one can write:
\begin{equation}\label{eq_statedis}
   \dot{\boldsymbol{x}}(t_k)\approx\frac{\boldsymbol{x}\langle k+1 \rangle-\boldsymbol{x}\langle k \rangle }{dT}
    \end{equation}
    
  where $\boldsymbol{x}\langle k+1 \rangle={\boldsymbol{x}}(t_{k+1})$ and $\boldsymbol{x}\langle k \rangle={\boldsymbol{x}}(t_{k})$. 
  Substituting~\eqref{eq_statedis} in~\eqref{eq_bla}, one can get:
  \begin{equation}
      \begin{split}
        \boldsymbol{x}\langle k+1 \rangle&= (I+A(t_k)dT) \boldsymbol{x}\langle k+1 \rangle + B(t_k)\boldsymbol{w}\langle k \rangle dT\\
        &= \Phi(t_{k+1},t_{k}) \boldsymbol{x}\langle k \rangle + G(t_{k+1},t_{k})\boldsymbol{w}\langle k \rangle
      \end{split}
  \end{equation}
  where $\Phi(t_{k+1},t_{k})$ is a state transition matrix that propagates the system state $\boldsymbol{x}_{k}$ one time step; $G(t_{k+1},t_{k})$ is an input matrix that plays the role of coloring the input system noise $\boldsymbol{w}_k$, which is usually white with zero mean. \par
  \textbf{Nb:} This discretization is only a first order approximation, but this is compromised by the fact that the uncertainties in the discrete model will be hidden in the coloured white noise vector of the system model equation.

\end{mybox}

%
\subsubsection{Velocity Error Dynamics in the \textit{n}-Frame}
The objective in the \textit{n}-frame coordinatization is to formulate the error dynamics with respect to the geodetic coordinates ($\phi, \lambda,h$). We write the perturbation from the compact form of the velocity navigation equation~\eqref{eq_nframe_dyn2}.
\begin{equation}\label{eq_erdyn}
\begin{aligned}
    \frac{d}{dt}\delta \boldsymbol{v}^n=-\delta({{\Omega}}_{in}^{n}+{{\Omega}}_{ie}^{n}){\boldsymbol{v}}^{n}&- ({{\Omega}}_{in}^{n}+{{\Omega}}_{ie}^{n})\delta{\boldsymbol{v}}^{n}\\
    &+ \delta \boldsymbol{a}^n +{\Bar{\boldsymbol{\Gamma}}}^n \delta \boldsymbol{p}^n +\delta{\Bar{\boldsymbol{g}}}^n\,.
    \end{aligned}
\end{equation}

Of note in the above equation are the errors in the computation of the local gravity vector, ${\Bar{\boldsymbol{g}}}$.
This vector is not a constant, and varies as a function of location as identified in the matrix of gravity gradients, ${\Bar{\boldsymbol{\Gamma}}}^n={\partial{\Bar{\boldsymbol{g}}}^n}/{\partial{{\boldsymbol{p}}}^n}$. Position errors, $\delta{{\boldsymbol{p}}}^n$, 
lead to errors in ${\Bar{\boldsymbol{g}}}$ which, in turn, lead to velocity errors. Using~\eqref{eq_wen} and~\eqref{eq_win} in Box~\ref{box:nav_frame_rel} on page \pageref{box:nav_frame_rel} it can be easily shown that:

\begin{strip}
\begin{equation}\label{eq_appro}
   \delta({{\Omega}}_{in}^{n}+{{\Omega}}_{ie}^{n}) =
\begin{bmatrix} 
0             &\delta\dot\lambda\sin\phi+(\dot\lambda+2\omega_e)\cos\phi\delta\phi& -\delta\dot\phi \\ 
-\delta\dot\lambda\sin\phi-(\dot\lambda+2\omega_e)\cos\phi\delta\phi & 0        &-\delta\dot\lambda\cos\phi+(\dot\lambda+2\omega_e)\sin\phi\delta\phi\\
\delta\dot\phi&\delta\dot\lambda\cos\phi-(\dot\lambda+2\omega_e)\sin\phi\delta\phi&0
\end{bmatrix}\,.
\end{equation}
\end{strip}
Since our intention is to give a hands-on experience for the reader of this manuscript, it is instructive to provide the  approximations usually applied in commercial integrated navigation systems to the error equations. Typically, for general applications where \textit{n}-frame velocities don't exceed, say, $\unitfrac[120]{m}{s}$, the term $\delta({{\Omega}}_{in}^{n}+{{\Omega}}_{ie}^{n}){\boldsymbol{v}}^{n}$ (in view~\eqref{eq_appro} is lower than $\unitfrac[10^{-5}]{m}{s^2}$. This can be neglected in case of low-cost MEMS are being used in the INS, since the the  acceleration errors, $\delta\boldsymbol{a}^n$, due to accelerometer biases and gyro drift are much higher. It is important to stress that gyro drifts lead to an erroneous transformation of sensed acceleration from body to navigation frame as shown later in this section. For small navigational velocities, the angular rate of the vehicle $\Omega_{en}^n$, is relatively much smaller  than the angular rate of the Earth $\Omega_{ie}^n$, which is already lower than the noise level found in typical MEMS sensors. Thus for an elementary analysis we may consider the second term on the right hand side  to be zero in the error dynamics of equation~\eqref{eq_erdyn}. Neglecting the gravity related\footnote{It can be showed that for relatively short time of navigation ($ \unit[1-2]{hr})$ the gravity gradient terms and the errors in computing the gravity vector are negligible.} terms in~\eqref{eq_erdyn} we obtain:
\begin{equation}\label{eq_erdynapp}
\begin{aligned}
    \frac{d}{dt}\delta \boldsymbol{v}^n=
     \delta \boldsymbol{a}^n \,.
    \end{aligned}
\end{equation}

Taking the second part of~\eqref{eq_delta_anmod} and using $\boldsymbol{a}^n \approx (a_n, a_e, -\Bar{{g}})$ since in case of low vehicle velocity in the navigation frame the accelerometer in the down direction is overshadowed by the gravity vector, $a_d\approx -\bar g$, we obtain:
\begin{equation}\label{eq_delta_anmodapp}
\begin{split}
    \frac{d}{dt}\delta \boldsymbol{v}^n&= C_{s}^{n} \delta \boldsymbol{a}^s +\boldsymbol{a}^{n}\times\boldsymbol{\psi}^n\\
      &=\begin{bmatrix}
      \delta{a}_{An}\\
     \delta{a}_{Ae}\\
      \delta{a}_{Ad}
      \end{bmatrix} +\begin{bmatrix}
     a_n\\
      a_e\\
      -\Bar{{g}}
      \end{bmatrix}
      \times\begin{bmatrix}
 {{\psi}}_n \\
  {{\psi}}_e\\
   {{\psi}}_d
\end{bmatrix}\\
&=\begin{bmatrix}
      \delta{a}_{An}\\
     \delta{a}_{Ae}\\
      \delta{a}_{Ad}
      \end{bmatrix} +\begin{bmatrix}
     a_e{{\psi}}_d+\Bar{g}{{\psi}}_e\\
      -a_n{{\psi}}_d-\Bar{g}{{\psi}}_n\\
      -a_n{{\psi}}_e-a_e{{\psi}}_n
      \end{bmatrix}\,.
    \end{split}
\end{equation}
where we have used the notation $ \delta \boldsymbol{a}_A=C_{s}^{n} \delta \boldsymbol{a}^s$ (notation adopted from~\cite{2004applied}) to identify the accelerometer error vector produced in the body frame, but projected on the navigation frame.\par
\begin{figure}[ht]
    \centering
      \begin{subfigure}{0.45\linewidth}
    \centering
    \includegraphics[width=\linewidth]{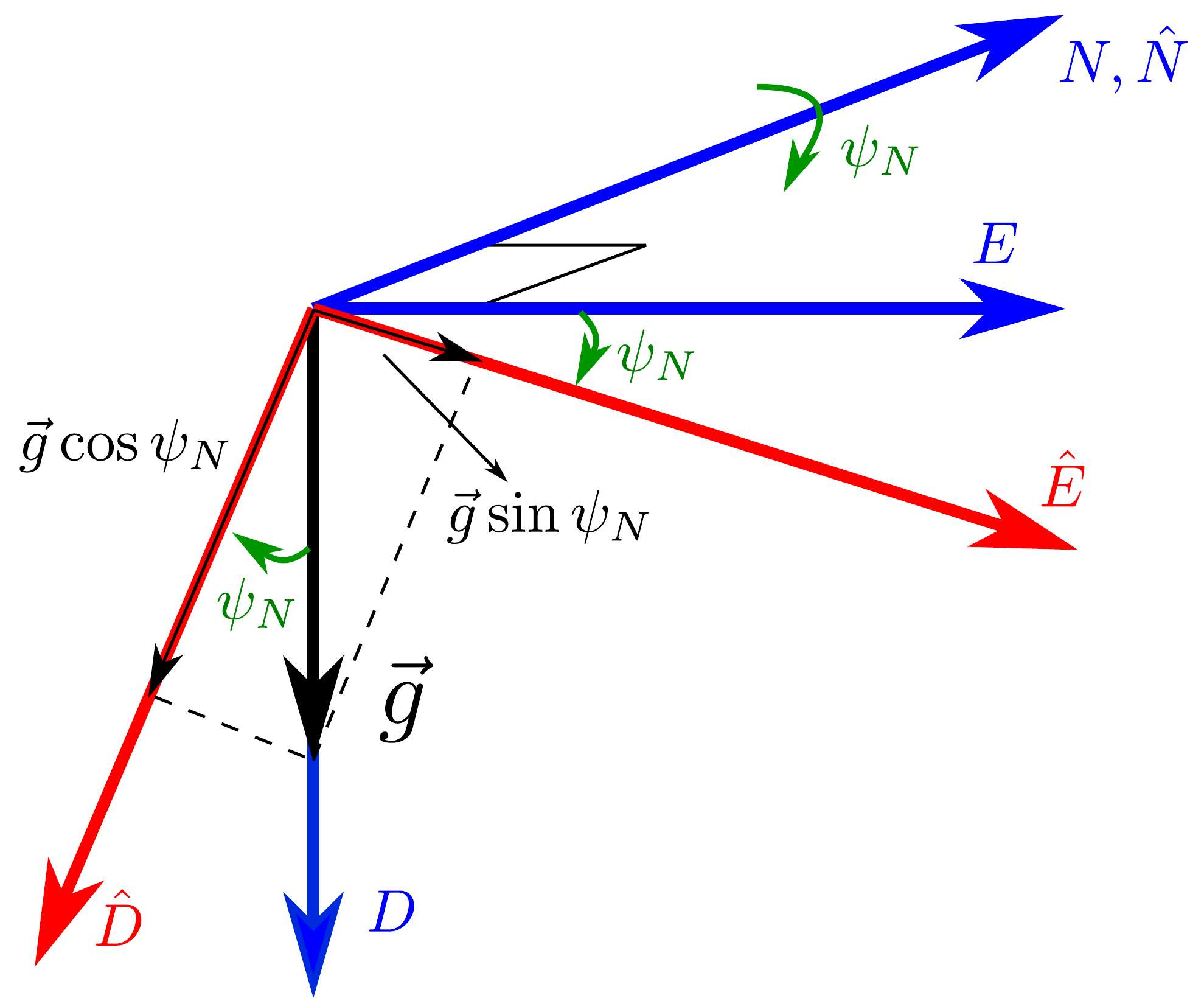}
    \caption{Due to  tilt error around the N-direction, the East pointing accelerometer picks up a reaction to the pull of gravity. If the platform was level, this component would be null.}
    \label{fig_secon}
    \end{subfigure}
    \begin{subfigure}{0.45\linewidth}
    \centering
    \includegraphics[width=\linewidth]{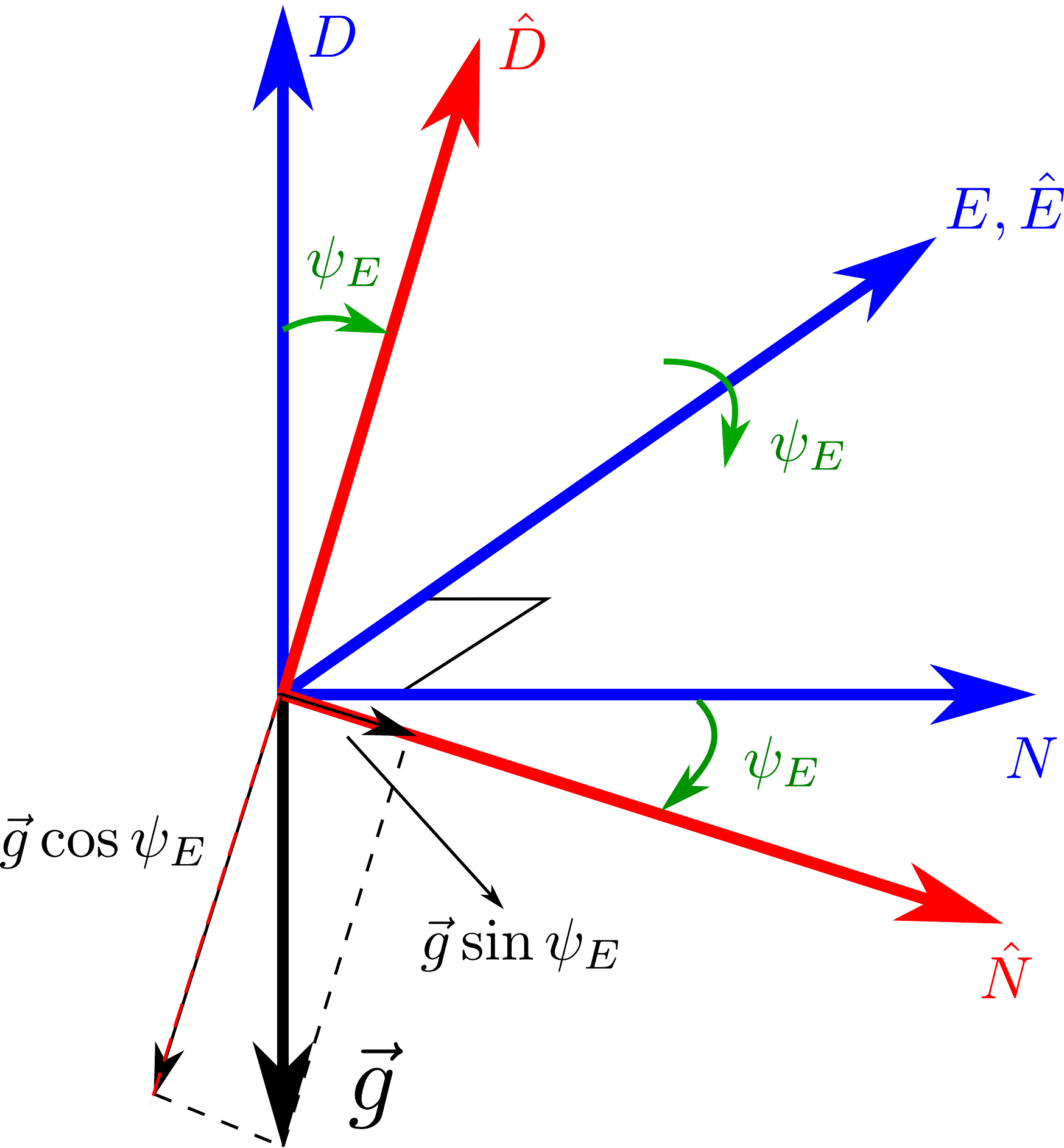}
    \caption{The tilt error around the East pointing axis also causes erroneous readings on other sensors.}
    \label{fig_thir}
    \end{subfigure}
         \caption{Attitude measurement error produces very predictable errors in velocity estimates .  If you can measure velocity and position errors (say, with GPS), then you can figure out your attitude error using a combination of GPS and inertial sensors.}
    \label{fig_mul_imagett}
\end{figure}

It is instructive to prove the  velocity error dynamics, simply by relying on insights into Fig.~\ref{fig_mul_imagett}. We shall  re-prove the error dynamics for the first component in~\eqref{eq_delta_anmodapp}, $\delta v_n$. The interested reader is invited to prove the error dynamic equations of the  other terms.
From  a geometrical viewpoint of Fig.~\ref{fig_secon}, we have:
\begin{equation}\label{eq_ff1}
    \hat{a}_n =a_n \cos\psi_e -a_d\sin\psi_e\,.
\end{equation}
For $\psi_e\approx 0$ and $a_d \approx -\bar g$,~\eqref{eq_ff1} becomes:
\begin{equation}\label{eq_ff2}
    \hat{a}_n =a_n + \bar g\psi_e\,,
\end{equation}
which implies that:
\begin{equation}\label{eq_ff3}
    \hat{a}_n -a_n=\delta a_n= \bar g\psi_e\,.
\end{equation}
Similarly, looking at Fig.~\ref{fig_thir}, we have:
\begin{equation}\label{eq_ff4}
    \hat{a}_n =a_n \cos\psi_d +a_e\sin\psi_d
\end{equation}

For $\psi_d\approx 0$ ,~\eqref{eq_ff4} becomes to first order:
\begin{equation}\label{eq_ff5}
    \hat{a}_n =a_n + a_e\psi_d\,,
\end{equation}
which implies that:
\begin{equation}\label{eq_ff6}
    \hat{a}_n -a_n=\delta a_n= a_e\psi_d\,.
\end{equation}

So the total error in $a_n$, is the superposition of the terms produced by $\psi_d$ and $\psi_e$, thus we have:
\begin{equation}\label{eq_ff7}
    \delta a_n= a_e\psi_d + \bar g\psi_e\,.
\end{equation}

Adding to~\eqref{eq_ff7} the error in acceleration caused by the bias term $\delta a_{An}$ we obtain:
\begin{equation}\label{eq_ff8}
    \delta a_n=\frac{d}{dt}\delta \boldsymbol{v}^n= \delta a_{An} +a_e\psi_d + \bar g\psi_e \,,
\end{equation}
which is nothing but the first row in~\eqref{eq_delta_anmodapp}\footnote{Due to continuity of the $v_n $ term, we have  $\delta a_n= \delta (\frac{d}{dt} v_n)= \frac{d}{dt}\delta v_n$}.

Discretizing~\eqref{eq_delta_anmodapp} with the aid of BOX~\ref{state_transition}, we can easily obtain the following discrete velocity error equations
\begin{equation}\label{eq_delta_anmoddisc}
\begin{split}
    \begin{bmatrix}
      \delta{v}_{n}\langle k+1 \rangle\\
     \delta{v}_{e}\langle k+1 \rangle\\
      \delta{v}_{d}\langle k+1 \rangle
      \end{bmatrix}&=   \begin{bmatrix}
      \delta{v}_{n}\langle k \rangle\\
     \delta{v}_{e}\langle k \rangle\\
      \delta{v}_{d}\langle k \rangle
      \end{bmatrix}  
     +dT\begin{bmatrix}
      \delta{a}_{An}\langle k \rangle\\
     \delta{a}_{Ae}\langle k \rangle\\
      \delta{a}_{Ad}\langle k \rangle
      \end{bmatrix}\\
      &+dT\begin{bmatrix}
    ( a_e\langle k \rangle{{\psi}}_d\langle k \rangle+\Bar{g}{{\psi}}_e\langle k \rangle)\\
     ( -a_n\langle k \rangle{{\psi}}_d\langle k \rangle-\Bar{g}{{\psi}}_n\langle k \rangle)\\
      (-a_n\langle k \rangle{{\psi}}_e\langle k \rangle-a_e\langle k \rangle{{\psi}}_n\langle k \rangle)
      \end{bmatrix}\,.
    \end{split}
\end{equation}

%
\subsubsection{Position Error Dynamics in the \textit{n}-Frame}
Referring to footnote~\ref{foot1} on page  \pageref{foot1} we can write:
\begin{equation}\label{eq_latlon1}
\begin{split}
        \delta\dot\phi&=\frac{\delta v_n}{R}\,,\\
        \delta\dot\lambda&= \frac{\delta v_e}{R\cos\phi}+\frac{v_e}{R}\frac{\sin\phi}{\cos^2\phi}\delta\phi\,,
    \end{split}
\end{equation}
but since $\dot\lambda= {\delta v_e}/{R\cos\phi}$
and $\dot\lambda\approx 0$ (for small vehicle velocity), then~\eqref{eq_latlon1} becomes:
\begin{equation}\label{eq_latlon2}
\begin{split}
        \delta\dot\phi&=\frac{\delta v_n}{R}\,,\\
        \delta\dot\lambda&= \frac{\delta v_e}{R\cos\phi}\,.
    \end{split}
\end{equation}
The radius of the Earth, $R$, is taken to be constant neglecting the ellipsoidal nature of the Earth.\par
We should also include altitude $h$ in our position error dynamic equations. To do so we can simply write\footnote{The negative sign in error dynamics for the altitude channel is due to the fact that, it is traditional in GPS-INS fusion systems to take altitude $h$ and the down axis in opposite directions.}:
\begin{equation}
    \delta\dot{h}=-\delta v_d\,.
\end{equation}
In order to discretize~\eqref{eq_latlon2} we approximate the derivatives by a finite difference, resulting in:
\begin{equation}\label{eq_latlon3}
\begin{split}
        \delta\phi\langle k+1 \rangle &=\delta\phi\langle k \rangle +dT\frac{\delta v_n\langle k+1 \rangle}{R}\,,\\
        \delta\lambda\langle k+1 \rangle &=\delta\lambda\langle k \rangle +dT\frac{\delta v_e\langle k \rangle}{R\cos\phi\langle k \rangle}\,,\\
        \delta h \langle k+1 \rangle &=\delta h\langle k \rangle -dT{\delta v_d\langle k \rangle}\,.
       \end{split}
\end{equation}

%
\section{Kalman Filter}\label{Kalman}
The Kalman filter is an estimation strategy, instead of being a filter. The fundamental strategy was designed by R. E. Kalman in 1960~\cite{kalman1960new}, and has been improved  further by various researchers  since. The filter refreshes the estimates  of the state vector which is persistently changing. These estimates are then updated using a set of measurements which are subject to noise~\cite{groves2015principles}. The measurements should be written in terms of the parameters estimated, yet the
measurements at a given time need not contain adequate information to uniquely decide the values of the state vector at the time. This is related to the concept of observability of the system. It closely mimics the case of solving a set of equations where the number of unknown variables exceeds the number of equations.\par
The Kalman filter utilizes information of statistical properties of the system in order to get ideal estimates of the data available. It maintains a set of uncertainties about the estimates that is carried from one iteration to another. It also carries a measure of correlation between the errors in the estimates of the states from iteration to iteration.

The Kalman filter is an efficient algorithm from computing point of view, since it is a recursive algorithm that only processes the latest measurements and forgets the old  ones. In contrast, non-recursive algorithms waits until all measurements are available before beginning any estimate which is time and memory consuming.
\begin{figure}[h]
    \centering
    \includegraphics[width=1\linewidth]{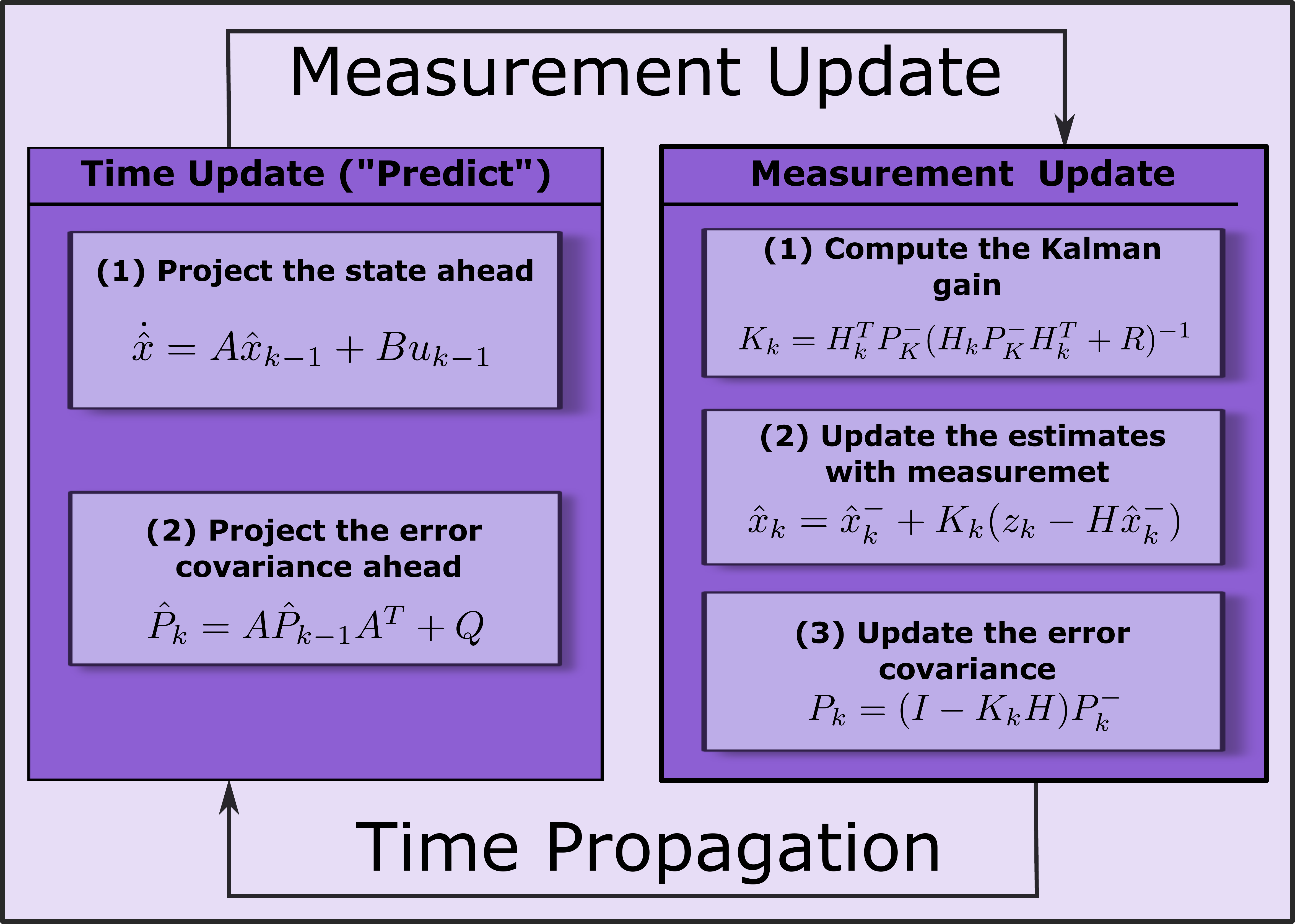}
    \caption{Key update equations of a Kalman filter}
    \label{fig_kal}
\end{figure}

%
\subsection{Components of the Kalman Filter}
The calculation scheme of the Kalman filter  algorithm is shown in Fig.~\ref{fig_kal}.
The Kalman filter has five major components:
\begin{itemize}
	\item The state vector and its covariance matrix
    \item The system model
     \item The measurement vector and its covariance
     \item The measurement model
     \item The algorithm
\end{itemize}
The state vector is a set of parameters that the filter estimates. It is usually composed of the position, velocity and other navigation states or their errors. In our demonstration of the Kalman filter algorithm, we will estimate the errors in the parameters of an INS system, $\delta(\cdot)$, instead of the parameters them self, $(\cdot)$. This implementation is called an error-state implementation. In contrast, when estimating absolute states of the system such as position, velocity, and orientation, the system is known as a total-state implementation. The error-state implementation separates the state into a ``large" nominal state $\hat{x}$, and a ``small" error state, $\delta x$, such that $x=\hat{x}+\delta x$. The error-state implementation can perform better due to the fact that the error dynamic equations we derived are linearized versions of their true equations (due to approximations) and therefore are more accurately evolved in time for small quantities. 

The error covariance matrix $P$, represents the expectation of the square of the deviation of the state vector estimate from the true value of the state vector. The diagonal elements are the variance of the state estimates, while the off-diagonal elements represent the correlation between the errors in the different state estimates. In a Kalman filter, it is required to initialize the state vector and the covariance matrix. Usually, in error-state implementations the state vector is initialized to zero, while the covariance matrix elements are chosen by the designer to reflect the level of confidence of his \emph{a priori} estimates of the initial state vector. 

Each complete iteration of a Kalman filter consists of a propagation and an update step. The state vector and covariance matrix after being propagated in time and before updating, are denoted by, $\hat{\boldsymbol{x}}_k^-$, and $P_k^-$, respectively. Their counterparts following  the measurement update are denoted by $\hat{\boldsymbol{x}}_k^+$, and $P_k^+$.

The vector $\boldsymbol{z}$ consists of a set of measurements related to the state-vector through a deterministic matrix $H$ and with added noise $\boldsymbol{v}$:
\begin{equation}
\label{eq_meas1}
    \boldsymbol{z}= H\boldsymbol{x}+\boldsymbol{v}\,.
\end{equation}

The measurement innovation,  $\delta\boldsymbol{z}^-$, is the difference between the true measurement vector and the one computed from the state vector before a measurement update:
\begin{equation}
    \delta\boldsymbol{z}^-=\boldsymbol{z}-H\hat{\boldsymbol{x}}^-\,.
\end{equation}
The measurement residual, $\delta\boldsymbol{z}^+$, is the difference between the true measurement vector and the one computed from the updated state-vector:

\begin{equation}
    \delta\boldsymbol{z}^+=\boldsymbol{z}-H\hat{\boldsymbol{x}}^+\,.
\end{equation}

The standard Kalman filter assumes that the measurement errors form a zero-mean Gaussian distribution, uncorrelated in time, and with a noise covariance matrix $R$. The covariance matrix $R$ is nothing but the expectation of the square of the measurement noise:
\begin{equation}
    R=E(\boldsymbol{v}\boldsymbol{v}^T)\,.
\end{equation}

%
\subsection{Kalman Filter Algorithm}
\begin{figure}[h]
    \centering
    \includegraphics[width=1\linewidth]{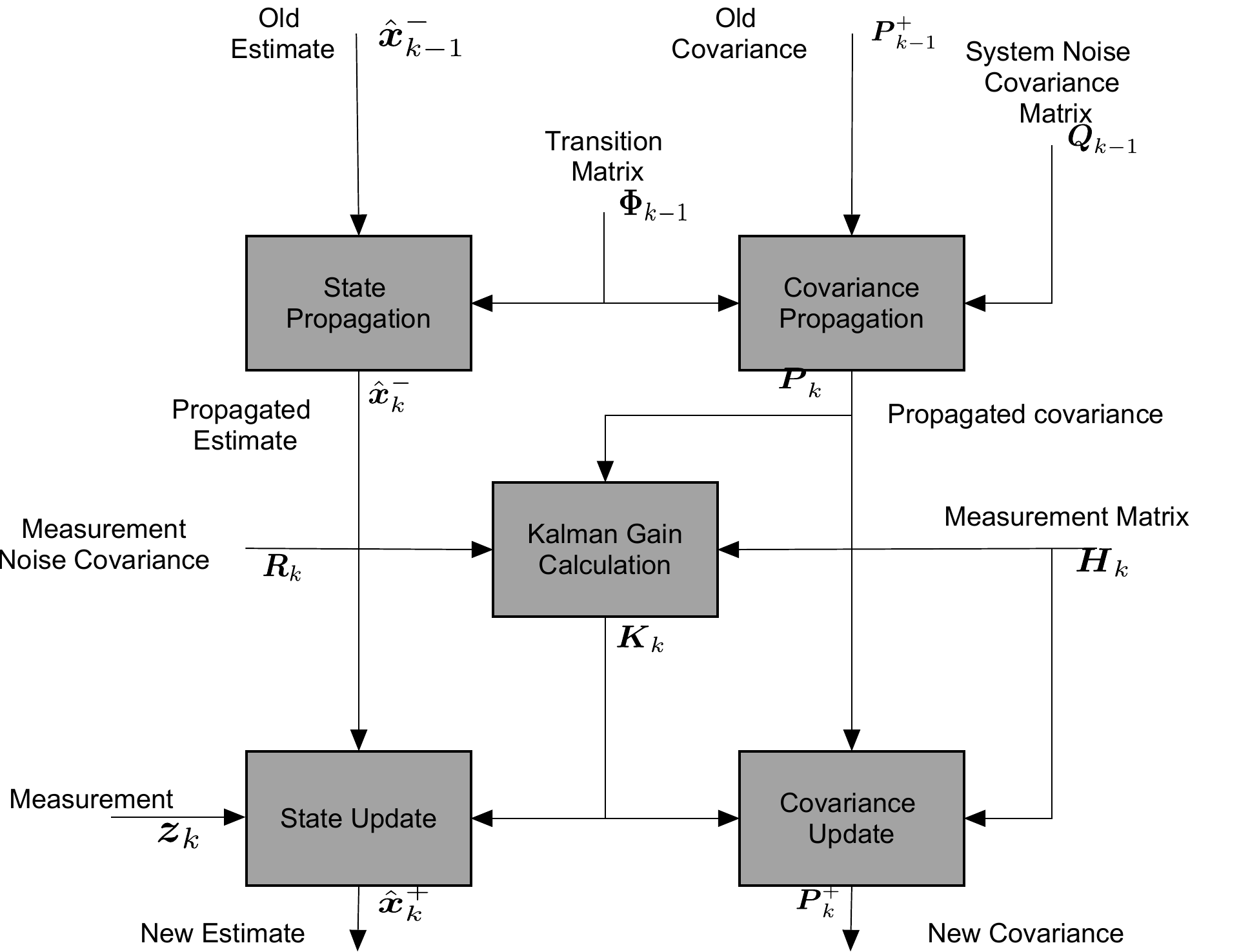}
    \caption{Dataflow graph of a Kalman filter}
    \label{fig_kal1}
\end{figure}

The data flow of the Kalman filter algorithm  is shown in Fig.~\ref{fig_kal1}

The following steps constitute the Kalman filter algorithm~\cite{farrell1999global}:
\begin{enumerate}
    \item Calculate the transition matrix, $\Phi_{k-1}$;
	\item Compute the noise covariance matrix, $Q_{k-1}$;
    \item Propagate the state vector estimate from $\hat{\boldsymbol{x}}_{k-1}^+$ to $\hat{\boldsymbol{x}}_{k}^-$;
     \item Propagate the error covariance matrix from $P_{k-1}^+$ to $P_k^-$;
     \item Compute the measurement matrix, $H_k$;
     \item Calculate the measurement noise covariance matrix, $R_k$;
       \item Calculate the Kalman gain matrix $K_k$;
     \item Extract the measurement, $\boldsymbol{z}_k$;
     \item Update the state vector estimate from $\hat{\boldsymbol{x}}_k^-$ to $\hat{\boldsymbol{x}}_k^+$;
     \item Update the error covariance matrix from $P_k^-$ to $P_k^+$;
\end{enumerate}

The first four steps comprise the system propagation phase of the Kalman filter. The last two steps comprise the update phase of the Kalman filter. Later we will see that it is not necessary to execute the update phase of the Kalman filter with every propagation step of the state-vector.
On the other hand, the system propagation phase should be executed in every iteration  of the Kalman filter. \par

%
\subsubsection{Transition matrix}
The transition matrix describes the dynamics of the system. It defines how the state-vector is propagated with time. It is not a function of any of the state vector parameters. If its elements are a function of time, then it should be updated with every iteration of the Kalman filter.
Since we will derive the equations of an error-state Kalman filter, the elements of the state-vector $\boldsymbol{x}$, will take the form of error terms (see Section~\ref{principles}). 
\begin{equation}
  \label{eq_statevector}
  \begin{split}
    \boldsymbol{x}^T=(\delta\omega_{Gx}^s,\delta\omega_{Gy}^s ,\delta\omega_{Gz}^s,&\delta{a}_{Az},{\psi}_n,{\psi}_e,{\psi}_d,\\
    &\delta v_n,\delta v_e,\delta v_d,\delta\phi,\delta\lambda,\delta h)\,.
\end{split}
\end{equation}

The first three terms are gyroscope biases in the $x, y$, and $z$ directions of the IMU sensor frame. They are considered as random constants, and they are easily modeled as unchanging elements in the state vector. For this reason, we have not derived their error dynamics in the previous section. The fourth term in the state vector is the accelerometer bias in the $z$ direction. It is also modeled as a random constant. We have deleted the
accelerometer biases in the $x$ and $y$ directions since they did not improve the accuracy of our estimated state vector. The collected set of transition elements are collected in one transition matrix~\eqref{eq_transition} in a convenient form to directly observe which parameters are correlated, simply by looking at the first row and the first column entries.
In contrast, it is very important to estimate the bias in the $z$ direction to prevent the vertical channel, ``$h$", in our Kalman filter from diverging. The transition matrix as seen in BOX~\ref{state_transition} is written as $\Phi_{k-1}=I +A*dT$. It is wise to write the transition matrix without the identity matrix  due to the efficiency in calculations achieved in our Kalman filter as will be seen in this section. The elements of the transition matrix will written in Algorithm~\ref{transitionalg}.
\setcounter{MaxMatrixCols}{20}
\begin{figure*}[t]
\begin{equation}\label{eq_transition}
\Phi_{k-1}=
    \begin{bmatrix}
    *\\
    \delta\omega_{Gx}^s\\
    \delta\omega_{Gy}^s\\ \delta\omega_{Gz}^s\\
    \delta{a}_{Az}\\
    {\psi}_n\\
    {\psi}_e\\
    {\psi}_d\\
    \delta v_n\\
    \delta v_e\\
    \delta v_d\\
    \delta\phi\\
    \delta\lambda\\
    \delta h
    \end{bmatrix}
    \begin{bmatrix}
    *   & T_0 &T_1&T_2&T_3&T_4&T_5&T_6&T_7&T_8&T_9&T_A&T_B&T_C\\
    T_0 & 0   & 0 & 0 & 0 & 0 & 0 & 0 & 0 & 0 & 0 & 0 & 0 & 0\\
    T_1 &0    &0  &0  &0  &0  &0  &0  &0  &0  &0  &0  &  0& 0\\ T_2&0&0&0&0&0&0&0&0&0&0&0&0&0\\
    T_3&0&0&0&0&0&0&0&0&0&0&0&0&0\\
    T_4&T_{40}&T_{41}&T_{42}&0&0&0&0&0&0&0&0&0&0\\
    T_5&T_{50}&T_{51}&T_{52}&0&0&0&0&0&0&0&0&0&0\\
    T_6&T_{60}&T_{61}&T_{62}&0&0&0&0&0&0&0&0&0&0\\
    T_7&0&0&0&0&0&T_{75}&T_{76}&0&0&0&0&0&0\\
    T_8&0&0&0&0&T_{84}&0&T_{86}&0&0&0&0&0&0\\
    T_9&0&0&0&T_{93}&T_{94}&T_{95}&0&0&0&0&0&0&0\\
    T_A&0&0&0&0&0&0&0&T_{A7}&0&0&0&0&0\\
    T_B&0&0&0&0&0&0&0&0&T_{B8}&0&0&0&0\\
    T_C&0&0&0&0&0&0&0&0&0&T_{C9}&0&0&0
    \end{bmatrix}\,.
  \end{equation}
\end{figure*}

\begin{algorithm}[!]
\caption{Computation of transition matrix elements}
 \hspace*{\algorithmicindent} \textbf{Input:} State parameters and time step $dT$.\\
 \hspace*{\algorithmicindent} \textbf{Output:} Transition matrix, $\Phi_{k-1}$, entries.
\begin{algorithmic}[1]
\State $kt[4][0]= c_{11}\langle k\rangle *dT$
\State $kt[4][1]= c_{12}\langle k\rangle *dT$
\State $kt[4][2]= c_{13}\langle k\rangle *dT$
\State $kt[5][0]= c_{21}\langle k\rangle *dT$
\State $kt[5][1]= c_{22}\langle k\rangle *dT$
\State $kt[5][2]= c_{23}\langle k\rangle *dT$
\State $kt[6][0]= c_{31}\langle k\rangle *dT$
\State $kt[6][1]= c_{32}\langle k\rangle *dT$
\State $kt[6][2]= c_{33}\langle k\rangle *dT$
\State $kt[7][5]= \Bar{g}  *dT$
\State $kt[7][6]= a_e\langle k \rangle *dT$
\State $kt[8][4]=-\Bar{g}*dT$
\State $kt[8][6]= -a_n\langle k \rangle *dT$
\State $kt[9][3]=c_{33}\langle k\rangle  *dT$
\State $kt[9][4]= -a_e\langle k \rangle *dT$
\State $kt[9][5]= -a_n\langle k \rangle *dT$
\State $kt[A][7]=(1/R) *dT$ \Comment\textit{$A$ is hexadecimal 10}
\State $kt[B][8]= {1}/{(R\cos\phi\langle k \rangle)} *dT$\Comment\textit{$B$ is hexadecimal 11}
\State $kt[C][9]=  -dT$\Comment\textit{$C$ is hexadecimal 12}
\end{algorithmic}
\label{transitionalg}

\end{algorithm}

%
\subsubsection{Error Propagation Matrix}
In order to propagate the covariance matrix we have to apply the following equation:
\begin{equation}
    P_k^-=\Phi_{k-1} P_{k-1}^+\Phi_{k-1}^T+Q_{k-1}\,.
\end{equation}
In order to compute the propagation matrix efficiently, it is beneficial to consider the sparsity of the transition matrix. We shall apply a divide-and-conquer strategy where only the matrix multiplications involving non-zero elements of the transition matrix are executed. We will first compute the matrix, $l=\Phi_{k-1} P_{k-1}^+$, and then multiply the resulting  matrix (we call it intermediate matrix) with $\Phi_{k-1}^T$.\par
To elaborate on the matrix multiplication issue, let us consider that we want to multiply two matrices, $T$ and $P$, where $T$ is sparse and $P$ is not, such as:

\begin{equation}
 T=   \begin{bmatrix}
    0 & t_{12}& 0\\
    0 & 0    & 0 \\
    0 & 0    & 0
\end{bmatrix},\quad  P=   \begin{bmatrix}
    p_{11} & p_{12} & p_{13}\\
    p_{21} & p_{22} & p_{23} \\
    p_{31} & p_{32} &p_{33}
\end{bmatrix}\,.
\end{equation}

The entry $t_{12}$ contributes only to the first row of the resulting matrix, since the first row of the product  uses  the terms, $t_{12}*p_{21}$, $t_{12}*p_{22}$, and $t_{12}*p_{23}$, respectively, in its computations.\par
This strategy decreases the number of  accesses to  memory where the matrices are stored, since the relevant entries are only loaded once for each non-zero transition matrix entry. This greatly reduces the execution time for the Kalman filter on low-cost embedded processors where resources are limited.\par
The detailed steps involved in the computation of $l$, will be shown in Algorithm (\ref{inter1}).\par

To continue the propagation computation of the covariance matrix we shall write the algorithm for  the second part of the matrix multiplication and then finally add the system noise covariance matrix, $Q_{k-1}$. We apply the same strategy as above in Algorithm (\ref{inter2}). The final stage in propagating the error covariance matrix is adding system noise as seen in Algorithm (\ref{inter3}).

%
\subsubsection{Measurement Matrix and Kalman Gain}
The measurement matrix defines how the measurement vector varies with the state vector. This relation in~\eqref{eq_meas1} is repeated here for convenience:
\begin{equation}
\label{eq_meas2}
    \boldsymbol{z}_k= H_k\boldsymbol{x}_k+\boldsymbol{v}_k\,.
\end{equation}

The measurement noise vector in most applications is considered white with a few exceptions. It has a measurement noise covariance matrix, $R_k$, that may be assumed constant.
In our typical implementation we are directly measuring some state vector elements (GNSS positions and velocities).\par
The Kalman gain matrix is used to determine the weighting of the measurement information in updating the state estimates. It is a function of the ratio of the uncertainty of the true measurement, $\boldsymbol{z}_k$ to the uncertainty of the measurements predicted from the state estimates, $H\boldsymbol{x}_k^-$.\par
The Kalman gain matrix is:
\begin{equation}
    K_k=P_k^-H_k^T(H_kP_k^-H_k^T+R_k)^{-1}\,.
\end{equation}

%
\subsubsection{State Vector and Error Covariance Update}
When ever we obtain a measurement the state vector is updated by the measurement vector using this formula:
\begin{equation}
    \hat{\boldsymbol{x}}_k^+=\hat{\boldsymbol{x}}_k^-+ K_k(\boldsymbol{z}_k- H_k\hat{\boldsymbol{x}}_k^-)\,.
\end{equation}
Similarly, the error covariance
matrix is updated with:
\begin{equation}
   P_k^+=(I- K_kH_k) P_k^-\,.
\end{equation}

As the updated state vector estimate is based on more information, the updated state uncertainties are smaller than before the update.\par
We will continue this section by providing the detailed algorithms for the second (update) phase of the Kalman filter. This phase is comprised of calculating the matrix gain, $K_k$, updated state-vector $\hat{\boldsymbol{x}}_k^+$, and updated error covariance matrix, $P_k^+$. These three steps will be implemented for each new measurement obtained (theoretical details deferred to Section~\ref{insights}).

%
\section{Filter Insights}\label{insights}

%
\subsection{Inverse Matrix Calculation}
The most complex and time consuming part of the Kalman filter algorithm is finding the inverse of the matrix in the Kalman filter gain, $K_k$. In order to avoid this tedious calculation, we will prove that it is possible to avoid this inverse matrix calculation by a small trick.\par
We can write the $K_k$ and the $H_k$ matrices as follows:
\begin{equation}
    K_k=\begin{bmatrix}
    \vdots & \vdots & \vdots\\
    K_1    &  K_2 & \cdots\\
     \vdots & \vdots& \vdots
    \end{bmatrix},
\end{equation}
and
\begin{equation}
    H_k=\begin{bmatrix}
    \cdots & H_1 & \cdots\\
    \cdots &  H_2 & \cdots\\
     \cdots & \vdots & \cdots
    \end{bmatrix}\,.
\end{equation}

Using this notation, we have, $K_kH_k= K_1H_1+K_2H_2+\cdots$. Thus, it is easily shown that the error covariance matrix can be written as follows:
\begin{equation}
   P_k^+=\underbrace{\underbrace{P_k^-+K_1H_1 P_k^-}_\text{$P_k^+$ after 
   first measurement }+K_2H_2 P_k^-}_\text{$P_k^+$ after 
   second measurement}+\cdots
\end{equation}

Note that the sum of the first two terms is nothing but the updated error covariance matrix associated with one measurement. Adding the third term to it, we obtain the error covariance matrix after the second measurement is manipulated. As shown, it is  possible to update the error covariance matrix after each reported measurement. This is legitimate as long as we do not propagate the error covariance matrix while manipulating the measurements. It is only required to compute the Kalman gain and correct the state vector after each update of the covariance matrix. The benefit of following this strategy is that the matrix inversion in the Kalman gain computation is transformed to a simple scalar inversion. So by taking any single measurement element only, we can update the covariance matrix and derive and apply system corrections for that single measurement element, much simpler than we can with multiple measurements. In this case, we avoid matrix inversion. It is important that we update everything with a single measurement before using the next measurement, and that we use the updated status before applying the next measurement. All updates must be performed before the next navigation integration cycle. We should always propagate the Kalman filter after every integration of the navigation equations, and  only  update  the  Kalman  filter  (and  also  make  system  corrections)  whenever  we  have  measurements.  The frequency of these updates could be the same or less than the Kalman filter propagation frequency. It depends on the source of our measurements.

%
\subsection{Error Dynamics Approximations}
It is also beneficial to address the approximations we made in deriving the error dynamic equations,
\eqref{eq_fffgggkkkjj} and \eqref{eq_erdynapp}, for  those readers who may be concerned. With low-grade IMU we simply forget the Earth's rotation and consider a local flat Earth. It is impossible to detect the very slow rotation of the earth with all the noise and random drift of the gyroscopes. There is no harm in including these factors but they are unlikely to make any improvements.\par

%
\subsection{Error-State Kalman Filter}
It is necessary to remember that we are performing an error-state Kalman filter where the estimated states are INS errors. After each Kalman filter iteration, the estimated states are applied to the corresponding navigation parameters, and thus the state vectors are reset to zero. Consequently, the state vector itself does not need to be propagated forward in time. This type of implementation is called a closed-loop implementation. It is still critical however that the state covariance be propagated using the following equation:
\begin{equation}
    P_k^-=\Phi_{k-1} P_{k-1}^+\Phi_{k-1}^T+Q_{k-1}\,.
\end{equation}

In this closed-loop technique, since the estimated  errors are fed back every iteration, the Kalman filter states are zeroed which keeps the errors of the filter small. This has the effect of minimizing the errors introduced in linearizing the system or process model, since higher order terms in the Taylor series expansion gets smaller and smaller. This is in contrast to  the open-loop implementation where there is no feedback, and thus, the states will get larger as time progresses.\par
We shall also explain a concept which is not very well common between Kalman filter practitioners, related to the idea of how velocity measurements ($v_n, v_e, v_d$) are essential in estimating  tilt errors $\psi_n$, $\psi_e$, and $\psi_d$.\par

Suppose that the INS is stationary (zero-velocity), and that we have a positive $\psi_n$ error, which means that the system model does not have an exact representation of the rotation of the IMU relative to the local north pointing axis. Instead, it is rotated clockwise by the angle $\psi_n$ about the north pointing axis.
 Then we will not have the correct transformation value for $dv_e$ after performing the direction cosine transformation of $dv_x$, $dv_y$ and $dv_z$ (Algorithm (\ref{velinteg})). The effect is that the east velocity will increment by an error equal to $(\psi_n * g * dT)$. This is the same  term relating the error $\delta v_e$ to $\psi_n$ in the transition matrix~\eqref{eq_transition}. Thus, the system computed velocity $v_e$ is no longer zero. But we know it is zero, so we give a measurement to the Kalman filter equal to $-v_e$. This generates a whole set of corrections for the system, including highly weighted corrections to $v_e$, $\psi_n$ and gyroscope biases. The other terms are only weakly coupled to $\delta v_e$ in the Kalman filter and have only very small, almost negligible corrections (for these corrections, you need the $v_n$ and $v_d$ measurements). We then apply quaternion correction, which reduces the error in $\psi_n$ (as well as the other terms) so that, next time round,  the error in $v_e$ is smaller. This way, after several cycles, we reduce the error in $\psi_n$  until the roll and pitch values and $v_e$ are correct (gyroscope biases take a bit longer).\par
So as we can see, it is the accelerometers (delta velocities) which determine the alignment and not the gyroscopes. The gyroscopes are only there so that sensed rotation can be immediately applied to the attitude solution instead of waiting until the Kalman filter, aided by velocity measurements, eventually finds the new attitude angles (direction cosine matrix).

These are some recommendations for Kalman filter designers:
\begin{itemize}[label={\tiny\raisebox{1ex}{\textbullet}}]
\item  Check that you can read and display the IMU data with your software.
\item Prove that quaternion integration, body to navigation direction-cosine-matrix computation and roll-pitch-heading routines  work without the Kalman filter, by starting level and manually rotating for a short time. To do this you do not need to transform and integrate delta velocities. 
\item Transform, integrate and display velocities; they should rapidly become very large.
\item Develop your Kalman filter by displaying all the variances adjacent to their respective state matrix elements.  Note that the typical error of the of a state element $``i"$ (squared) should approximately equal to its corresponding i'th diagonal entry in the  covariance matrix $P_k$.

\end{itemize}

%
\subsection{System Model Error Sources}
In the last few years, Microelectromechanical system (MEMS) gyroscopes and accelerometers are beginning to take market away from traditional inertial sensors like fiber optic gyroscopes (FOG) and ring laser gyroscopes (RLG). This take over has been occurring due to improved error characteristics, environmental stability, better bandwidth, enhanced g-sensitivity, and the plethora of  of embedded computational power that can run advanced fusion and sensor error modeling algorithms. This transition couldn't have been achieved if not for the advancements in MEMS technology which had  remarkable improvements on the error characteristics of the sensors~\cite{niu2006development}\cite{goodall2013battle}\cite{schmidt2015navigation}.\par
In system modeling in general,  we  only  consider  the  main  error  sources.  We ignore  the  terms  which  have  little  effect  on  system  performance,  but  allow  for  these  small  effects  as  additional noise  in   other  terms.  This  is  not perfectly correct,  but  its  is a good compromise, since it  simplifies the mathematics significantly, minimises the size of the Kalman filter and works surprisingly well.
\begin{figure}[h]
    \centering
    \includegraphics[width=0.8\linewidth]{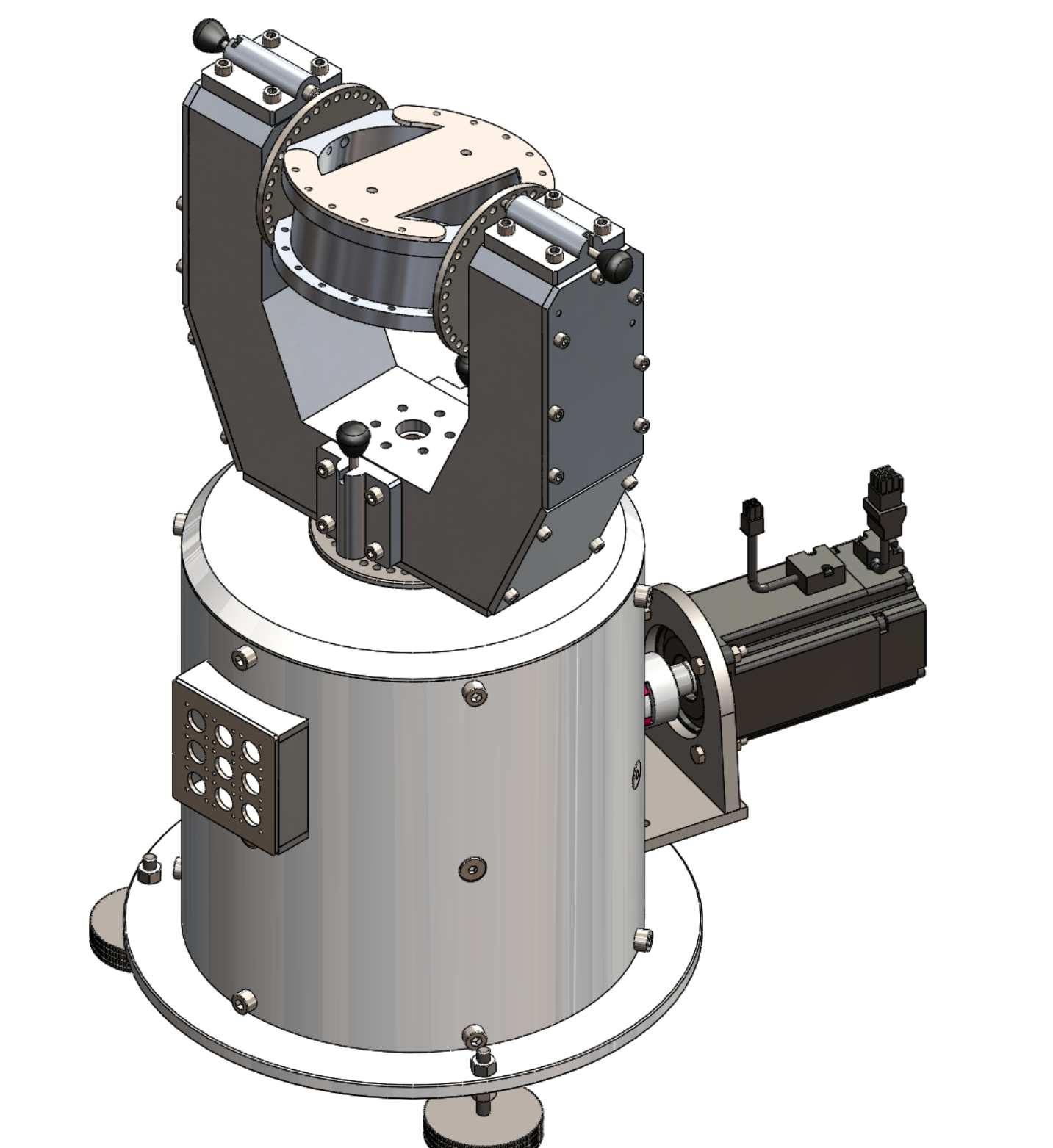}
    \caption{Rate-tables are used for gyroscopes/IMU calibration or hardware-in-loop (HWIL) testing. Payloads are mounted on the table top platen. A pattern of threaded holes accept a variety of test loads. It is often equipped with high speed optical or electrical slip-rings to transmit data to and from the payload under test. It contains a direct drive brushless motor with dedicated amplifiers,  controllers and a heavy duty power supply. Some of the main error sources that may be acquired from the calibration process include: sensor-to-axis misalignments, gyroscope and accelerometer scale factor errors, and bias errors. The rate-table is usually mounted inside a temperature chamber in order for the payload unit to be calibrated at equally spaced temperatures ranging from $\unit[-40]{\degree C}$ to $\unit[+85]{\degree C}$~\cite{lawrence2012modern}.}
    \label{fig_ratetable}
\end{figure}

For example $d\phi_x, d\phi_y $, and $d\phi_z$ are actual measurements and their values would be absolutely correct if there were no errors in scale factors, biases and mis-alignments, and therefore we do not consider errors in these terms (corrected for by rate-table Fig.~\ref{fig_ratetable}).
However, if there were additional errors in $d\phi_x, d\phi_y $, and $d\phi_z$, then these errors would come from other sources such as measurement electronics, g-sensitivity and non-linearity, which we have ignored. We have not modelled these errors in the mathematics but instead we do account for these additional errors by having larger values of system noise in the other terms.\par
A good example where we add "unnecessary" noise in the Kalman filter is in $\delta v_n, \delta v_e$, and $\delta v_d$. Where could noise appear in these terms? A very minute amount could come from computing noise (e.g., loss of numerical values which are smaller than the least-significant-bit when integrating), but that's all. So why do we add velocity noise in the Kalman filter? The answer is, it compromises for errors caused by non-considered effects and thus prevents the mathematics in the Kalman filter from exploding due to a reduced mathematical model.

%
\subsection{IMU and GNSS Time Synchronization}
We have assumed that GNSS and IMU sytems are time synchronized until now. In practice, they are not.
Suppose there is a small time lag between both systems. If the vehicle is moving at constant velocity then the solutions provided by both systems will perfectly match, and thus no error is introduced. In contrast, in case of a small acceleration experienced by the vehicle, the timing lag will manifest itself as a position and velocity difference between the two system solutions. Therefore, it is mandatory to have dedicated hardware in your system that compensates for this time lag, or take care of it by software.

\tcbset{width=\textwidth}

\begin{tcolorbox}[float*=!tbph]
\begin{algorithm}[H]
\caption{Efficient propagation of the error covariance matrix (part a)}
\begin{multicols}{2}
 \hspace*{\algorithmicindent} \textbf{Input:} Transition matrix, $\Phi_{k-1}$, and covariance matrix, $P_{k-1}^+$.\\
 \hspace*{\algorithmicindent} \textbf{Output:} Intermediate propagated covariance matrix, $l=\Phi_{k-1} P_{k-1}^+$. 
\begin{algorithmic}[1]
\\\textbf{\color{Red}Compute Intermediate covariance matrix $l=\Phi_{k-1} P_{k-1}^+$}\\
 \Function{KO}{$i,j$}

 \For{\texttt{(k= 0;k <13;k++)}}
        \State \texttt{$ko[i][k] = ko[i][k] + kt[i][j] * kp[j][k]$;}
      \EndFor
\EndFunction
\\
\Function{KL}{$void$}

 \For{\texttt{(i= 0;i <13;i++)}}
 \For{\texttt{(j= 0;j <13;j++)}}
        \State \texttt{$ko[i][j] = kp[i][j]$;}
      \EndFor
      \EndFor

\State \texttt{$i=4;j=0$;}
\State {KO($i,j$);}
\State \texttt{$i=4;j=0$;}
\State {KO($i,j$);}
\State \texttt{$i=4;j=2$;}
\State {KO($i,j$);}
\State \texttt{$i=5;j=0$;}
\State {KO($i,j$);}
\State \texttt{$i=5;j=1$;}
\State {KO($i,j$);}
\State \texttt{$i=5;j=2$;}
\State {KO($i,j$);}
\State \texttt{$i=6;j=0$;}
\State {KO($i,j$);}
\State \texttt{$i=6;j=1$;}
\State {KO($i,j$);}
\State \texttt{$i=6;j=2$;}
\State {KO($i,j$);}
\State \texttt{$i=7;j=5$;}
\State {KO($i,j$);}
\State \texttt{$i=7;j=6$;}
\State {KO($i,j$);}
\State \texttt{$i=8;j=4$;}
\State {KO($i,j$);}
\State \texttt{$i=8;j=6$;}
\State {KO($i,j$);}
\State \texttt{$i=9;j=3$;}
\State {KO($i,j$);}
\State \texttt{$i=9;j=4$;}
\State {KO($i,j$);}
\State \texttt{$i=9;j=5$;}
\State {KO($i,j$);}
\State \texttt{$i=10;j=7$;}
\State {KO($i,j$);}
\State \texttt{$i=11;j=8$;}
\State {KO($i,j$);}
\State \texttt{$i=12;j=9$;}
\State {KO($i,j$);}
\EndFunction
\end{algorithmic}
\end{multicols}
\label{inter1}
\end{algorithm}

\end{tcolorbox}

\tcbset{width=\textwidth}

\begin{tcolorbox}[float*=!tbph]
\begin{algorithm}[H]
\caption{Efficient propagation of the error covariance matrix (part b)}
\begin{multicols}{2}
 \hspace*{\algorithmicindent} \textbf{Input:} Intermediate propagated covariance matrix, $l=\Phi_{k-1} P_{k-1}^+$.  \\
 \hspace*{\algorithmicindent} \textbf{Output:} Propagated Covariance matrix, $P_k^-=\Phi_{k-1} P_{k-1}^+\Phi_{k-1}^T$.
\begin{algorithmic}[1]
\\\textbf{\color{Red}Compute Intermediate covariance matrix $l*\Phi_{k-1}^T$}\\
 \Function{KP}{$i,j$}

 \For{\texttt{(k= 0;k <13;k++)}}
        \State \texttt{$kp[k][i] = kp[k][i] + kt[i][j] * ko[k][j]$;}
      \EndFor
\EndFunction
\\
\Function{KL}{$void$}

 \For{\texttt{(i= 0;i <13;i++)}}
 \For{\texttt{(j= 0;j <13;j++)}}
        \State \texttt{$kp[i][j] = ko[i][j]$;}
      \EndFor
      \EndFor

\State \texttt{$i=4;j=0$;}
\State {KP($i,j$);}
\State \texttt{$i=4;j=0$;}
\State {KP($i,j$);}
\State \texttt{$i=4;j=2$;}
\State {KP($i,j$);}
\State \texttt{$i=5;j=0$;}
\State {KP($i,j$);}
\State \texttt{$i=5;j=1$;}
\State {KP($i,j$);}
\State \texttt{$i=5;j=2$;}
\State {KP($i,j$);}
\State \texttt{$i=6;j=0$;}
\State {KP($i,j$);}
\State \texttt{$i=6;j=1$;}
\State {KP($i,j$);}
\State \texttt{$i=6;j=2$;}
\State {KP($i,j$);}
\State \texttt{$i=7;j=5$;}
\State {KP($i,j$);}
\State \texttt{$i=7;j=6$;}
\State {KP($i,j$);}
\State \texttt{$i=8;j=4$;}
\State {KP($i,j$);}
\State \texttt{$i=8;j=6$;}
\State {KP($i,j$);}
\State \texttt{$i=9;j=3$;}
\State {KP($i,j$);}
\State \texttt{$i=9;j=4$;}
\State {KP($i,j$);}
\State \texttt{$i=9;j=5$;}
\State {KP($i,j$);}
\State \texttt{$i=10;j=7$;}
\State {KP($i,j$);}
\State \texttt{$i=11;j=8$;}
\State {KP($i,j$);}
\State \texttt{$i=12;j=9$;}
\State {KP($i,j$);}
\EndFunction
\end{algorithmic}
\end{multicols}
\label{inter2}
\end{algorithm}

\end{tcolorbox}

\begin{algorithm}[!h]
\caption{Addition of system noise to error covariance matrix}
 \hspace*{\algorithmicindent} \textbf{Input:} Propagated Covariance matrix $P_k^-$ minus noise.\\
 \hspace*{\algorithmicindent} \textbf{Output:} Propagated Covariance matrix $P_k^-$.
\begin{algorithmic}[1]
\\\textbf{\color{Red}Add system noise covariance matrix to $\Phi_{k-1} P_{k-1}^+\Phi_{k-1}^T$}\\
\Function{KP}{$void$}
 \For{\texttt{(i= 0;i <13;i++)}}
    \State \texttt{$kp[i][i] = kp[i][i] + kq[i] * dt$;}
      \EndFor
\EndFunction
\end{algorithmic}
\label{inter3}
\end{algorithm}

\tcbset{width=\textwidth}

\begin{tcolorbox}[float*=!tbph]
\begin{algorithm}[H]
\caption{State-vector and error covariance matrix update}
\begin{multicols}{2}
 \hspace*{\algorithmicindent} \textbf{Input:} Propagated state-vector $\boldsymbol{x}_k^-$ and error covariance matrix ${P}_k^-$.\\
 \hspace*{\algorithmicindent} \textbf{Output:} Updated state-vector $\boldsymbol{x}_k^+$ and error covariance matrix ${P}_k^+$.\\
\begin{algorithmic}[1]
\\\textbf{\color{Red}Update the Kalman filter for each individual measurement $z_k$}
\State \texttt{$i=7$;}
\State \texttt{$kz[i] = GpsVeln -v_n;$}
\State {UPDATE($i$);}
\State {CORRECTION($void$);}
\State {ZEROING($void$);}

\State \texttt{$i=8$;}
\State \texttt{$kz[i] = GpsVele -v_e;$}
\State {UPDATE($i$);}
\State {CORRECTION($void$);}
\State {ZEROING($void$);}

\State \texttt{$i=9$;}
\State \texttt{$kz[i] = GpsVeld -v_d;$}
\State {UPDATE($i$);}
\State {CORRECTION($void$);}
\State {ZEROING($void$);}

\State \texttt{$i=10$;}
\State \texttt{$kz[i] = GpsLat -\phi;$}
\State {UPDATE($i$);}
\State {CORRECTION($void$);}
\State {ZEROING($void$);}

\State \texttt{$i=11$;}
\State \texttt{$kz[i] = GpsLon -\lambda;$}
\State {UPDATE($i$);}
\State {CORRECTION($void$);}
\State {ZEROING($void$);}

\State \texttt{$i=12$;}
\State \texttt{$kz[i] = GpsHeight -h;$}
\State {UPDATE($i$);}
\State {CORRECTION($void$);}
\State {ZEROING($void$);}

 \Function{UPDATE}{$i$}
 \For{\texttt{(j= 0;j <13;j++)}}
        \State \texttt{$kw[i][j] = kp[i][j] * tmp ;$}
      \EndFor
      
 \For{\texttt{(j= 0;j <13;j++)}}
 \For{\texttt{(k= 0;k <13;k++)}}
        \State \texttt{$ko[j][k] = kp[j][i] * kw[i][k]$;}
      \EndFor
      \EndFor 
      
 \For{\texttt{(j= 0;j <13;j++)}}
 \For{\texttt{(k= 0;k <13;k++)}}
        \State \texttt{$kp[j][k] = kp[j][k] - ko[j][k]$;}
      \EndFor
      \EndFor 
        
   \For{\texttt{(j= 0;j <13;j++)}}
        \State \texttt{$kx[j] = kx[j] + ki[i] * kw[i][j]$}
      \EndFor
      
\EndFunction

\Function{CORRECTION}{$void$}
\State \texttt{$\omega_{GBx}=\omega_{GBx}+kx[0]$;}\Comment\textit{Gyro Bias Correction}
\State \texttt{$\omega_{GBy}=\omega_{GBy}+kx[1]$;}\Comment\textit{Gyro Bias Correction}
\State \texttt{$\omega_{GBz}=\omega_{GBz}+kx[2]$;}\Comment\textit{Gyro Bias Correction}
\State \texttt{$a_{Bz}=a_{Bz}+kx[3]$;}\Comment\textit{Accel Bias Correction}
\State \texttt{$\psi_n=-kx[4]$;}
\State \texttt{$\psi_e=-kx[5]$;}
\State \texttt{$\psi_d=-kx[6]$;}
\State \texttt{$d\phi_x = \psi_n * c_{11} + \psi_e * c_{21} +\psi_d* c_{31};$}
\State \texttt{$d\phi_y = \psi_n * c_{12} + \psi_e * c_{22} +\psi_d* c_{32};$}
\State \texttt{$d\phi_z = \psi_n * c_{13} + \psi_e * c_{23} +\psi_d* c_{33};$}
\State \textit{Do Quaternion Integration (Algorithm~\ref{qint});}
\State \texttt{$v_n = v_n + kx[7];$}  
\State \texttt{$v_e = v_e + kx[8];$}
\State \texttt{$v_d = v_d + kx[9];$}
\State \texttt{$\phi = \phi + kx[10];$}  
\State \texttt{$\lambda = \lambda + kx[11];$}
\State \texttt{$h = h + kx[12];$}

\EndFunction

\Function{ZEROING}{$void$}

 \For{\texttt{(i= 0;i <13;i++)}}
        \State \texttt{$kx[i] = 0$;}
      \EndFor

\EndFunction

\end{algorithmic}
\end{multicols}
\label{inter4}
\end{algorithm}

\end{tcolorbox}

\section{Other Attitude Filters}\label{other}
\begin{figure*}[!h]
\centering
\includegraphics[width=0.8\textwidth]{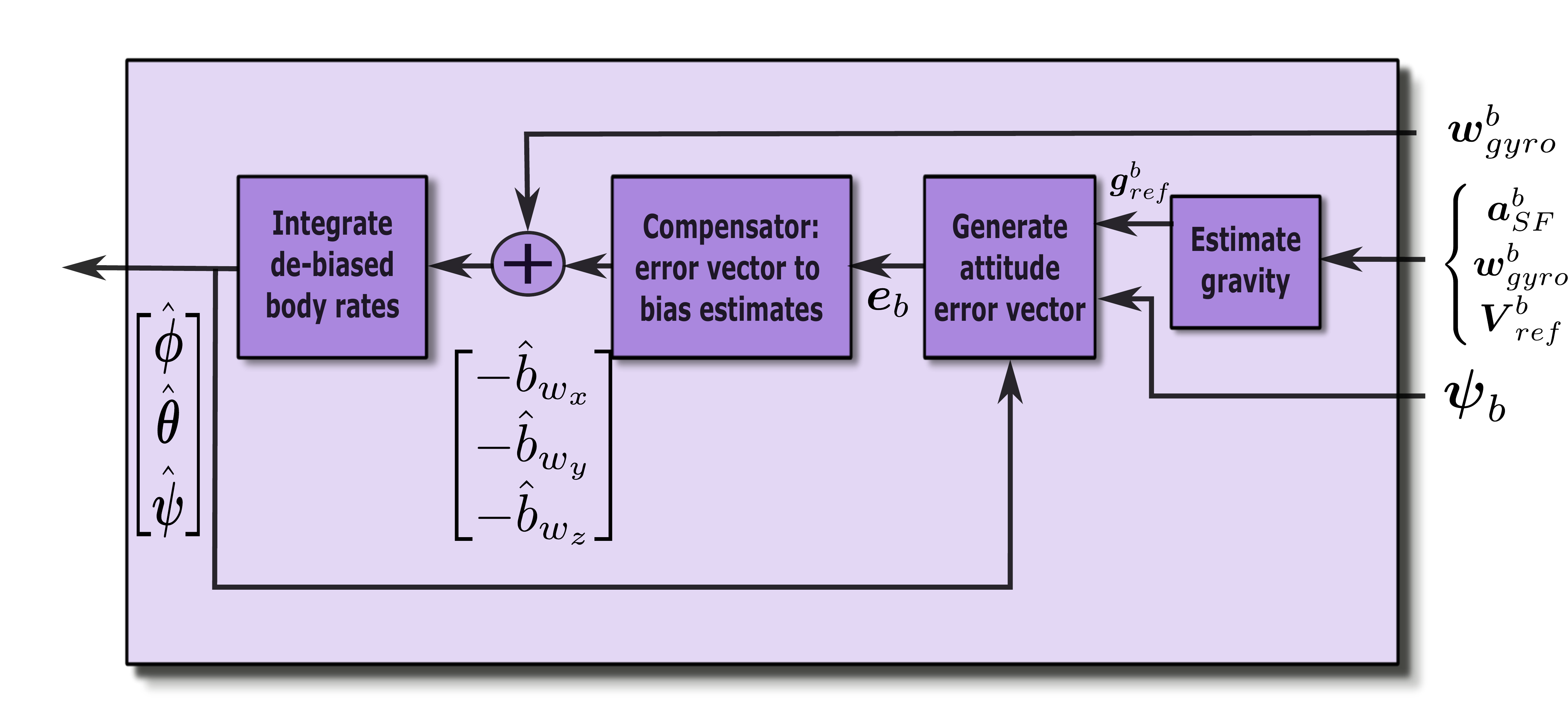}
\caption{An attitude heading reference system feedback controller estimates the body orientation by fusing high-bandwidth gyro angular rate measurements with low-bandwidth attitude references. The yaw reference comes from a magnetometer or a GPS course-based estimate. Pitch and roll references are acquired from an estimate of the gravity vector via centripetally corrected accelerometer measurements. Essentially, this feedback controller uses a compensator to estimate the gyro biases
by regulating the error between the estimated orientation and the orientation expressed by the
low-bandwidth attitude references.}
\label{fig_nav_aw}
\end{figure*}

A computationally simpler alternative to a Kalman filter
for the GPS-INS algorithms
in Fig.~\ref{fig_kal} is to use a feedback
controller to correct for gyro
drift~\cite{euston2008complementary}\cite{beard2005autonomous}\cite{bryson2004vehicle}.
In the GPS-INS
feedback controller shown
in Fig.~\ref{fig_nav_aw}, a compensator typically based on proportional–integral control, is
used to estimate gyro biases
based on a body-frame attitude error vector, $\boldsymbol{e}^b$. $\boldsymbol{e}^b$
is derived by noting that if
the output attitude estimate
were correct and other error
sources were negligible, then
the estimated North and Down directions would align with those observed from a yaw reference ($\boldsymbol{\psi}_{ref}$) and a
body-frame gravity vector reference ( $\boldsymbol{g}_{ref}^b$). The gravity vector reference is derived
from the centripetally corrected accelerometer measurements:
\begin{equation}\label{eq_spec_force}
    \boldsymbol{g}_{ref}^b=\boldsymbol{w}_{gyro}^b\times \boldsymbol{V}_{ref}^b-\boldsymbol{f}_{accel}^b\,.
\end{equation}

The error vector $\boldsymbol{e}^b$ is expressed as the summation of errors generated from the yaw reference ($\boldsymbol{\psi}_{ref}$) and body-frame gravity vector reference ( $\boldsymbol{g}_{ref}^b$).
So we have:
\begin{equation}
    \boldsymbol{e}^b= \boldsymbol{e}_\psi^b+\boldsymbol{e}_g^b\,,
\end{equation}
where
\begin{equation}\label{eq_hjl}
\begin{split}
   \boldsymbol{e}_\psi^b&=\left(C_x(\hat{\phi})C_y(\hat{\theta})C_z({\psi}_{ref}) \begin{bmatrix}
   1\\
   0\\
   0
   \end{bmatrix}
   \right)\times\left( \hat{C}_{ned}^b \begin{bmatrix}
   1\\
   0\\
   0
   \end{bmatrix}\right)\,,\\[1em]
   \boldsymbol{e}_g^b&= \left(\frac{\boldsymbol{g}_{ref}^b}{\|\boldsymbol{g}_{ref}^b\|} \right)\times\left( \hat{C}_{ned}^b \begin{bmatrix}
   1\\
   0\\
   0
   \end{bmatrix}\right),
   \end{split}
\end{equation}
and $\hat{C}_{ned}^b=\left(C_x(\hat{\phi})C_y(\hat{\theta})C_z({\hat{\psi}})\right)$ is the current estimate of the transformation matrix from the $ned$ frame to the body frame.\par

In~\eqref{eq_hjl}, $\boldsymbol{e}_\psi^b$  is the rotation vector expressed in body coordinates between the observed North direction defined by $\boldsymbol{\psi}_{ref}$ and the system estimated North-direction.(Note that the cross-product between two vectors, $\times$, between two vectors yields a vector orthogonal to both with a magnitude proportional to the sine of the angle between them).
Similarly, $\boldsymbol{e}_g^b$ is the rotation vector in body coordinates between the centripetally corrected gravity direction estimated from the accelerometers and the system-estimated Down direction. As a result, the combined error vector $\boldsymbol{e}^b$, expresses the angular error and the rotation axis between the reference NED coordinate frame and the system-estimated NED frame. This feedback error vector id filtered via a compensator (proportional-integral controller) to generate the estimated gyroscope biases. Subtracting these estimated gyroscope biases from the gyroscope measurements and performing a quaternion integration on the de-biased gyroscope data, we obtain the desired Euler angles.\par
The drawback of this filer is it its assumption of zero average linear acceleration for~\eqref{eq_spec_force} to apply. While this assumption is true for most vehicles spending most of the time cruising with zero acceleration, it may lead to unaccepted results in terms of attitude estimation in case of prolonged or transient linear accelerations. Nevertheless whenever this transient acceleration is gone, it can recover and converge to the true values of attitude in fast manner.\par
\begin{figure*}[!t]
    \centering
    \includegraphics[width=\textwidth]{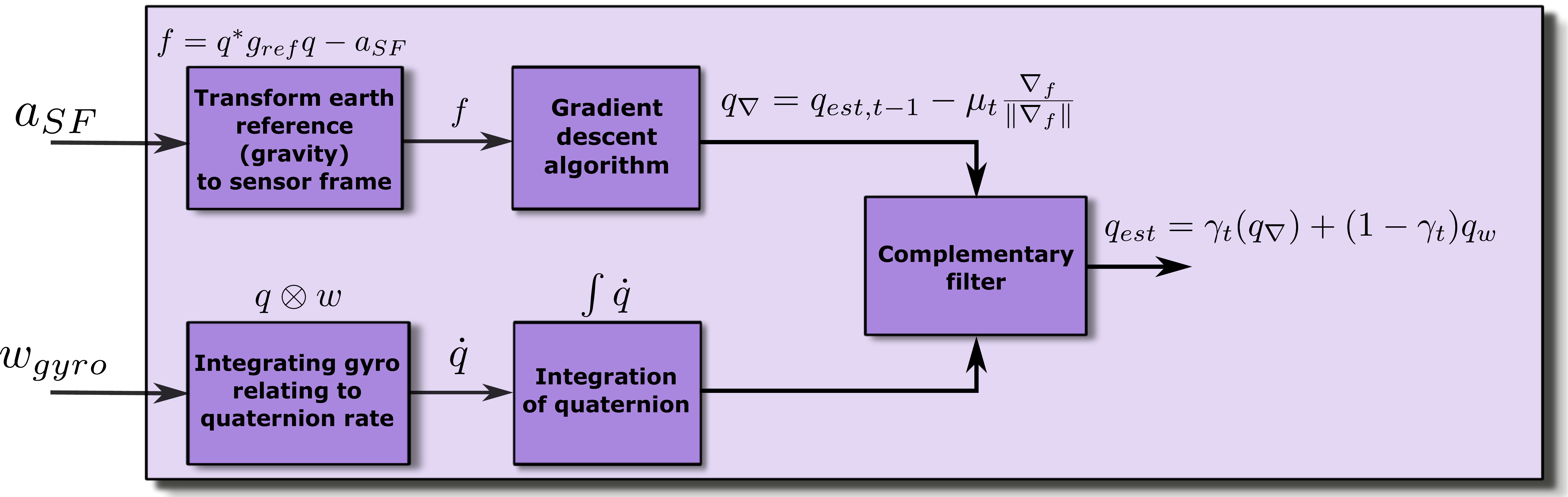}
    \caption{Madgwick Filter based on gradient descent optimization}
    \label{fig_nav_alg111}
\end{figure*}

Yet, another computationally efficient non-linear attitude filter is shown in Fig.~\ref{fig_nav_alg111}. The algorithm uses a quaternion
representation, allowing accelerometer and magnetometer data to be used in an analytically derived and optimised gradient
descent algorithm to compute the direction of the gyroscope measurement error as a quaternion derivative. The quaternion
derivative describing rate of change of the NED frame relative to the sensor frame can be calculated as~\cite{cooke1992npsnet} shown inequation~\eqref{eq_madg}:
\begin{equation}\label{eq_madg}
    \dot{\boldsymbol{q}}= \frac{1}{2}\hat{\boldsymbol{q}}\otimes\boldsymbol{w}\,,
\end{equation}
where $\boldsymbol{w}=
(0,w_x,w_y,w_z)$ is augmented  angular rate measurement vector delivered by the gyroscopes and $\hat{\boldsymbol{q}}$ is the normalized quaternion vector representing the relative orientation between the NED frame and the body coordinate frame.\par 
Provided the initial conditions are known, equation~\eqref{eq_madg} can be numerically integrated to solve for $\boldsymbol{q}$. If the time step considered is $dT$ the integration can be done using the following
\begin{equation}\label{eq_madg2}
  \hat{\boldsymbol{q}}_t=   \hat{\boldsymbol{q}}_{t-1}+ \dot{\boldsymbol{q}}_tdT\,.
\end{equation}

In the context of an orientation estimation algorithm, it will
initially be assumed that an accelerometer will measure only
gravity and a magnetometer will measure only the earth’s
magnetic field. If the direction of an earth’s reference field is known
in the earth frame, a measurement of the field’s direction
within the sensor frame will allow an orientation of the
sensor frame relative to the earth frame to be calculated.
However, for any given measurement there will not be a unique
sensor orientation solution, instead there will be infinite solutions
represented by all those orientations achieved by the rotation of the true sensor frame around an axis parallel with the field's direction.
A quaternion representation requires a single solution to be
found. This may be achieved through the formulation of an
optimisation problem where an orientation of the sensor, $\hat{\boldsymbol{q}}$,
is found as that which aligns a predefined reference direction in the earth frame, $\boldsymbol{d}_{ref}$, with the measured field in the body coordinate  frame, $\boldsymbol{s}$; thus solving~\eqref{eq_madg3} where $\boldsymbol{f}$ in~\eqref{eq_madg4},
defines the objective function~\cite{madgwick2011estimation}:
\begin{equation}\label{eq_madg3}
     \min_{\hat{\boldsymbol{q}}} \boldsymbol{f}(\hat{\boldsymbol{q}}, \boldsymbol{d}_{ref}, \boldsymbol{s})\,,
\end{equation}
where
\begin{equation}\label{eq_madg4}
  \boldsymbol{f}(\hat{\boldsymbol{q}}, \boldsymbol{d}_{ref}, \boldsymbol{s} )=\hat{\boldsymbol{q}}^*\otimes \boldsymbol{d}_{ref} \otimes \hat{\boldsymbol{q}}- \boldsymbol{s}\,.
\end{equation}

Many optimisation algorithms exist but the gradient descent
algorithm is one of the simplest to both implement and
compute. The equation used to implement the gradient descent algorithm is shown in the block diagram (Fig.~\ref{fig_nav_alg111}).\par

%
\section{Summary}\label{summary}

The use of attitude estimation filters have seen an explosive growth in the past few decades, specifically with the growth of the smartphones market. The use of touchscreens for gaming has popularized motion detection and orientation estimation within the portable computing platform.
Orientation estimation algorithms for inertial sensors is a is a mature field of research. Modern techniques
~\cite{bachmann2001inertial}, ~\cite{mahony2008nonlinear}, and~\cite{martin2010design} have focused on simpler algorithms that ameliorate the computational load and parameter tuning burdens
associated with conventional Kalman-based approaches. The algorithms presented in this paper employs some cost-effective techniques and is able to offer some key advantages in terms of energy efficiency with out sacrificing accuracy, aiming at deploying this technology in low-cost hardware.\par
In this article, we have showed the advantages of fusing GNSS and INS systems, in terms of their complementary properties. Various integration architectures are also exploited, that demonstrate how the lower bandwidth of the GNSS system can act as an online calibrator for the INS system. Specifically, how the gyroscope and accelerometer errors are compensated for, using  estimated drift and biases.\par
This article points the reader's attention to various insights necessary for a successful GNSS-INS design. It is a hands-on approach that when followed step-by-step, by applying the presented navigation  and fusing filter algorithms, guarantees a completely working and efficient stand-alone system. The authors have made their best to keep this article self-contained. To the best of our knowledge, this is the one of the few articles that fills the gaps between the theoretical and applied aspects of inertial navigation systems.\par
\appendix

%
Algorithm~(\ref{alg_arctan}) is an accurate polynomial implementation of the arc-tangent function. It takes two arguments as inputs. It is similar to the four-quadrant inverse tangent function, ``atan2(x,y)" in Matlab.

\begin{algorithm}[h]
\caption{Efficient computation of $\arctan(C1, C2)$}
 \hspace*{\algorithmicindent} \textbf{Input:} Arguments $C1$ and $C2$. \\
 \hspace*{\algorithmicindent} \textbf{Output:} Arc tangent of the input arguments. 
\begin{algorithmic}[1]
\If {$\abs{C1} \geq \abs{C2}$}
  \State $R1 =\frac{C2}{C1};$
\Else
\State $R1 =\frac{C1}{C2};$
\EndIf
\If {$\abs{C1} == -\abs{C2}$}
  \State $R1 =-1;$
\EndIf
\If {$\abs{C1} == \abs{C2}$}
  \State $R1 =1;$
\EndIf
\State $R2 = R1 * R1;$
\State $R3 = R1 * R2;$
\State $R5 = R3 * R2;$
\State $R7 = R5 * R2;$
\State $R9 = R7 * R2;$
\State $ANG = 0.999896 * R1 - 0.330756 * R3 + 0.181946 * R5 - 0.0876858 * R7 + 0.021997 * R9;$
\If {$\abs{C1} < \abs{C2}$}
  \State $ANGLE = ANG;$
\Else
    \If {$ANG>0$}
    \State $ANGLE =  \frac{\pi}{2} - ANG;$
    \EndIf
     \If {$ANG<0$}
    \State $ANGLE = - \frac{\pi}{2} - ANG;$
    \EndIf
\EndIf
\end{algorithmic}
\label{alg_arctan}
\end{algorithm}


\ifCLASSOPTIONcaptionsoff
  \newpage
\fi



\pagebreak

\bibliographystyle{IEEEtran}

\bibliography{attitudesurvey.bib}
%

%

\begin{IEEEbiographynophoto}{Hussein Al Jlailaty}
received the B.E. and M.E. degrees in electro-mechanical engineering from the Lebanese University (FEA), in 2003 and 2006, respectively, and the M.S. degree in physics and the M.E. degree in mechanical engineering from the Lebanese University and from the American University of Beirut (AUB), Lebanon, in 2005 and 2019, respectively. He worked as a research assistant with the Vision and Robotics Lab (VRL) for two years.  He is currently pursuing
the Ph.D. degree in electrical engineering with
the American University of Beirut
(AUB).
His research interest lie in the areas of robotics and digital signal processing, inertial navigation systems and reliable designs by algorithms.
\end{IEEEbiographynophoto}

\begin{IEEEbiographynophoto}{Mohammad Mansour}
received the B.E. and M.E. degrees in computer and communications engineering from the American University of Beirut (AUB), Lebanon, in 1996 and 1998, respectively, and the M.S. degree in mathematics and the Ph.D. degree in electrical engineering from the University of Illinois at Urbana–Champaign (UIUC), Champaign, IL, USA, in 2002 and 2003, respectively. He is currently a tenured Professor and Chairperson of the ECE Department at AUB. His research interests are in energy-efficient, high-performance, and reliable designs by algorithm, architecture, and circuit co-optimizations, with emphasis on emerging applications in 5G wireless communications, signal processing, deep learning networks, computing, and security. He has held visiting and consulting positions with Qualcomm, Broadcom, Intel, and Tensorcom, where he was involved in algorithm and architecture design for baseband receivers and computing systems. He is an active Senior Member of the IEEE.
\end{IEEEbiographynophoto}





\end{document}